\documentclass[preprints,article,accept,pdftex,moreauthors]{Definitions/mdpi} 
\firstpage{1} 
\pubvolume{1}
\issuenum{1}
\articlenumber{0}
\pubyear{2022}
\copyrightyear{2022}
\datereceived{} 
\dateaccepted{} 
\datepublished{} 
\hreflink{https://doi.org/} 

\usepackage{bm}
\usepackage{subcaption} 

\Title{Fast and Noise-Resilient Magnetic Field Mapping on a Low-Cost UAV Using Gaussian Process Regression}

\TitleCitation{Fast and Noise-Resilient Magnetic Field Mapping on a Low-Cost UAV Using Gaussian Process Regression}

\Author{Prince E. Kuevor $^{1,}$*\orcidA{}, 
Maani Ghaffari$^{2}$\orcidD{}, 
Ella M. Atkins $^{3}$\orcidC{}, and
James W. Cutler $^{4}$\orcidB{}}

\AuthorNames{Prince E. Kuevor, James W. Cutler and Ella M. Atkins}

\AuthorCitation{Kuevor, P.E.; Ghaffari, M.; Atkins, E.M.; Cutler, J.W.}

\address{%
$^{1}$ \quad Robotics Department, University of Michigan, Ann Arbor, MI 48109, USA\\
$^{2}$ \quad Naval Architecture and Marine Engineering, University of Michigan, Ann Arbor, MI 48109, USA;maanigj@umich.edu (M.G.) \\
$^{3}$ \quad Aerospace and Ocean Engineering, Virginia Tech, Blacksburg, VA 24061, USA;  ematkins@vt.edu (E.M.A.) \\
$^{4}$ \quad Aerospace Engineering, University of Michigan, Ann Arbor, MI 48109, USA; jwcutler@umich.edu (J.W.C.) \\
}

\corres{Correspondence: kuevpr@umich.edu (P.E.K.)}

\abstract{
This work presents a number of techniques to improve the ability to create magnetic field maps on a UAV which can be used to quickly and reliably gather magnetic field observations at multiple altitudes in a workspace. 
Unfortunately, the electronics on the UAV can introduce their own magnetic fields, distorting the resultant magnetic field map.
We show methods of reducing and working with UAV-induced noise to better enable magnetic fields as a sensing modality for indoor navigation. 
First, some gains in our flight controller create  high-frequency motor commands that introduce large noise in the measured magnetic field. 
Next, we implement a common noise reduction method of distancing the magnetometer from other components on our UAV. 
Finally, we introduce what we call a \textit{compromise} GPR (Gaussian process regression) map that can be trained on multiple flight tests to learn any flight-by-flight variations between UAV observation tests. 
We investigate the spatial density of observations used to train a GPR map then use the compromise map to define a \textit{consistency test} that can indicate whether or not the magnetometer data and corresponding GPR map are appropriate to use for state estimation. 
The interventions we introduce in this work facilitate indoor position localization of a UAV whose estimates we found to be quite sensitive to noise generated by the UAV.
}


\begin{document}

\section{Introduction}

Earth's magnetic field provides a reference measurement anywhere on the planet to assist with navigation. Magnetometers, or 3D compasses, onboard many modern robotics platforms can measure the local magnetic field to assist with navigation. To do this, we often rely on simplifying assumptions (i.e., the magnetic field points northward and is constant in a target region) or on maps of Earth's magnetic field like the World Magnetic Model (WMM). 
However, when testing in indoor environments, these assumptions and worldwide maps are inaccurate due to the contribution of the magnetic field from metallic objects inside buildings. Because of this, measurements from magnetometers are often ignored for indoor navigation. 

There are a number of existing works that leverage indoor magnetic field measurements to estimate position, or attitude \cite{Kuevor2021, Kok2018, Vallivaara2011, Vallivaara2010, Akai2015, Akai2017, Suksakulchai2000, Li2012, Haverinen2009, Wu2019}.
However, a common theme with the position localization and SLAM (simultaneous localization and mapping) papers \cite{Robertson2013, Vallivaara2010, Vallivaara2011, Wang2017} is that they do not investigate the accuracy their magnetic map. 
Thus it is not clear if their maps can be used as a valuable prior for future experiments in the mapped region. 
Similarly, for indoor attitude estimation, many previous works see Earth's magnetic field (mapped by WMM) as the signal and any magnetic fields from the building itself to be a source of noise that must be rejected \cite{Afzal2011, Yin2018, Karimi2020}. Though this paradigm works for attitude estimates, it strips away valuable information that is essential to using magnetic fields for position localization since Earth's magnetic field is fairly constant over many kilometers.

\begin{figure*}
    \centering
  \captionsetup{justification=centering}
  \includegraphics[width= 0.98\textwidth,trim= 10 100 20 80, clip]{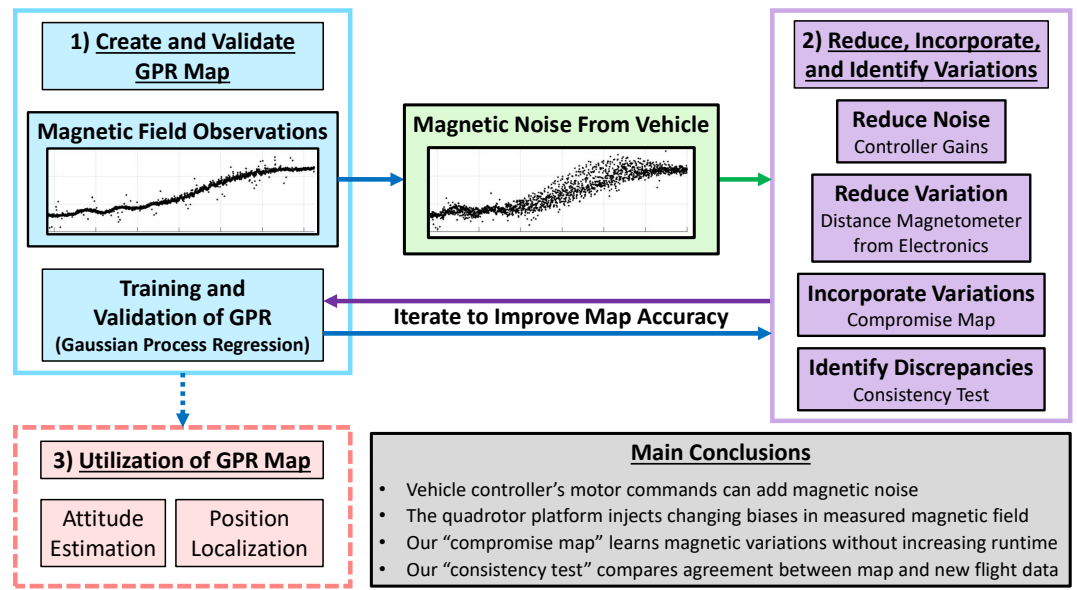}
  \caption{Methods to reduce, incorporate, and identify vehicle-induced magnetic field noise. 
  Our suggested interventions can be progressively added until the user's GPR map is sufficiently accurate for the application.}
 \label{fig:paper_outline}
 \vspace{-10pt}
\end{figure*}

This paper, outlined in Figure \ref{fig:paper_outline}, aims to augment indoor sensor suites with magnetic field data by demonstrating how to create and validate magnetic field maps of indoor spaces. 
The key idea is to enable the creation of local indoor magnetic field maps that can then be leveraged to improve position and attitude estimates. 
The value of indoor magnetic field maps is demonstrated in works that use them to estimate the position of pedestrians 
and robotic vehicles 
inside buildings and to estimate the attitude of a drone indoors.
As shown by ref. \cite{Akai2017}, gathering observations for a magnetic field map by hand is time-consuming and does not scale to mapping increasingly large spaces.
As such, we propose using an unmanned aerial vehicle (UAV) to gather observations in the target volume of the workspace, using Gaussian Process Regression (GPR) to interpolate between these observations, then finally leveraging the map in future experiments within the mapped volume. 

The contributions of this work are as follows.
\begin{enumerate}
    \item The magnetic noise induced by the motors and ESCs on a UAV can be reduced by removing high-frequency commands from the flight controller to the motors.
    \item Our UAV produces magnetic biases that occasionally change in magnitude and direction. We were unable to identify the underlying cause but found that distancing the magnetometer from the electronics, a commonly used intervention to reduce the measured magnetic field noise, decreased the measured variation of these biases.
    \item Our ``compromise map'' can be trained on large datasets to prevent overfitting to a single flight test and learn the flight-by-flight variations of the drone without incurring high computation costs when predicting the magnetic field.
    \item We find that our compromise map has similar accuracy if the location of training observations is no further than 0.55m apart. This agrees with a similar, qualitative study by Akai and Ozaki in ref. \cite{Akai2017}.
    \item We introduce a consistency test to indicate if a given GPR map has good agreement with new measurements from a subsequent flight test. This consistency test was important for contextualizing our position localization accuracy in another project.
    \item In a comparison of two methods of mapping the norm (magnitude) of the magnetic field, we find (empirically) that creating a specialized map on the norm of the field is equivalent to taking the norm of estimates from a vector-valued magnetic field map.
\end{enumerate}

At a high level, this paper rings similar to our previous work in \cite{Kuevor2021} where we used a drone to create a magnetic field map, interpolate with GPR, then finally demonstrate the value of the map in improving attitude estimates of the drone. 
This work focuses on improving the accuracy of magnetic field maps in the face of vehicle-induced noise and comparing different methods of creating magnetic field maps.

The remainder of this paper is structured as follows. 
Section \ref{sec:related_work} presents related works, and Section \ref{sec:math} introduces our mathematical notation for Gaussian process regression and our compromise map.
Section \ref{sec:experimental_procedure} introduces our UAV and testing methodology for this work.
Finally, Section \ref{sec:results_and_discussion} presents our results, while conclusions and future work appear in Section \ref{sec:conclusion_future_work}.

\section{Related Works}
\label{sec:related_work}

A magnetic field map must have 
an input dimension $p$ and output dimension $m$, which we will denote as a $p \rightarrow m$ map. 
The choice of input dimension $p$ often depends on the agent making the map. 
For example, $p=2$ for wheeled robots 
\cite{Li2012, Vallivaara2011, Vallivaara2010, Wu2019, Akai2015, Hanley2021} 
and most pedestrian localization while $p=3$ for UAVs \cite{Brzozowski2016, Brzozowski2017}, multi-floor pedestrian localization \cite{Kok2018}, or specially-outfitted ground robots \cite{Akai2017, Hanley2021}. 
The output $m$ is set by how the map will be utilized. 
Works like \cite{Suksakulchai2000, Haverinen2009} use $m=1$ for the orientation or magnitude of the magnetic field vector, while other works like \cite{Kuevor2021, Vallivaara2010, Vallivaara2011, Wu2019, Li2012, Akai2015, Akai2017, Kok2018} use all three vector components of the magnetic field with $m=3$.

A number of works using indoor magnetic field maps are focused on pedestrian localization. Such works often utilize foot-mounted, or calf-mounted sensor suites \cite{Frassl2013, Robertson2013, Liu2021} that can leverage zero-velocity updates to help with localization. Alternatively, many such works utilize the ubiquity of smartphones \cite{Gozick2011} and even consider constructing maps by crowdsourcing magnetic field measurements from many users \cite{Ayanoglu2018}. To our knowledge, all pedestrian-localization-based magnetic field maps are $2 \rightarrow m$ (or ``$2.5 \rightarrow m$'' maps \cite{Liu2021}), which assume a constant (or near-constant) altitude. 

Ground vehicles are another common platform for indoor magnetic field mapping \cite{Li2012, Vallivaara2011, Vallivaara2010, Wu2019, Akai2015, Hanley2021, Almeida2021, Frassl2013}.
These platforms benefit from wheel encoder odometry to assist their state estimates. Most of such works also create $2 \rightarrow m$ maps since the platform is confined to the ground. However, some researchers outfit their wheeled robots with a magnetometer on a vertical actuator \cite{Akai2017} or vertically-spaced magnetometers \cite{Hanley2021} that let them create $3 \rightarrow m$ maps. The latter is a recent work by Hanley et al. showing that indoor magnetic fields can be quite sensitive to changes in altitude \cite{Hanley2021} and emphasizing some issues with $2 \rightarrow m$ planar magnetic field maps.

UAVs are used more regularly for \textit{outdoor} magnetic field surveys than for \textit{indoor} mapping.
A recent review paper by Zhang et al. \cite{Zheng2021} presents works on outdoor magnetic field surveillance and also addresses methods of characterizing and suppressing UAV-induced magnetic noise. From the papers in their review, it is clear that the size and cost of the outdoor survey vehicles and sensors tend to be much larger. For example, \cite{Versteeg2007} mounted a Geometrics G823A cesium magnetometer on a gas-powered helicopter and measured 800nT of magnetic variation caused by components on the vehicle when it is not powered. This reduced to 80nT and 40nT when the magnetometer was attached to a boom of length 0.5m and 1.2m, respectively. Additionally, they consider vehicles that cost \$2K - \$45K USD and weigh 7lbs - 51lbs. 
The larger vehicle size of outdoor magnetic surveys allows for interventions like suspending a magnetometer at the end of a 4.5m-long cable \cite{Koyama2013}.
Thus, the vehicle size and project budget of outdoor surveying techniques allow for solutions to vehicle-induced magnetic noise (and magnetometers with better sensing capabilities) that are not applicable to (or not used on) indoor platforms.
Zheng et al. cover other UAV noise characterization and mitigation efforts for outdoor magnetic surveys in Section 3 of ref. \cite{Zheng2021}.

By contrast, some works use UAVs with magnetometers indoors, but do not create a map of the magnetic field for their implementation.
Brzozowski et al. describe methods to \textit{support} the use of indoor magnetic fields on UAVs in refs. \cite{Brzozowski2016, Brzozowski2017}, but mostly present methods of gathering and visualizing magnetic field observations. They stop short of interpolating their magnetic field observations to create a queryable map or perform any state estimation.
Furthermore, they do not actually use a flight vehicle (real or simulated) to gather or validate their magnetic field measurements. 
The authors in ref. \cite{Vasiljevic2022} perform 3D position localization of a UAV near a two-wire power line setup in their lab space, but do not attempt to map the magnetic field near the wires in the process.
Li et al. estimate the 6DOF pose of a UAV indoors using magnetometers (among other sensors), but only use the magnetometer for heading estimates and achieve this without a magnetic field map \cite{Li2019}.
Zahran et al. leverage hall effect sensors (a type of magnetometer) in a clever way to estimate the velocity of a quadrotor and improve the dead reckoning of such flight vehicles (again, without magnetic field maps) \cite{Zahran2019}. 

There are works that use UAVs to map signals like ultra-wideband (UWB) radio \cite{Tiemann2018} and visual light communication (VLC) \cite{Niu2021}. Some works often use the term ``fingerprinting'' to describe the association of unique features in a reference signal to specific locations in a workspace. In this work, we describe the same process as mapping.

To our knowledge, however, there are only two works that use a UAV (drone, multirotor, flight vehicle, etc...) to make or leverage \textit{indoor} magnetic field \textit{maps}. 
One is our previous work \cite{Kuevor2021} which uses Gaussian process regression (GPR) to make $3 \rightarrow 3$ magnetic field maps of the DC magnetic field in an indoor workspace of dimensions [4$\times$3$\times$2.25]m. 
The other work, by Lipovsky et al. \cite{Lipovsky2021}, uses observations from a UAV and cubic spline interpolation to create $3 \rightarrow 3$  maps of the DC, 50Hz, and 150Hz magnetic fields in a [5$\times$5$\times$2.5]m room. Ref. \cite{Lipovsky2021} mounts their magnetometer 40cm away from the drone based on lessons learned from their previous investigation on drone-induced magnetic noise \cite{Blazek2018}.

In this work, we use a UAV to create $3 \rightarrow 3$ indoor magnetic field maps. 
A flight vehicle allows for quick and repeatable mapping and validation of an indoor volume, but can also impede accurate measurements of the ambient field with magnetic noise generated by the motors and ESCs (electronic speed controllers). 
The focus of this work is on identifying the type of magnetic noise our UAV creates, reducing the amount of magnetic noise, and finally incorporating any remaining measurable variations into the magnetic field map. 

Additionally, we study how the spatial density of magnetic observations affect the accuracy of a magnetic field map (similar to a study done in \cite{Akai2017}) and empirically demonstrate the equivalence of two maps that estimate the norm of the ambient magnetic field. 
Finally, we introduce a consistency check to reason about when a GPR-based magnetic field map is appropriate to use for state estimation.

The interventions and studies done in this paper are lessons we learned while investigating the 3D position localization of a UAV using a low-cost IMU and magnetometer.
Though a presentation of our position localization methods and accuracy is outside the scope of this paper, 
it is important to emphasize that some of our interventions may not be needed for other applications of indoor magnetic field maps (e.g., they were not applied for our study on attitude estimation using indoor magnetic field maps \cite{Kuevor2021}).


\section{Mathematical Preliminaries and Methodology}
\label{sec:math}

This section introduces our use of Gaussian Process Regression (GPR) to interpolate a set of magnetic field observations in our workspace. Special notation is used to distinguish a set of $n_2$ observations used to train hyperparameters and a separate set of $n_1$ observations used to perform inference. Additionally, we introduce performance metrics used later in our analysis of our GPR-based magnetic field maps.

We first introduce some notation. A single measurement of the magnetic field at an unspecified location is $\tilde{\bm{y}} \in \mathbb{R}^3$ with the $x$, $y$, and $z$ components of this measurement denoted as $\tilde{y}_x$, $\tilde{y}_y$, $\tilde{y}_z \in \mathbb{R}$ respectively. 
In general, a subscript of $_x$, $_y$, or $_z$ denotes that respective component of the magnetic field while an overhead tilde $\tilde{ }$ denotes a \textit{measured} value. 
$\tilde{\bm{Y}}^{n} \in \mathbb{R}^{3 \times n}$ is a set of $n$ magnetic field measurements while $\tilde{\bm{Y}}^{n}_z \in \mathbb{R}^{n}$ denotes the $z$ component of each magnetic field measurement in the set. 
Similarly, $\bm{X}^n \in \mathbb{R}^{3 \times n}$ is a collection of $n$ spatial locations in our workspace. 
The \textit{predicted} or \textit{estimated} magnetic field $\hat{\bm{m}}$ at some location $\bm{r} \in \mathbb{R}^3$ is denoted as $\hat{\bm{m}}(\bm{r})\in \mathbb{R}^3$ and the $x$ component of this prediction is $\hat{m}_x(\bm{r}) \in \mathbb{R}$. 
Thus, a collection of $n$ magnetic field estimates will be $\hat{\bm{M}}^n(\bm{X}^n)  \in \mathbb{R}^{3 \times n}$.
Similarly, $\hat{\bm{M}}^n$ is a set of predictions of the magnetic field where the location of these predictions is arbitrary and $\hat{\bm{M}}^n_y \in \mathbb{R}^n$ gives just the predictions from $\mathcal{GP}_y$.
Generally, an overhead hat $\hat{ }$ denotes a prediction/estimate while a superscript integer $n$ denotes the number of measurements/predictions/locations of the respective matrix.




\subsection{Gaussian Process Regression}
\label{sec:gpr}
Gaussian Process Regression (GPR) is a machine learning tool that can be used to estimate a signal given a set of noisy measurements. Here, we use GPR as the backbone of our magnetic field map leveraging methods from Rasmussen et al. in \cite{carledwardrasmussen2005}. The goal is to have a magnetic field map of the flight workspace which will provide an estimated magnetic field vector at any location in the working volume.

This work uses GPR to create three separate $3 \rightarrow 1$ maps each responsible for estimating the $x$, $y$, and $z$ components of the magnetic field vector in our workspace. To create the GPR-based map, we first gather $n$ observations of the magnetic field throughout the workspace. 
Observation sets give the three components of the \textit{measured} magnetic field $\tilde{\bm{Y}}^n_x$, $\tilde{\bm{Y}}^n_y$, $\tilde{\bm{Y}}^n_z \in \mathbb{R}^n$ at each 3D position in our design matrix $\bm{X}^n \in \mathbb{R}^{3 \times n}$. Together, these quantities define the training sets $D_x = (\bm{X}, \tilde{\bm{Y}}_x)$, $D_y = (\bm{X}, \tilde{\bm{Y}}_y)$, and $D_z = (\bm{X}, \tilde{\bm{Y}}_z)$ for the $x$, $y$, and $z$ GPRs respectively. 

In Ref. \cite{carledwardrasmussen2005}, Rasmussen and Williams define a Gaussian process as a distribution over functions written as 
\begin{equation}
    f(\bm{r}) = \mathcal{N} \left(\bar{f}, \mathbb{V}(f) \right) \sim \mathcal{GP} \left(\bm{0}, k(\bm{r}, \bm{r}') \right)
    \label{eq:gp_definition}
\end{equation}

\noindent where $\bm{0}$ is the zero-mean function, $k$ is the covariance function (or kernel) and $\bm{r}$, $\bm{r}' \in \mathbb{R}^3$ are 3D positions. This work uses the squared exponential covariance function

\begin{equation}
    k(\bm{r}, \bm{r}') = \sigma_f^2 \exp \left( -\frac{1}{2l^2} (\bm{r} - \bm{r}') ^\top (\bm{r} - \bm{r}')  \right) + \sigma_n^2
\end{equation}

\noindent with $\sigma_f$ as the signal variance, $l$ as the length scale, and $\sigma_n$ as the sensor noise variance which define our set of \textit{hyperparameters} $\bm{\Theta} = \{ \sigma_f, l, \sigma_n\}$ for this kernel. In essence, the squared exponential kernel is a measure of similarity between two 3D locations $\bm{r}$ and $\bm{r}'$  (scaled by the hyperparameters) and will serve as a weighting term in the inference of the magnetic field at unobserved locations.

Recall that we perform regression using three Gaussian processes $\mathcal{GP}_x$, $\mathcal{GP}_y$, and $\mathcal{GP}_z$ to represent the full magnetic field anywhere in the workspace. 
As such, we need three sets of hyperparameters ($\bm{\Theta}_x$, $\bm{\Theta}_y$, $\bm{\Theta}_z$) each computed using the Gaussian Processes for Machine Learning \texttt{gpml-matlab} toolbox by minimizing the negative log marginal likelihood of the respective training sets $D_x$, $D_y$, and $D_z$. 
We define hyperparameters optimized over the observations in a training set $D_x$ as $\bm{\Theta}^*_x(D_x)$.
(\texttt{gpml-matlab} was created by Carl Edward Rasmussen and Hannes Nickisch \url{http://www.gaussianprocess.org/gpml/code/matlab/doc/} accessed on September 2021).
\label{fn:gpml}

Now, to estimate the magnetic field at some location $\bm{r}^*$, we need a training set $D$ and a set of hyperparameters $\bm{\Theta}$ corresponding to a selected kernel $k$. 
A squared exponential kernel $k$ using optimal hyperparameters $\bm{\Theta}^*(D)$ is denoted as $k_{\bm{\Theta}^*(D)}(\bm{r}, \bm{r}^*)$ where the subscript on the kernel $k$ specifies which set of hyperparameters are used and $\bm{r} \in \bm{X}$ is a location in the training set.

This is the extent of our notation, but an example is helpful. Say we want to estimate the  $z$ component of the magnetic field $\hat{m}_z(\bm{r}^*)$ at some location $\bm{r}^*$. 
For this, $\mathcal{GP}_z$ requires a kernel $k$ to compare the target location $\bm{r}^*$ against locations $\bm{X}^{n_1}$ in its training set $D_z^{n_1} = (\bm{X}^{n_1}, \tilde{\bm{Y}}_z^{n_1})$. Additionally, its hyperparameters $\bm{\Theta}^*(D_z^{n_2})$ are optimized over a separate set of observations $D_z^{n_2} = (\bm{X}^{n_2}, \tilde{\bm{Y}}_z^{n_2})$. Mathematically, we express this as


\begin{equation}
\begin{split}
    \hat{m}_z(\bm{r}^*) & = GPR_z(\bm{r}^*) \\
    & = \mathbb{E} \left[ \mathcal{GP}_z \left( \bm{0}, \; K_{\bm{\Theta}_z^*(D_z^{n_2})} (\bm{X}^{n_1}, \bm{r}^* ) \right) \right]
\end{split}
\label{eq:gpr_inference}
\end{equation}
\noindent where a matrix $K(\bm{X}, \bm{r}^*)$ has scalar elements $k\left(\bm{X}^{\{i\}}, \bm{r}^* \right)$ with each $\bm{X}^{\{i\}} \in \mathbb{R}^3$ a column of $\bm{X} \in \mathbb{R}^{3 \times n}$ while $\hat{m}_z(\bm{r}^*) \in \mathbb{R}$. 
Finally, $GPR_z(\bm{r}^*)$ is a shorthand for the predicted magnetic field value of the $z$-component Gaussian process $\mathcal{GP}_z$ at $\bm{r}^*$.

The reason for the verbosity in Equation \ref{eq:gpr_inference} is to allow an observation set for optimizing hyperparameters $D_z^{n_2} = (\bm{X}^{n_2}, \tilde{\bm{Y}}_z^{n_2})$ and a separate ``inference set'' $D_z^{n_1} = (\bm{X}^{n_1}, \tilde{\bm{Y}}_z^{n_1})$  for computing the similarity weights when predicting the magnetic field at some location $\bm{r}^*$. 
This is important because hyperparameter optimization, typically done offline, scales with $\mathcal{O}\left({n_2}^3 \right)$ while inference scales with $\mathcal{O}\left({n_1}^2 \right)$. With this formulation, our map can encode information from a large set of flights offline, then use a subset of these observations during the actual inference to reduce computational load. We leverage this idea in Section \ref{sec:create_query_gpr} to create and query our magnetic field map.

\subsection{Magnetometer Calibration}
\label{sec:mag_calibration}

To calibrate our vehicle's magnetometer, we use the model and iterative least squares solver from Ref. \cite{Springmann2012_magCalibration} along with the two-step calibration procedure from Ref. \cite{Wu2020}. 
The magnetometer model from \cite{Springmann2012_magCalibration} is repeated here
\begin{gather}
 \tilde{B}_x = \theta_a B_x + \theta_{x_0} + \eta_x
 \label{eq:magModelX} \\
 \tilde{B}_y = \theta_b (\tilde{B}_y \cos( \theta_\rho) + B_x \sin(\theta_\rho)) + \theta_{y_0} + \eta_y
 \label{eq:magModelY} \\
 \begin{split}
  \tilde{B}_z &= \theta_c(B_x \sin(\theta_\lambda) + B_y \sin(\theta_\phi)\cos(\theta_\lambda) \\
  &+ B_z \cos(\theta_\phi)\cos(\theta_\lambda)) + \theta_{z_0} + \eta_z
 \end{split}
 \label{eq:magModelZ}
\end{gather}

\noindent where ($\tilde{B}_x$, $\tilde{B}_y$, $\tilde{B}_z$) are measured magnetic field values, ($B_x$, $B_y$, $B_z$) are the true magnetic field, and the parameters to solve for are bias ($\theta_{x_0},\theta_{y_0}, \theta_{z_0}$), scale factor ($\theta_{a}, \theta_{b}, \theta_{c}$), and non-orthogonality terms ($\theta_{\rho}, \theta_{\lambda}, \theta_{\phi}$). 

Ultimately, we aim to find parameters $\theta = [\theta_{a}, \theta_{b}, \theta_{c}, \theta_{x_0}, \theta_{y_0}, \theta_{z_0}, \theta_{\rho}, \theta_{\lambda}, \theta_{\phi}]^\top$ by minimizing the following cost function via an iterative non-linear least squares solver
\begin{equation}
 \begin{split}
  \Delta B &= B_R^2 - B^2 \\
  & = B_R^2 - (B_x^2 + B_y^2 + B_z^2) \\
  &= B_R^2 - g(\tilde{B}_x, \tilde{B}_y, \tilde{B}_z, \alpha)
 \end{split}
\end{equation}

\noindent where $B_R$ is the reference magnetic field strength (taken from the World Magnetic Model \cite{WMM2020}) and $g()$ is obtained by solving Equations \ref{eq:magModelX} - \ref{eq:magModelZ} for ($B_x, B_y, B_z)$. 

The key difference from Springmann's method \cite{Springmann2012_magCalibration} is to estimate the nine $\theta$ parameters in two separate minimization steps like Wu et. al. demonstrate in \cite{Wu2020}. 
We found that this two-step approach gave more consistent results for another magnetometer (not used in this work) on our UAV.
As such, we adopted the two-step calibration technique for our primary magnetometer as well.

First, a simplified magnetometer model is created by setting all scaling terms to 1 and non-orthogonalities to 0 leaving just the bias terms ($\theta_{x_0},\theta_{y_0}, \theta_{z_0}$) in Equations \ref{eq:magModelX} - \ref{eq:magModelZ}. The first minimization is done on this simplified model to find an optimal set of bias terms ($\theta_{x_0}^*,\theta_{y_0}^*, \theta_{z_0}^*$) which are in turn used as initial conditions for the second optimization where
$\theta_0 =$
$[1, 1, 1,$
$\theta_{x_0}^*,\theta_{y_0}^*, \theta_{z_0}^*,$
$0, 0, 0]^\top$. 
The nine optimal parameters from this second optimization are used as the calibration terms for the magnetometer. 



\subsection{Performance Metrics}
We use the root mean squared error (RMSE) of each GPR's prediction $\hat{m}_x$ against the corresponding component of the measured field $\tilde{y}_x$ across a validation set. With this, the RMSE for $\mathcal{GP}_x$ over some validation set $D^{n_v}_x = (\bm{X}^{n_v}, \tilde{\bm{Y}}_x^{n_v})$ is defined as
\begin{equation}
    RMSE_x = \sqrt{ \frac{1}{n_v} \sum_{i=1}^{n_v} (\hat{m}^{\{ i\}}_x - \tilde{y}^{\{ i\}}_x)^2}
    \label{eq:rmse}
\end{equation}
where $\hat{\bm{M}}_x^{n_v} = GPR_x(\bm{X}^{n_v})$ is the  $x$ component of the predicted magnetic vector, $\hat{m}^{\{ i\}}_x$ is the $i$th prediction in $\hat{\bm{M}}_x^{n_v}$, and $\tilde{y}^{\{ i\}}_x$ is the corresponding $i$th measurement in $\tilde{\bm{Y}}_x^{n_v}$.

Although we have three GPRs, it is sometimes convenient to summarize the accuracy of their composite estimate as 
\begin{equation}
    RMSE_{norm} = \sqrt{{RMSE_x}^2 + {RMSE_y}^2 + {RMSE_z}^2}.
    \label{eq:rmse_norm}
\end{equation}

\section{Experimental Procedure}
\label{sec:experimental_procedure}
This section introduces our quadrotor unmanned aerial vehicle (UAV), testing locations, and flight trajectories used to construct and validate our magnetic field maps.

\subsection{Equipment, Facilities, and Setup}
\begin{figure}
    \centering
  \captionsetup{justification=centering}
  \includegraphics[width= 0.5\textwidth,trim= 270 110 270 20, clip]{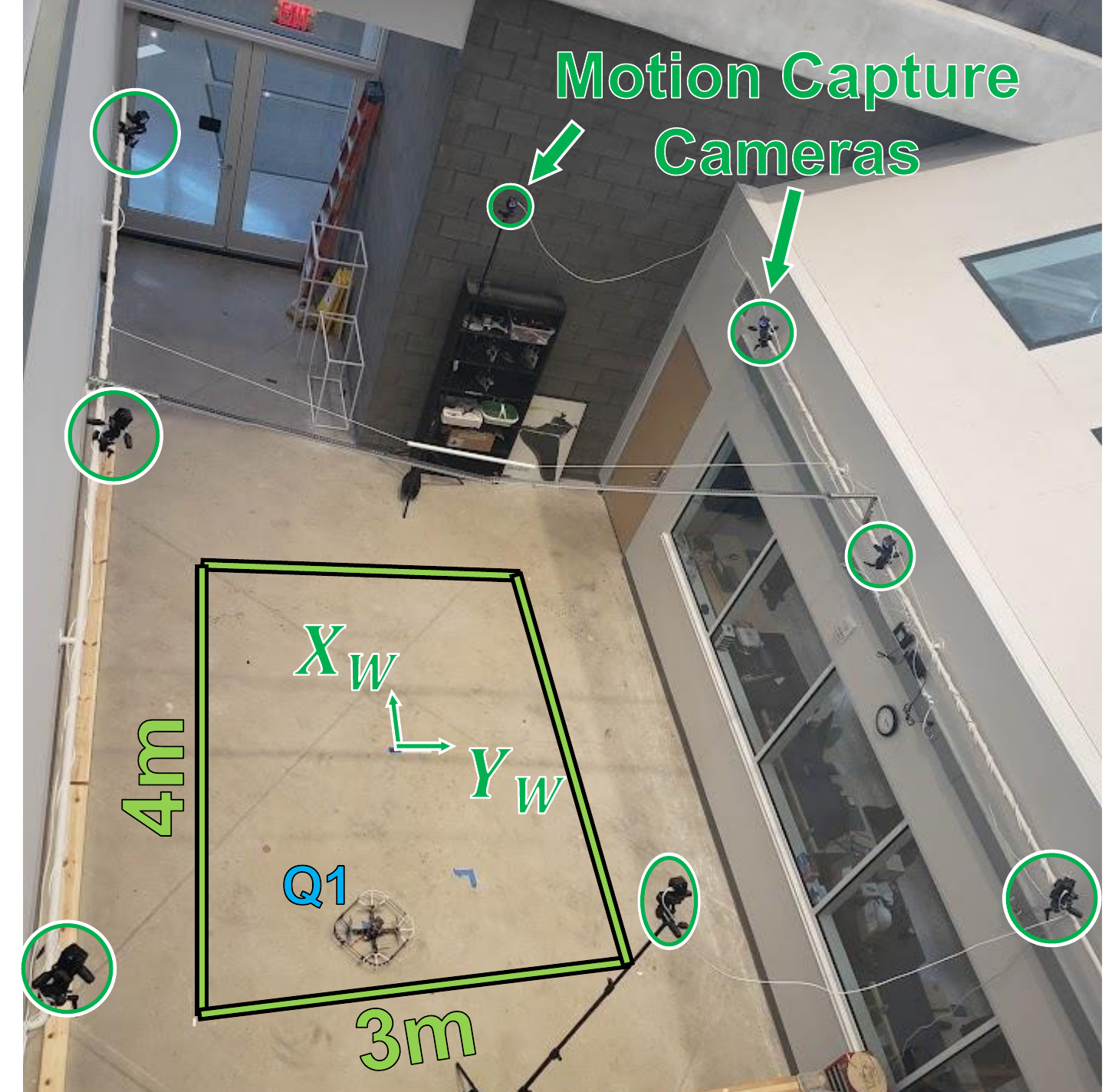}
  \caption{Flight vehicle Q1 in the Robot Fly Lab within the Ford Motor Company Robotics Building at the University of Michigan.}
 \label{fig:indoor_flight_space}
\end{figure}

All tests were conducted at the Robot Fly Lab in the University of Michigan Ford Motor Company Robotics Building. 
This indoor flight arena is equipped with eight OptiTrack motion capture cameras that give a working volume of 4m $\times$ 3m $\times$ 2.25m as shown in Figure \ref{fig:indoor_flight_space}. 
A ground station computer connected to the OptiTrack system provides ground-truth position and attitude estimates of the vehicle at 120Hz. 
The communication setup for the ground station, pilot transmitter, and flight vehicle is the same as that depicted in Figure 3 of Ref. \cite{Kuevor2021}, but with one less BeagleBone Blue on the UAV.

The motion capture pose estimate is streamed to the flight vehicle in real-time, but with a communication latency of about 40ms. For this, the vehicle's onboard estimate of roll and pitch is used for control while the remaining states (position and heading) are taken from the 40ms-delayed motion capture packets. 

Time synchronization is important to properly associate each onboard magnetometer measurement with a ground-truth pose when creating and validating our magnetic field maps. 
During flights, our UAV's BeagleBone Blue is synchronized with the ground station laptop and ground station BeagleBone Green using a Linux tool called \texttt{chrony} which was created by the Red Hat Software company (\url{https://chrony.tuxfamily.org/} accessed on February 2022).
\label{fn:chrony}
In post-processing, this synchronization allows us to associate the non-delayed motion capture data with the drone's magnetometer readings.
Additionally, we use motion capture pose (position and attitude) estimates to rotate the magnetometer data into the world frame and determine the location each magnetic field measurement was collected.

This work uses time-invariant magnetometer calibration combining techniques from Refs. \cite{Springmann2012_magCalibration} and \cite{Wu2020} (see Section \ref{sec:mag_calibration}). Magnetometer calibration is performed outdoors, just South of the University of Michigan's outdoor netted flight facility, M-Air. This location is far enough away from any buildings that the magnetic field strength is constant over a few meters and should be accurately reflected by the World Magnetic Model (WMM) \cite{WMM2020}.
For the data gathered on September 1st, 2022 (test series t6), the ambient magnetic field reference term used in our calibration is $B_R = 53.1351\mu T$ as taken from a WMM online calculator for M-Air's location at 42.294431$^\circ$N, 83.710442$^\circ$W, and 270m above sea level.

\begin{figure}
\centering
 \begin{subfigure}[b]{0.49\textwidth}
  \captionsetup{justification=centering}
  \includegraphics[width=\textwidth,trim= 120 85 100 100, clip]{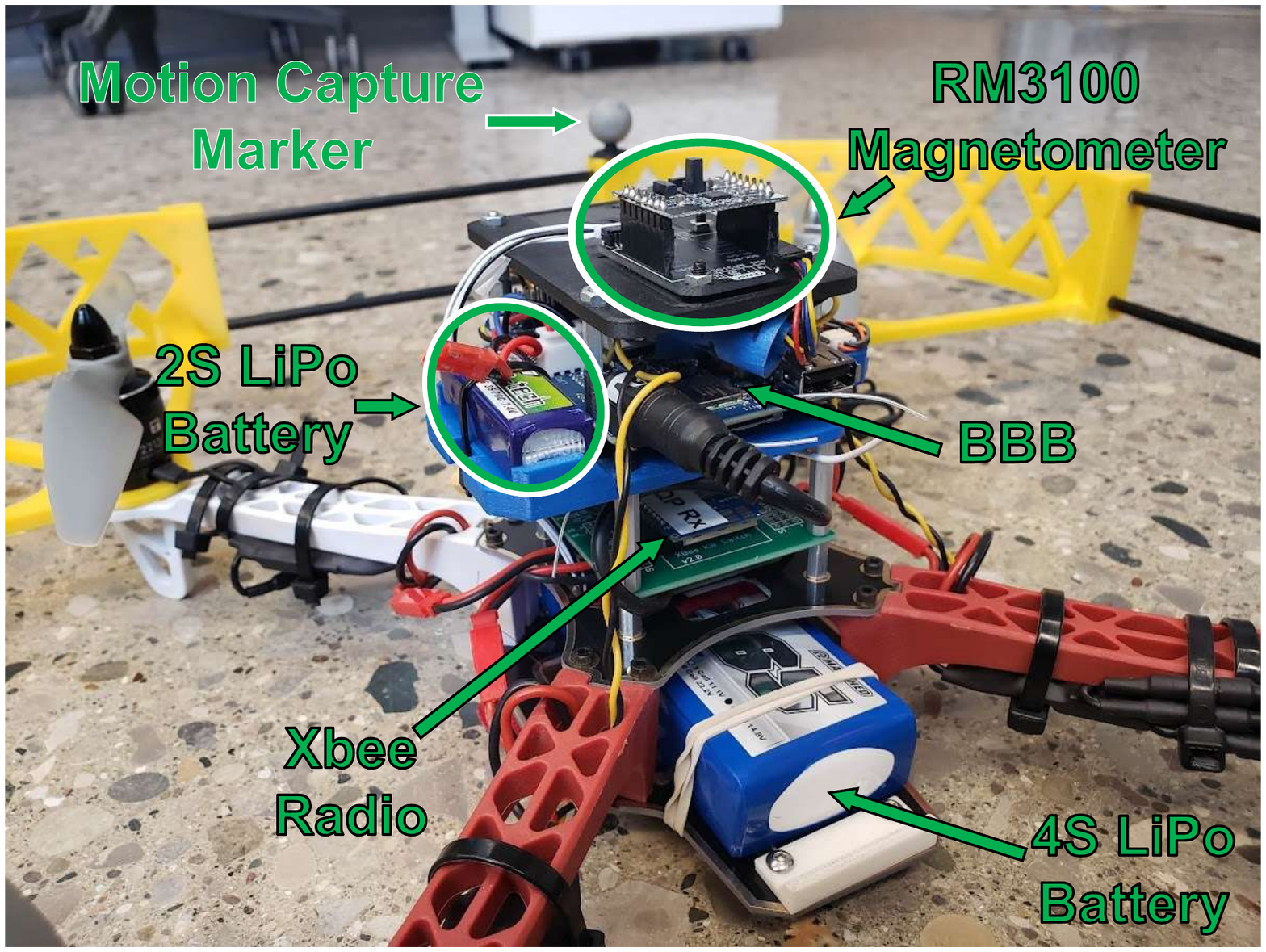}
  \caption{Main components of Q1. $S_2$ = 2cm.}
  \label{fig:q1_avionics_pic}
 \end{subfigure}
 \begin{subfigure}[b]{0.39\textwidth}
  \captionsetup{justification=centering}
  \includegraphics[width=\textwidth,trim= 230 200 230 100, clip]{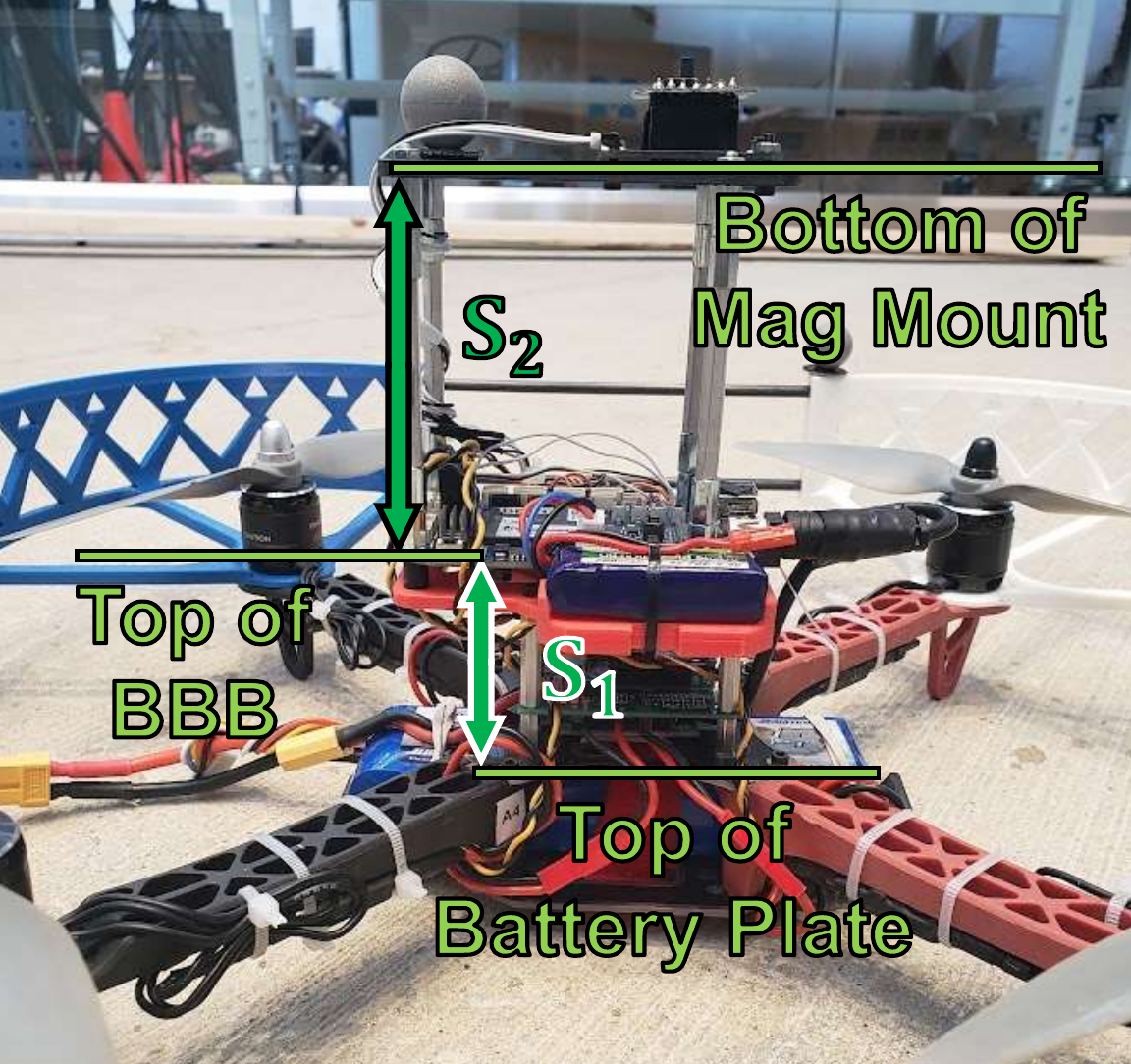}
  \caption{Definition of $S_1$ and $S_2$. $S_2$ = 8cm.}
  \label{fig:q1_vertical_dist}
 \end{subfigure}
 \caption{Q1: The flight vehicle used for experiments in this work.}
 \label{fig:drone_pic}
\end{figure}

Our flight vehicle ``Q1'' (Quadrotor one) is shown in Figure \ref{fig:drone_pic} and has an RM3100 magnetometer that is sampled at 200Hz and an MPU9250 IMU (gyroscope, accelerometer, and magnetometer) sampled at 200Hz. This work made no use of the magnetometer on the MPU9250 relying solely on the RM3100 for all magnetic field measurements. The BeagleBone Blue (BBB) microprocessor handles communication with the ground station via Xbee radio, logs all sensor data locally, and commands the four motors to achieve the desired flight trajectory. A Turnigy 4S 2650mAh 20C LiPo battery provides the main power on Q1 while a Turnigy 2S 300mAh LiPo battery keeps the BBB powered on when swapping 4S batteries between flights.

Figure \ref{fig:q1_vertical_dist} shows Q1 with different distances from the magnetometer to the rest of the vehicle's electronics. $S_1$ is defined as the distance from the top of the battery mounting plate to the top of the BBB. This distance is $S_1$ = 4.79cm on Q1 and will not change in any experiments through this paper. 
$S_2$ is an adjustable distance defined from the top of the BBB to the bottom of the magnetometer's mounting plate that allows us to investigate how the magnetometer's proximity to other electronics affects the consistency of magnetic field measurements on a quadrotor.

Finally, we placed a stationary RM3100 magnetometer in the corner of the flight lab that gathers data every 10 seconds (0.1Hz). This stationary sensor is on the ground at approximately (-3.5m, +2.5m) in the ($X_W$, $Y_W$) frame defined in Figure \ref{fig:indoor_flight_space} and was used to confirm that the ambient magnetic field remained constant during our experiments.


\begin{figure}
    \centering
  \captionsetup{justification=centering}
  \includegraphics[width= 0.7\textwidth,trim= 20 160 90 180, clip]{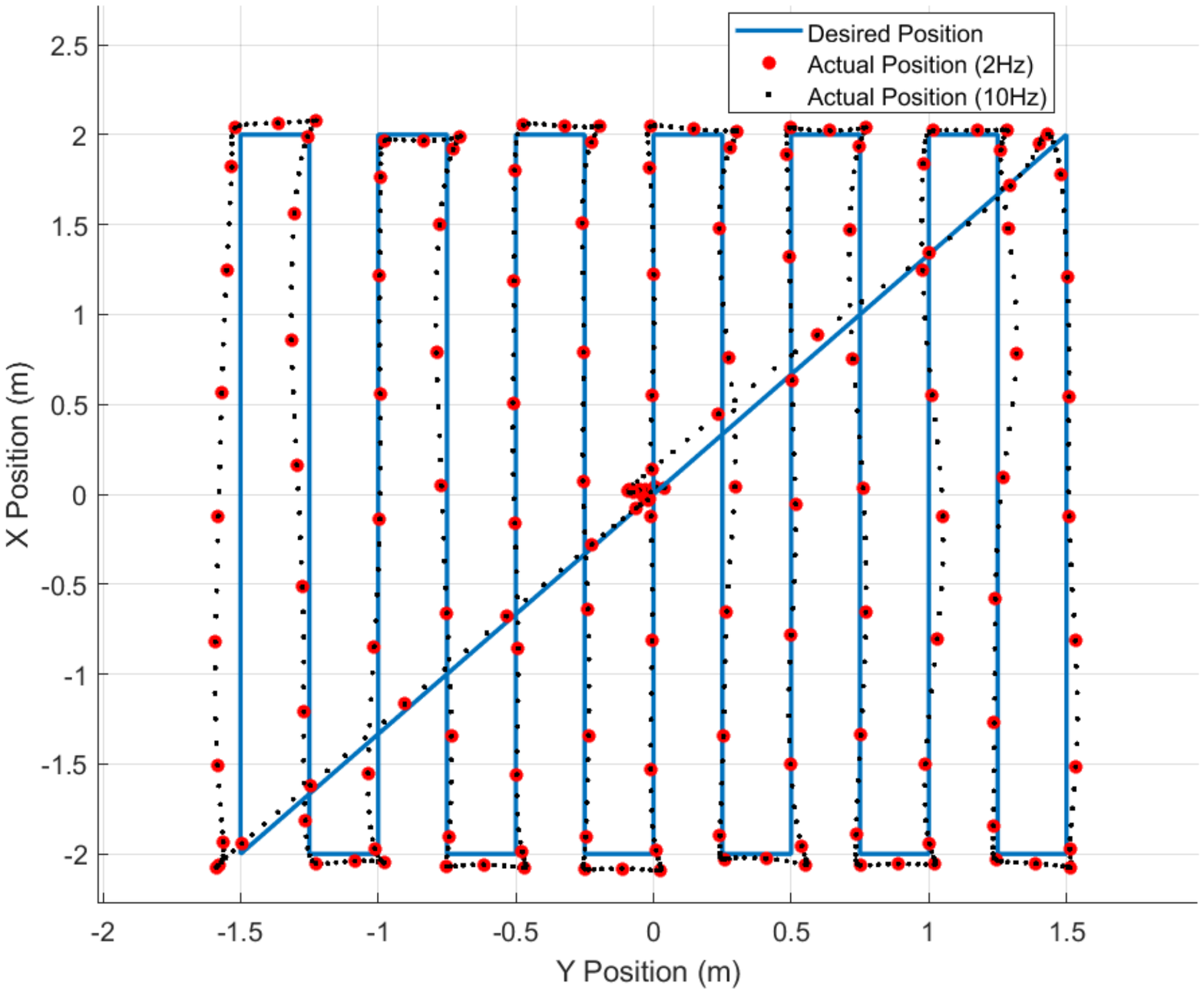}
  \caption{Single-altitude lawnmower trajectory used in all trajectories in this work.}
 \label{fig:1_5m_traj}
\end{figure}

\subsection{Flight Profiles}
\label{sec:flightProfiles}

The flight tests conducted for this work are all designed around single-altitude scanning patterns that gather observations at a planar slice of the working volume. 
Figure \ref{fig:1_5m_traj} shows the desired trajectory (blue, solid line) has 4m $x$-axis strides separated by 0.25m $y$-axis strides. 
Since all flights begin with the drone at the origin (0m, 0m) of our working volume, the trajectory in Figure \ref{fig:1_5m_traj} includes diagonal strides to leave and return to the origin at the start and end of each flight test.
Finally, some trajectories gather observations at multiple altitudes in a single flight and include $z$-axis strides between each planar trajectory. For multi-altitude tests, the drone traverses a single diagonal stride from the origin to the corner at the first altitude, and another diagonal stride from the corner to the origin at the final altitude.

All linear strides are created from quintic spline trajectories that are hand-tuned to enforce a 1.9m/s speed limit which is a compromise between the desire for shorter flight tests (that allow for more frequent data collection) and the limitations of our flight controller. 

The 0.25m $y$-axis spacing is selected to enable a study on how the spatial density of observations used to train a GPR-based map affects the accuracy of the said map (Section \ref{sec:compromise_map_spatial_density_analysis}). 
A similar study was done in ref. \cite{Akai2017} where magnetic field observations in a 3m $\times$ 3m $\times$ 2.2m volume are hand-gathered at 0.2m increments in all axes.
For our work, having 0.2m $y$-axis and 0.2m $z$-axis spacing that would scan all altitudes in our working volume created a flight trajectory duration that exceeded the maximum flight time of our vehicles.
Thus, a 0.25m minimum separation distance is used in this work instead. 

\subsubsection{Flight Profile Nomenclature}
Throughout this paper, we refer to flights with ID tags like ``tY\_XX'' where \textit{Y} refers to the flight test \textit{series} and \textit{XX} is the two-digit ID of the flight test in that series. 
This allows the reader to reference our raw experimental data to improve the reproducibility of our results. Appendix \ref{apndx:flight_test_names} has more details on the matter.

\subsection{Pre-Processing Magnetic Observations}
When creating our hyperparameter ($D^{n_2}$) and inference sets ($D^{n_1}$), we first pre-process the magnetic field data gathered by the RM3100. 

\begin{figure}
    \centering
  \captionsetup{justification=centering}
  \includegraphics[width= 0.7\textwidth,trim= 10 160 30 170, clip]{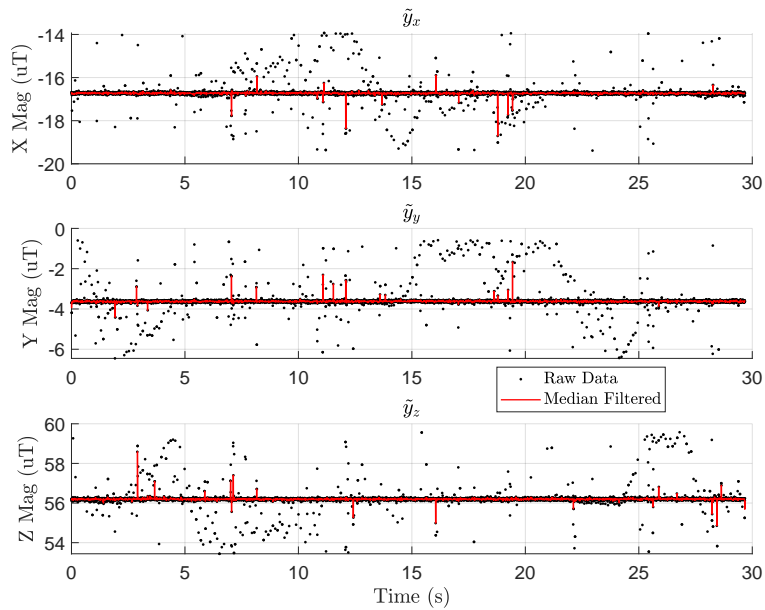}
  \caption{Data from a stationary RM3100 placed at the origin of our flight arena and sampled at 200Hz. Spurious measurements are reduced with a moving median filter with a window size of 5.}
 \label{fig:stationary_rm3100_data}
\end{figure}

Figure \ref{fig:stationary_rm3100_data} shows data from the UAV's RM3100 for a segment of time when the drone is not moving and the motors are not spinning. 
The black dots are raw magnetometer measurements that are mostly constant, but frequently have spurious measurements that vary by $\pm 3 \mu$T for each axis respectively. 
We have seen such spurious measurements from three different RM3100 sensors; even one mounted on a platform with no motors or ESCs (electronic speed controllers). 
Other works using RM3100s have not reported such spurious readings \cite{Hanley2021, Regoli2018}.
Thus, we believe this could be a result of our 200Hz sampling rate (600Hz for the whole device to achieve 200Hz per axis) and something with the RM3100's firmware that does not like such a fast sampling rate. Alternatively, this could be caused by some poor voltage regulation on the BeagleBone Blue. We have not tried sampling from other microprocessors to see if the BeagleBone is the cause of this issue.

Nonetheless, all RM3100 data we use in this work is first put through a moving median filter with a window size of 5.
The resultant observations from the median filter are shown in red in Figure \ref{fig:stationary_rm3100_data}.
Clearly, there are still some outliers in red, but increasing the window size of the moving median filter to remove these outliers delays the signal enough to cause problems for state estimation.
Since we want to construct and validate maps with the same pre-processing steps used when performing state estimation, we fix a window size of 5 and work with the remaining outliers.

In addition to a median filter, we downsample the 200Hz of data gathered during the flight test. Since the drone does not move very far in 1/200th of a second, many observations are spatially redundant. As such, we temporally downsample our observations to 2Hz, 4Hz or 10Hz when creating or validating our GPR maps. Figure \ref{fig:1_5m_traj} shows the spatial distribution of 2Hz and 10Hz downsampling during a flight test.

Since motion capture data is used to rotate the magnetometer data from the body frame to the world frame, we perform one final check to ensure a timely ground-truth pose estimate. 
Any temporally downsampled observation with a motion capture pose older than 50ms is considered ``stale'' and replaced with the nearest (in time) observation that has a more timely motion capture pose.
If the ``fresh'' replacement is already in the downsampled observation set, the redundant observation is removed leaving one instance of it in the final set.

\subsection{Creating and Querying the ``Compromise'' Map}
\label{sec:create_query_gpr}
As mentioned in Section \ref{sec:gpr}, creating a GPR-based map requires each Gaussian process $\{\mathcal{GP}_x$, $\mathcal{GP}_y$, $\mathcal{GP}_z\}$ to have a set of hyperparameters optimized over an observation set
$\{ 
\bm{\Theta}^*_x(D_x^{n_2}),$
$\bm{\Theta}^*_y(D_y^{n_2}),$
$\bm{\Theta}^*_z(D_z^{n_2}) 
\}$ 
and an inference set $\{ D_x^{n_1}, D_y^{n_1}, D_z^{n_1}\}$ to compare the target locations. 
This section explains how we use flight data to create the hyperparameter observation sets (with $n_2$ observations) and then use inference from an intermediate magnetic field map to make ``compromise'' inference sets (with $n_1$ observations) where $n_1 < n_2$. 
The goal of the compromise map is to leverage the flight-by-flight variation in magnetic field measurements (Section \ref{sec:s2_and_flight_by_flight_variations}) by training hyperparameters on $n_2$ observations from many flights  without incurring the computational cost of having a large inference set size $n_1$.

The ``hyperparameter observation set'' for the $y$ component Gaussian process $\mathcal{GP}_y$ is $D_y^{n_2} = (\bm{X}^{n_2}, \tilde{\bm{Y}}_y^{n_2})$ and similar for $D_x^{n_2}$ and $D_z^{n_2}$. All three hyperparameter observation sets share the same observed locations $\bm{X}^{n_2} \in \mathbb{R}^{3 \times n_2}$ taken from downsampled observations from any number of \textit{training} flights. 

These observations are used to compute $\{ \bm{\Theta}^*_x(D_x^{n_2}), \; \bm{\Theta}^*_y(D_y^{n_2}), \; \bm{\Theta}^*_z(D_z^{n_2}) \}$ by minimizing the negative log marginal likelihood of the observations over the hyperparameters of each respective Gaussian process. The optimal hyperparameters, along with their corresponding observation sets of size $n_2$, will serve as an ``intermediate'' magnetic field map.

From here, we query our intermediate map at $n_1$ user-selected locations to generate $n_1$ magnetic field observations for the compromise map. 
The result of this step is a set of $n_1$ locations around the workspace that make up $\bm{X}^{n_1} \in \mathbb{R}^{3 \times n_1}$.
The magnetic field ``measurements'' that complete the inference set $D_x^{n_1} = (\bm{X}^{n_1}, \hat{\bm{M}}_x^{n_1})$ come from the estimated magnetic field values at each location in $\bm{X}^{n_1}$ where 

\begin{equation}
    \hat{\bm{M}}_x^{n_1} = \mathbb{E} \left[ \mathcal{GP}_x \left( \bm{0}, \; K_{\bm{\Theta}_x^*(D_x^{n_2})} (\bm{X}^{n_2}, \bm{X}^{n_1}) \right) \right] \in \mathbb{R}^{n_1}
\label{eq:intermediate_map_inference}
\end{equation}
and similarly for the magnetic field ``measurements'' $\hat{\bm{M}}_y^{n_1}$ and $\hat{\bm{M}}_z^{n_1}$ for inference sets $D_y^{n_1}$ and $D_z^{n_1}$ respectively. Here, the GPR maps estimate the magnetic field at locations $\bm{X}^{n_1}$ by computing their similarity to locations $\bm{X}^{n_2}$. The notation used in Equation \ref{eq:intermediate_map_inference} was introduced in Section \ref{sec:gpr}.

At last, we have what we call our ``compromise inference sets'' $(D_x^{n_1}, D_y^{n_1}, D_z^{n_1})$ (or just ``inference sets'') comprised of magnetic field estimates from the intermediate map at $n_1$ user-selected locations around our workspace. 

Our construction of the compromise map gives two benefits. First, it allows our GPR-based maps to avoid overfitting to a single training flight by allowing large hyperparameter observation sets ($n_2$) that can incorporate observations from several flights of a quadrotor (or potentially even flights from multiple quadrotors for a multi-agent system). 
Next, it allows us to perform inference in $\mathcal{O}({n_1}^2 n_q)$ rather than $\mathcal{O}({n_2}^2  n_q)$ where $n_q$ is the number of points queried.

\section{Results and Discussion}
\label{sec:results_and_discussion}

This section starts by explaining how the derivative gains from the PID controller of our UAV's autopilot create measurable magnetic noise (Section \ref{sec:flight_ctrl_noise}).
Next, Section \ref{sec:s2_and_flight_by_flight_variations} introduces the unusual magnetic biases our UAV injects into our measurements and how distancing the magnetometer from the electronics improves the consistency of measurements. 
We then show that our compromise map (which uses $n_1 = 511$ observations for inference) yields estimates within 0.013$\mu$T of our intermediate map (which uses $n_2 = 2001$ observations for inference) in Section \ref{sec:accuracy_of_compromise_map}.

We follow with an analysis of the spatial density of observation points used to train the GPR map in Section \ref{sec:compromise_map_spatial_density_analysis} and find that using observations within 0.55m of one another is sufficient to accurately represent the magnetic field in our flight arena.
Section \ref{sec:3_3_vs_3_1_norm_map} shows that it is equivalent to either create a specialized map on the norm of the field or simply take the norm of estimates from a vector-valued magnetic field map.
Finally, Section \ref{sec:consistency_check} presents our consistency metric which is a tool to identify when a user can rely on the predictions from their GPR-based magnetic field map.

\subsection{Flight Controller Creating Magnetic Field Noise}
\label{sec:flight_ctrl_noise}
Our vehicles use the \texttt{rc\_pilot\_a2sys} autopilot originally forked from the open-source \texttt{rc\_pilot}
repository by James Strawson and Librobotcontrol (\url{https://github.com/StrawsonDesign/rc_pilot}).
\texttt{rc\_pilot\_a2sys} uses a four-stage cascaded PID controller as explained in Section IV.B.1 of \cite{Romano2022}. Our work started with the gains used in Ref. \cite{Romano2022} but adjusted them to reduce noise from the motors and ESCs. 

\begin{figure}
 \begin{subfigure}[b]{0.48\textwidth}
  \includegraphics[width=\textwidth,trim= 20 170 30 170, clip]{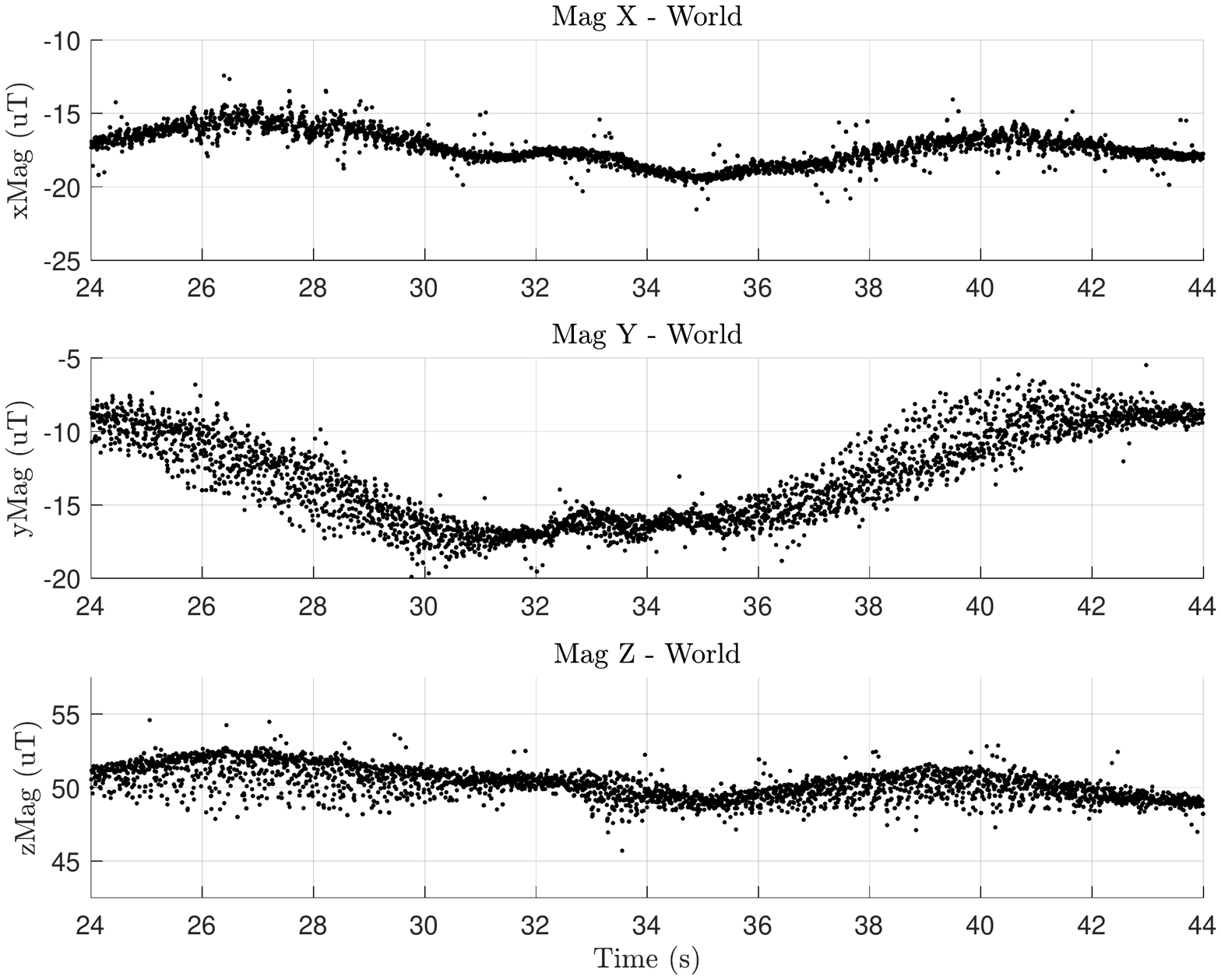}
  \caption{Test t5\_00: Flight with ``noisy'' gains.}
  \label{fig:noisy_maggie}
 \end{subfigure}
 \hfill
 \begin{subfigure}[b]{0.48\textwidth}
  \includegraphics[width=\textwidth,trim= 20 170 30 170, clip]{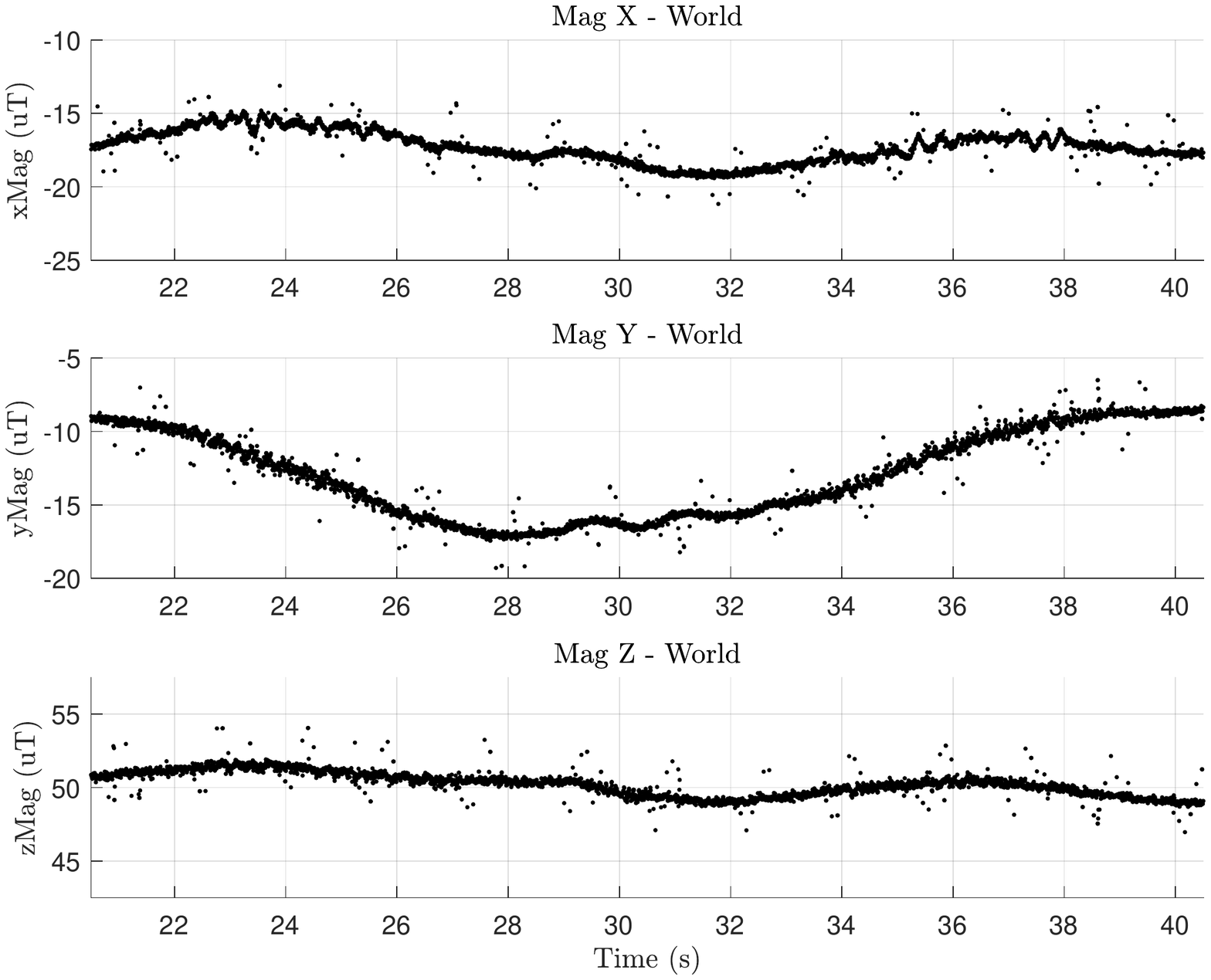}
  \caption{Test t5\_01: Flight with ``quiet'' gains.}
  \label{fig:quiet_maggie}
 \end{subfigure}
 \caption{Vehicle ``Maggie'' flying two instances of the same flight trajectory showing how the flight controller can inadvertently create noise in the measured magnetic field.}
	\label{fig:ctrl_and_mag}
\end{figure}

Figure \ref{fig:ctrl_and_mag} shows magnetometer data from a segment of an experiment where the flight vehicle (not Q1, but another vehicle of the same construction) was commanded to a single-altitude scanning pattern (Figure \ref{fig:1_5m_traj}) at 1.5m altitude. 
The data in Figure \ref{fig:ctrl_and_mag} was rotated from the vehicle's body frame to the world frame using attitude estimates from motion capture cameras.

Figure \ref{fig:noisy_maggie} (t5\_00) is the magnetometer data with the gains inherited from ref. \cite{Romano2022}. Here, the drone moves along the $x$ axis (4m stride) of the flight space from 24s - 30s, and again from about 36s - 42s. During these time segments, there is a large variance in the measured magnetic field due to commands from the flight controller to the motors and ESCs.

Figure \ref{fig:quiet_maggie} (t5\_01) shows another instance of the same trajectory, but with new controller gains. The plots are temporally aligned to easily compare corresponding flight segments. At a glance, it is clear that the variance in the magnetic field measurements is much lower in Figure \ref{fig:quiet_maggie} than in Figure \ref{fig:noisy_maggie}, but not completely gone. This is most evident in the ``Mag Y - World'' plot of Figure \ref{fig:quiet_maggie} where the drone moves along the $x$ axis (4m stride) of the flight space from 21s-26s and again from 34s - 40s.

As explained in Section IV.B.1 of ref. \cite{Romano2022}, the controller used on this vehicle is a four-stage cascaded PID controller with an outer loop that operates on position error and an inner-most loop controlling angular rate error. 
The math for the inner loop is shown below in Equation \ref{eq:att_rate_pid_ctrlr}
\begin{equation}
    \label{eq:att_rate_pid_ctrlr}
    \begin{split}
       \bm{\tau}_{\phi} &= K_{4,\phi}^{P} \, (\dot{\phi}_d - \; \tilde{\omega}_x) \\
                        &+ K_{4,\phi}^{I} \, \int_{0}^{t} (\dot{\phi}_d - \; \tilde{\omega}_x) \, ds \\ 
                        &+ K_{4,\phi}^{D} \, \frac{d}{dt} (\dot{\phi}_d - \; \tilde{\omega}_x)\\
    \end{split},
\end{equation}

\noindent where $\bm{\tau}_{\phi}$ is the desired torque for roll $(\phi)$, $\{K_{4,\phi}^{P}, K_{4,\phi}^{I}, K_{4,\phi}^{D}\}$ are the P, I, and D gains for the fourth-stage controller along the roll axis, $\dot{\phi}_d$ is the desired roll rate, and $\tilde{\omega}_x$ is the measured roll rate from the gyroscope. There are similar fourth-stage PID loops for pitch ($\dot{\theta}$) and yaw ($\dot{\psi}$) rates respectively.

The `noisy' gains, inherited from \cite{Romano2022}, used $K_{4,\phi}^{D} = K_{4,\theta}^{D} = K_{4,\psi}^{D} = 0.01$ while the modified `quiet' gains set all three of these values to 0. This change alone is responsible for the reduction in motor-induced noise in Figure \ref{fig:ctrl_and_mag}.

\begin{figure}
    \centering
  \captionsetup{justification=centering}
  \includegraphics[width= 0.7\textwidth,trim= 10 160 30 170, clip]{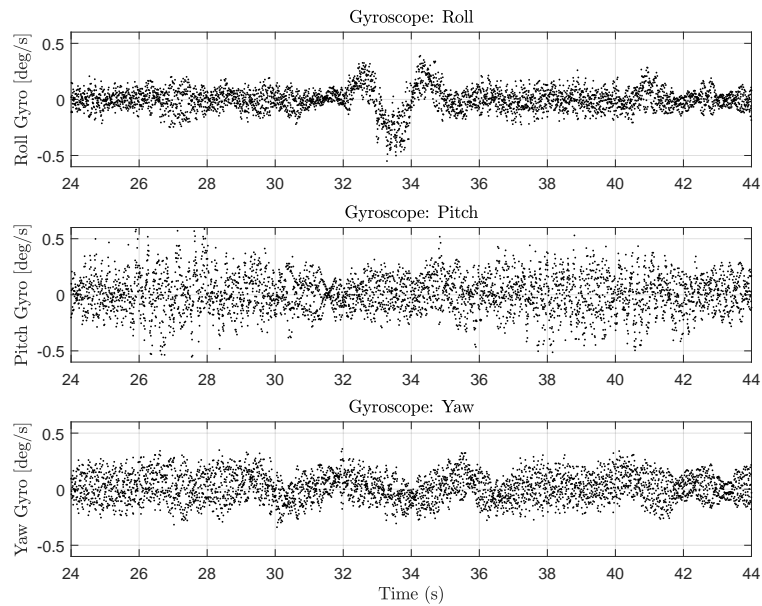}
  \caption[Raw gyroscope data from t5\_00.]{Raw gyroscope data from t5\_00 . Vibrations from UAV propellers create noisy gyroscope data. Differentiating this noisy data creates the high-frequency magnetic noise seen in Figure \ref{fig:noisy_maggie}.
  }
 \label{fig:noisy_gyro}
\end{figure}

The derivative term of the attitude rate controllers (e.g., Equation \ref{eq:att_rate_pid_ctrlr} for roll rate) uses numerical derivatives of gyroscope measurements (Figure \ref{fig:noisy_gyro}) as part of their computation.
As the quadrotor flies, the propellers induce vibrations that make the gyroscope (and accelerometer) measurements rather noisy. 
Taking numerical derivatives of these noisy gyroscope measurements causes the motors to be commanded with high-frequency inputs that induce measurable electromagnetic noise. 
We believe this problem is amplified as the total amount of current pulled from the 4S batteries increases.
This would explain why the magnetometer variance is larger during the 4m, $x$-axis stride (with a max commanded velocity of 0.75m/s for this set of tests) than during the 0.25m, $y$-axis stride (max commanded velocity of 0.47m/s).

Aside from changing controller gains, other possible solutions include commanding less-aggressive maneuvers during flight tests, moving the magnetometer further from the motors and ESCs (Section \ref{sec:s2_and_flight_by_flight_variations}), or time-varying magnetometer calibration \cite{Springmann2012_magCalibration}.
In addition, from some preliminary data gathered when trying to understand this problem, we believe that differences in motors and ESCs (even those of the same make and model) may create different magnitudes of measurable magnetic noise.


The remainder of this paper uses only the `quiet' controller gains.
Note that the analysis for this section used $S_2$ = 2cm (Figure \ref{fig:q1_vertical_dist}). In the following section, we analyze how the distance of the magnetometer from the motors and ESCs affects the \textit{variation} in the measured magnetic field.

\subsection{Varying \texorpdfstring{$S_2$}{TEXT}: Distance from Magnetometer to Electronics}
\label{sec:s2_and_flight_by_flight_variations}
Though the flight controller created motor-induced, high-frequency magnetic field noise, there is another measurable magnetic anomaly at play with our quadrotors that causes flight-by-flight \textit{variation} in the measured magnetic field.
Here, we distinguish variance (the spread of a distribution of points) from variation (more of an offset).
We will reserve variance to describe the spread of points around a probabilistic mean. 
For this, removing the derivative gains from a particular PID loop of our flight controller reduced the variance of motor-induced magnetic noise (Figure \ref{fig:ctrl_and_mag}). To avoid ambiguity, variation will be used to describe the differences in the magnetic field between subsequent flights of the same trajectory. As we will show in this section, the differences in measured magnetic field values are not obviously spread around some average signal. 

Recall that $S_2$ (Figure \ref{fig:q1_vertical_dist}) is the distance from the top of the BeagleBone Blue (BBB) to the bottom of the magnetometer's mounting plate. In this section, we will vary the parameter $S_2$ to show that increasing the distance of the magnetometer from the other electronics reduces the flight-by-flight variations we see in the measured field. It is easier to explain these anomalies graphically, which we will do in a moment, but first, we explain our testing methodology. 

\subsubsection{Testing Methodology}
Q1 was commanded to fly the same single-altitude scanning trajectory (Figure \ref{fig:1_5m_traj}) at an altitude of 1.5m. Initially, in our investigation of these flight-by-flight variations, we believed that each 4S LiPo battery could have a different amount of ``resting current'' it provided to the motors to achieve the same flight maneuver as another 4S battery. Though we could not conclusively prove or disprove this hypothesis with our data, it did drive the design of our experiments. For this, each battery started at full charge and Q1 flew the same 1.5m scanning trajectory several times in a row. This allowed us to see if the flight-by-flight variations were due to the battery's voltage during a flight. After a few repetitions with the first 4S battery, a new (fully charged) 4S battery was used to power Q1 as it flew more repetitions of the same trajectory. 
By testing different batteries, we could see if the cause of the flight-by-flight anomalies was tied to which battery was powering the motors and ESCs. 


Three Turnigy 4S 2650mAh 20C batteries (denoted as \#02, \#04, and \#14) were used for this analysis. The charge/discharge and storage history of these batteries are largely unknown as they have been used by multiple members in the lab since they were purchased in May of 2021. 
The only thing we could rigorously control for is fully charging each battery to 16.8V just before a flight test. 

For this analysis, each flight was downsampled to 10Hz or 2Hz to reduce spatially redundant observations resulting in $\sim$900 or $\sim$180 observations respectively for each flight. Figure \ref{fig:1_5m_traj} shows the observation set from one flight with 10Hz and 2Hz downsampling. Although 10Hz clearly shows better coverage of the flight space, the RMSE values for this analysis changed by at most 0.03$\mu$T on the $S_2$=8cm dataset when training on 10Hz vs 2Hz downsampling sets. 
Thus, we sometimes use 10Hz downsampling to illustrate specific points but typically use 2Hz downsampling for faster training and validation of the magnetic field maps. 

With this gathered data, we can train a GPR-based map of the magnetic field at a 1.5m single-altitude slice of the working volume and compare the predictions of that single-altitude map to the measurements gathered from the other repetitions of the same trajectory. 
Since we assume the ambient magnetic field is not changing throughout our experiments, we expect the magnetic field measurements to be nearly identical between each flight.

\subsubsection{Flight-by-flight variations}
\label{sec:flight_by_flight_variations}
\begin{figure*}[ht]
 \begin{subfigure}[b]{0.48\textwidth}
  
  \includegraphics[width=\textwidth,trim= 10 160 20 130, clip]{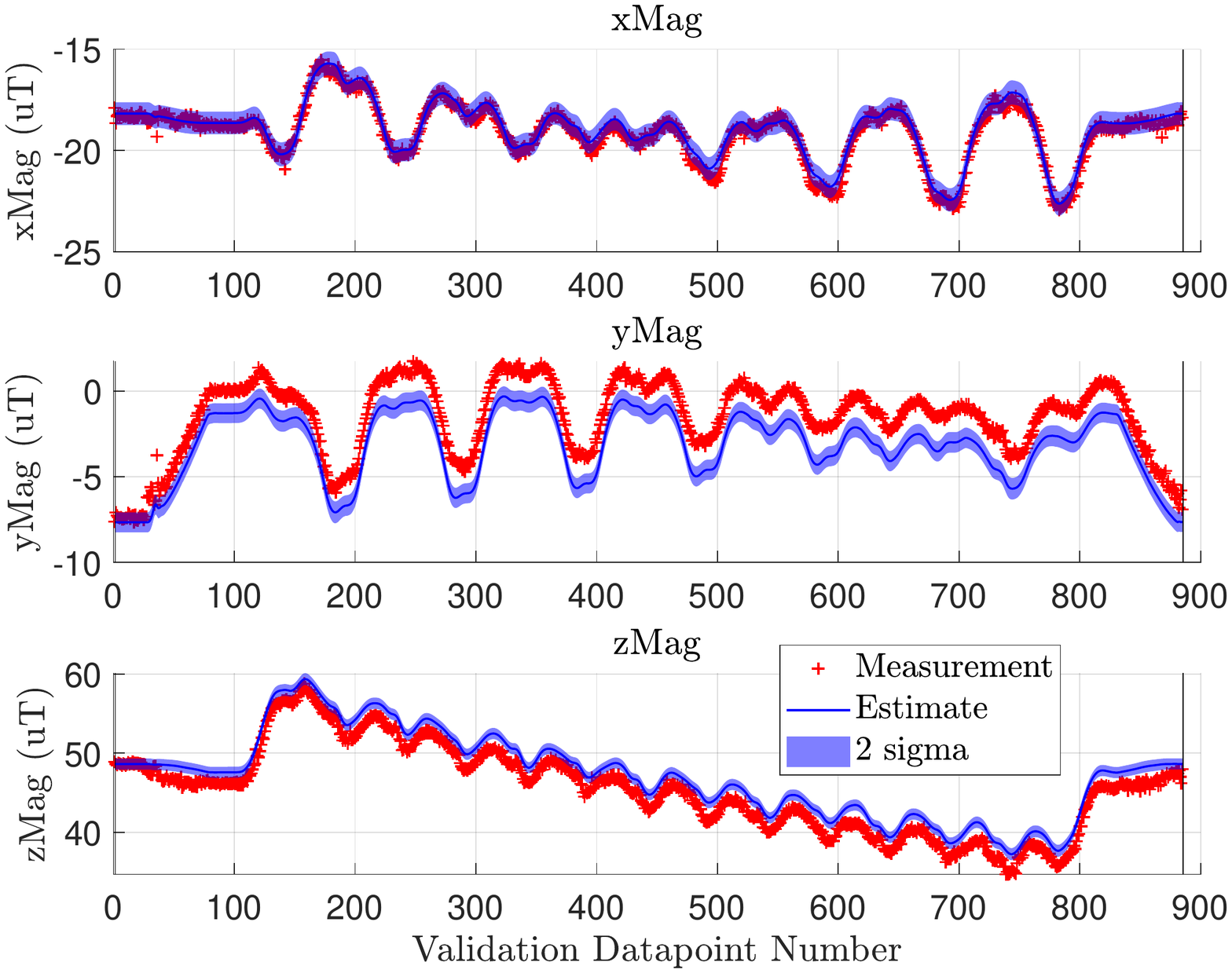}
  \caption{$S_2$=2cm. Trained on t1\_09 (\#04). Validated on t1\_05 (\#14). Same data as Figure \ref{fig:s2_gprErr_2}}
  \label{fig:s2_gprErr_1}
 \end{subfigure}
 \hfill
 \begin{subfigure}[b]{0.48\textwidth}
  
  \includegraphics[width=\textwidth,trim= 20 160 20 130, clip]{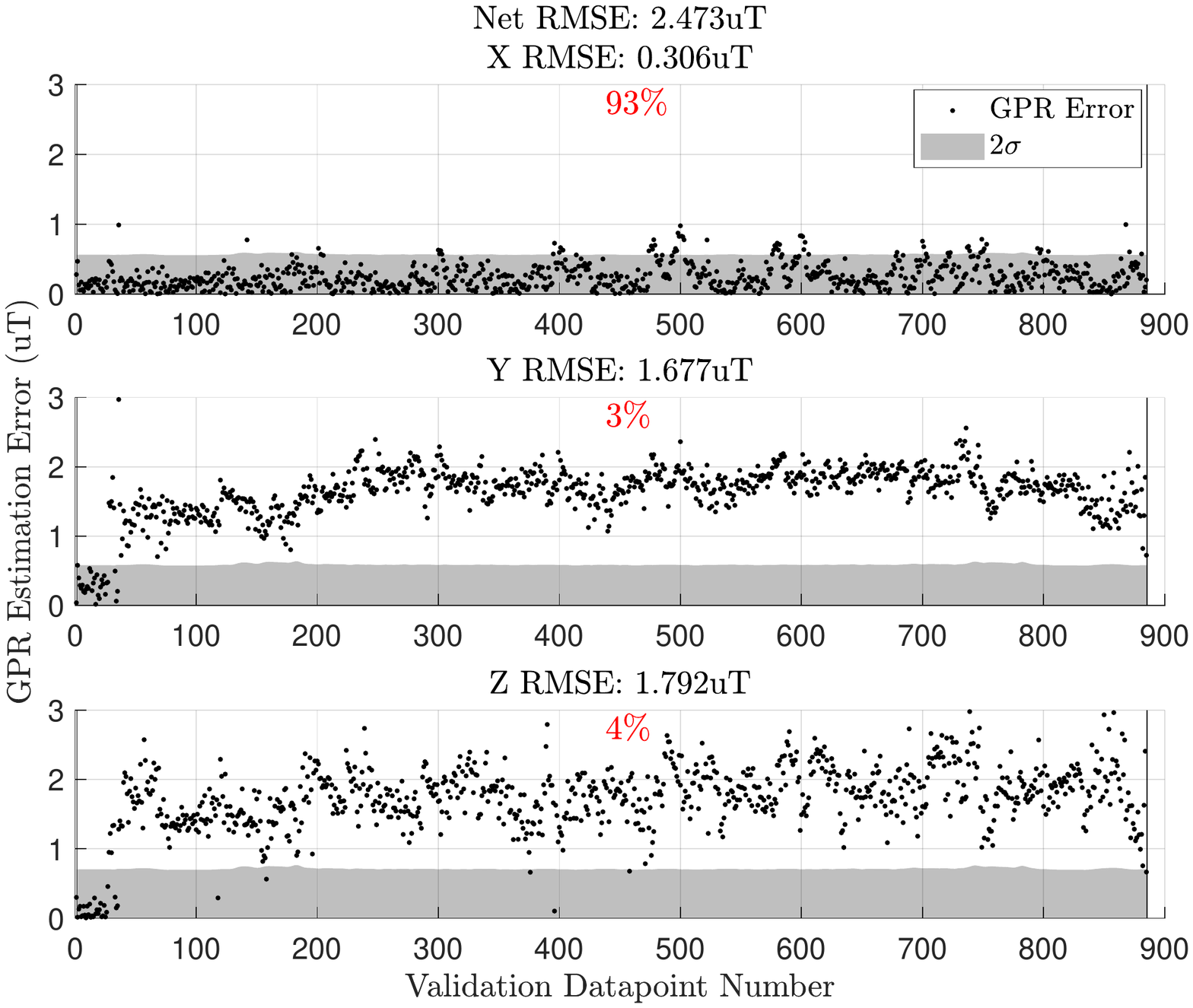}
  \caption{$S_2$=2cm. Trained on t1\_09(\#04). Validated on t1\_05(\#14). Same data as Figure \ref{fig:s2_gprErr_1}}
  \label{fig:s2_gprErr_2}
 \end{subfigure}
 \begin{subfigure}[b]{0.48\textwidth}
  
  \includegraphics[width=\textwidth,trim= 20 160 20 130, clip]{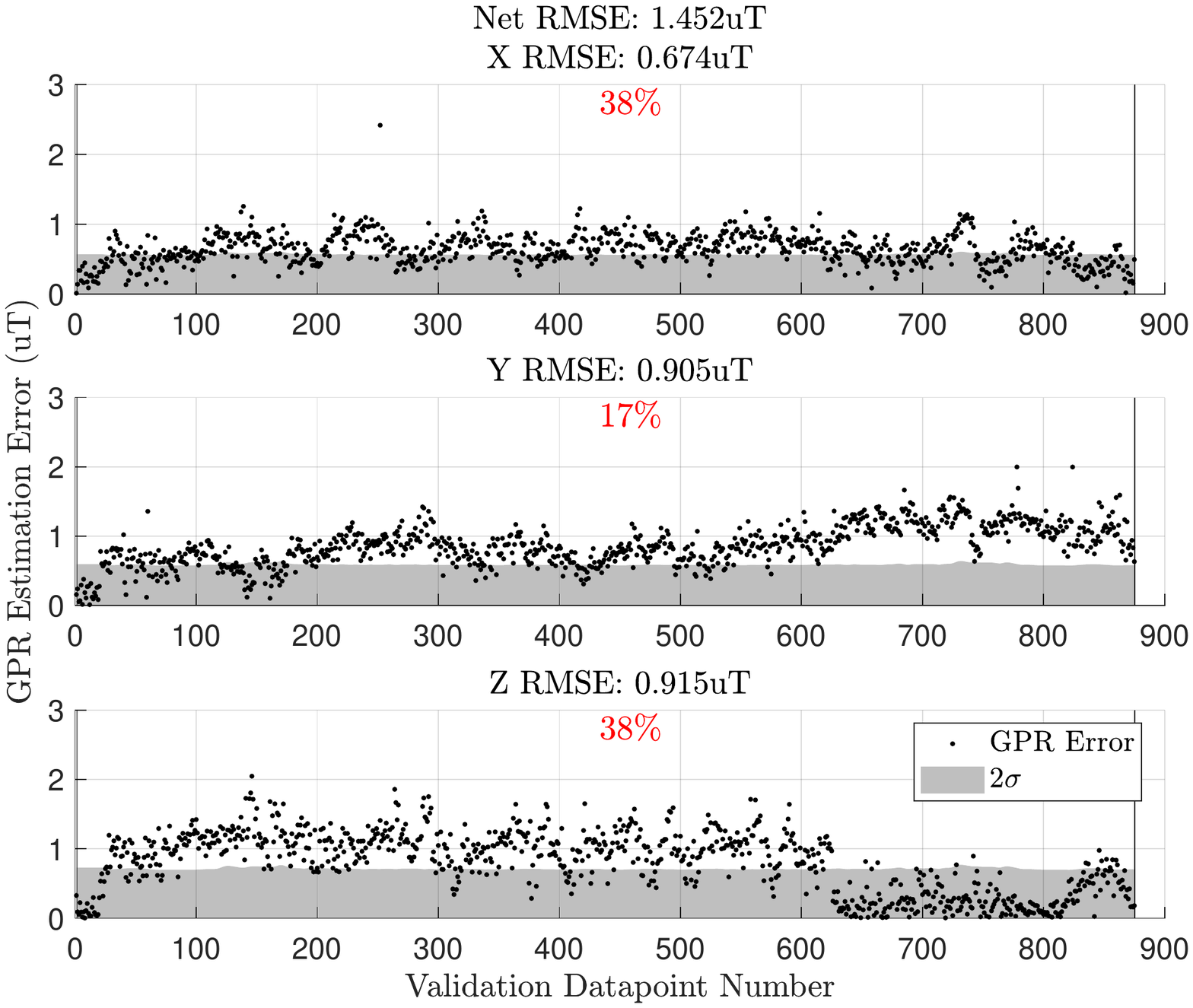}
  \caption{$S_2$=2cm. Trained on t1\_09(\#04). Validated on t1\_04(\#14).}
  \label{fig:s2_gprErr_3}
 \end{subfigure}
 \hfill
 \begin{subfigure}[b]{0.48\textwidth}
  
  \includegraphics[width=\textwidth,trim= 20 160 20 130, clip]{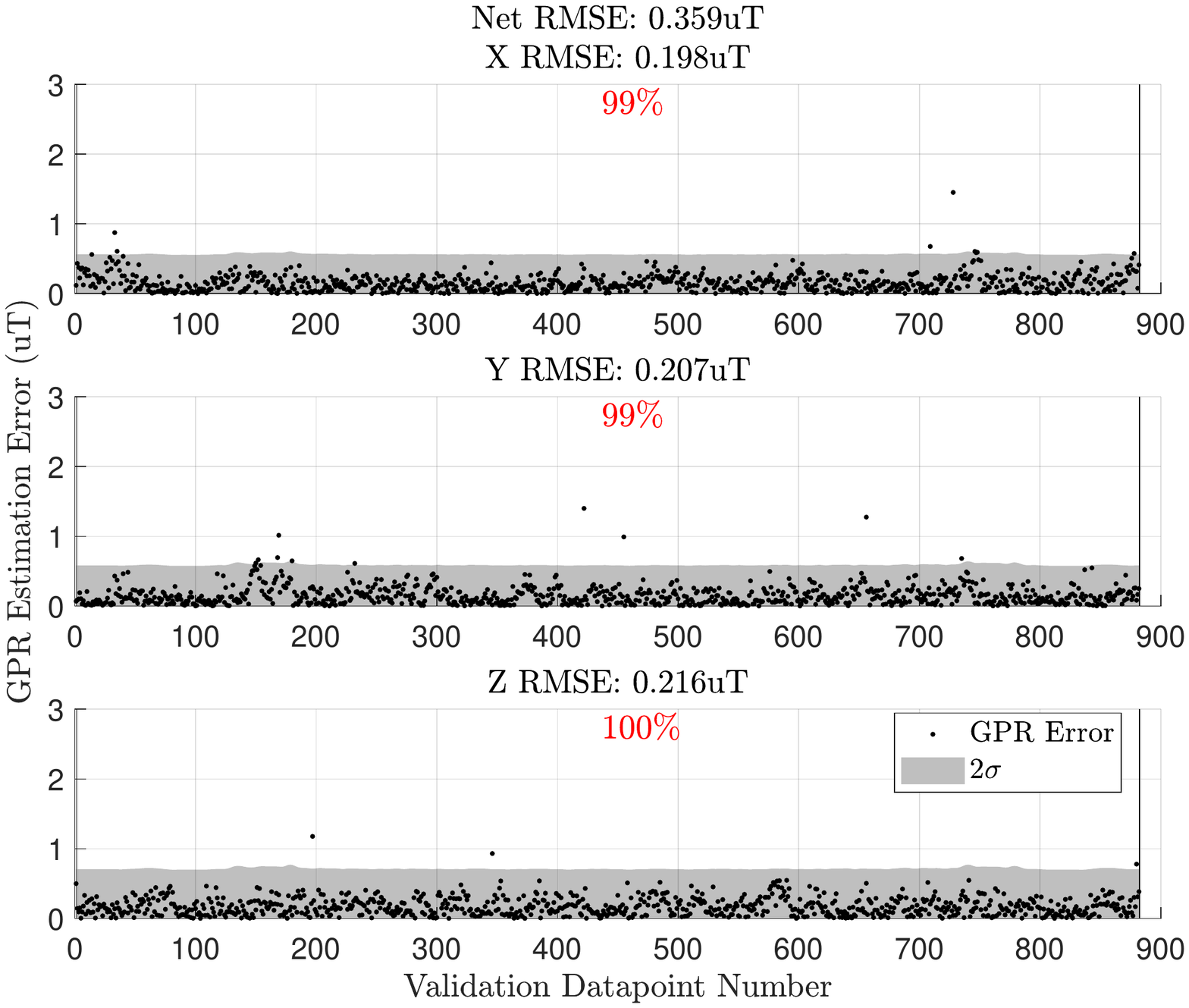}
  \caption{$S_2$=2cm. Trained on t1\_09(\#04). Validated on t1\_15(\#02).}
  \label{fig:s2_gprErr_4}
 \end{subfigure}
 
\caption{Flight-by-flight variations shown through three validation tests. 
}
	\label{fig:s2_gprErr}
\end{figure*}

By training the map on a single repetition and validating on another, we get data that looks like Figure \ref{fig:s2_gprErr} which has two types of plots: one with red and blue lines with blue shading and another type of plot in grayscale. 

For the first type (e.g., Figure \ref{fig:s2_gprErr_1}), each actual magnetometer measurement (in the world frame) from the \textit{validation} dataset is shown as a red cross. The blue line is the GPR's predicted mean at the location the drone was during that timestamp in the validation flight. 
The blue shading around the solid blue line depicts the two standard deviations ($2\sigma$) of the GPR's uncertainty with its prediction. 

The second type of plot (e.g., Figure \ref{fig:s2_gprErr_2}) depicts a black dot as the error between the red cross and blue line while the GPR's $2\sigma$ uncertainty is the gray shading. 
Here, we plot the absolute value of the error since the sign of the error does not indicate anything of interest. 
The percentage in the grayscale plots depicts how often the black dots (GPR error) are within the gray shading (GPR's $2\sigma$ uncertainty). 

In Figure \ref{fig:s2_gprErr}, the GPR map is trained on 2Hz-downsampled observations from a single flight (t1\_09) and validated against 10Hz-downsampled observations from three flight tests from the same t1 flight series. These three validation examples are generally representative of the different flight-by-flight magnetic field anomalies we have seen. Figure \ref{fig:s2_gprErr_2} shows a relatively large steady-state prediction error throughout, Figure \ref{fig:s2_gprErr_3} has a steady-state error that changes suddenly partway through the flight, while Figure \ref{fig:s2_gprErr_4} is an ideal case where the mapping and validation flights have agreeable measurements.

This is why we refer to this anomaly as ``flight-by-flight \textit{variation}''. The type of GPR error we see here across consecutive flights in the same test segment (e.g., t1\_XX) does not seem to be cleanly distributed around some mean bias value. Instead, repetition t1\_09 had similar magnetic observations as t1\_15 (Figure \ref{fig:s2_gprErr_4}) yet starkly different observations than t1\_05 (Figure \ref{fig:s2_gprErr_2}). Further complicating the matter are cases like t1\_04 (Figure \ref{fig:s2_gprErr_3}) with a time-varying bias. 

There is evidence in Figure \ref{fig:s2_gprErr} that suggests these magnetic anomalies are caused by the ESCs and motors. 
The initial black dots in Figures \ref{fig:s2_gprErr_2} and \ref{fig:s2_gprErr_3} have low errors before the anomalous bias takes effect. 
These beginning points are observations gathered when the drone is at rest on the ground (before it has taken off) and the bias sets in, if at all when the quadrotor is in the air. 
This suggests that before the motors and propellers are spinning, the observations from t1\_05 and t1\_04 are similar to those from t1\_09. 

This insight begs the obvious experiment of constraining the motion of the quadrotor and sampling the magnetometer with and without the propellers spinning. Unfortunately, we could not safely conduct such a test with our current setup, but we plan to test this in the future. 

\begin{figure}
    \centering
  \captionsetup{justification=centering}
  \includegraphics[width= 0.7\textwidth,trim= 0 170 30 180, clip]{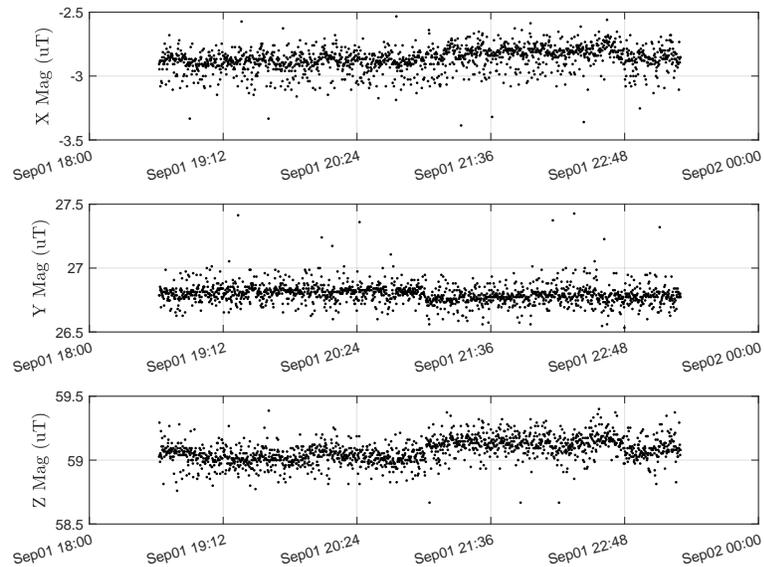}
  \caption{Stationary RM3100 data sampled at 0.1Hz over 3+ hours during t6\_XX flight tests. $x$, $y$ or $z$ components of the magnetic field never change by more than 0.15$\mu T$ respectively.}
 \label{fig:long_duration_rm3100}
\end{figure}

Finally, it is possible that the ambient magnetic field in our flight arena is changing over time.
To check this, placed a stationary RM3100 in our workspace for 3+ hours sampling data once every 10 seconds (0.1Hz).
The raw data from this experiment is shown in Figure \ref{fig:long_duration_rm3100}.
Here, we still have regular, spurious measurements (like in Figure \ref{fig:stationary_rm3100_data}) but the main signal does not change by more than 0.15$\mu T$ within each respective component of the measured magnetic field. 
This is much smaller than the 1.8$\mu T$ of variation we see in the Z component of Figure \ref{fig:s2_gprErr_2}.
Thus, we believe the flight-by-flight variations are due primarily to UAV-induced noise and not by time-varying changes in the ambient magnetic field.

\subsubsection{LiPo Batteries and Magnetic Variation}
From the three cases in Figure \ref{fig:s2_gprErr}, it is tempting to conclude that differences in batteries are the cause of these flight-by-flight anomalies. Here, we see that batteries \#04 (used for t1\_09) and \#02 (used for t1\_15) yield agreeable observations. However, from Figures \ref{fig:s2_gprErr_2} and \ref{fig:s2_gprErr_3}, it seems that battery \#14 (used for t1\_05 and t1\_04) is disagreeable with the other two. 

To investigate this idea, we flew 60 repetitions of the 1.5m-scanning trajectory from Figure \ref{fig:1_5m_traj} over four test sessions (t1\_XX, t2\_XX, t3\_XX, and t4\_XX) each corresponding to a different value of $S_2$ (2cm, 4cm, 6cm, and 8cm). For these tests, we used the same three batteries (\#02, \#04, and \#14) and flew consecutive repetitions of the scanning trajectory for each battery. 
For each test session, we trained the GPR map on the \textit{first} battery \#04 flight and validated the map on all flight tests of the same test session. 
Note that the data in Figure \ref{fig:s2_gprErr} (tests t1\_04, t1\_05, t1\_09, and t1\_15) are four of the $S_2$=2cm repetitions used in this larger analysis.

\begin{figure}
    \centering
  \captionsetup{justification=centering}
  \includegraphics[width= 0.7\textwidth,trim= 60 200 100 230, clip]{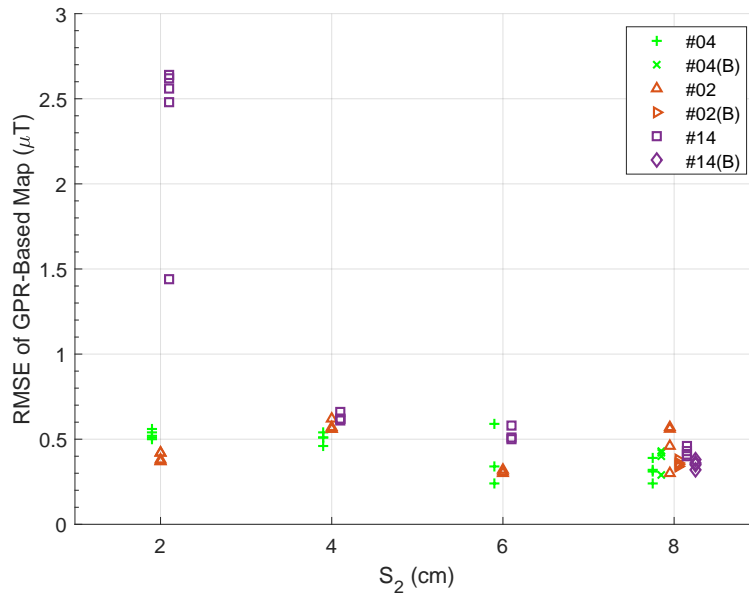}
  \caption{$S_2$ versus vector GPR RMSE. At each $S_2$, the map is trained on \textit{one} flight from battery \#04 (green +). For $S_2$=\{2,4,6,8\}cm there are $N$=\{5,4,3,8\} reps per battery.}
 \label{fig:mag_and_s2}
\end{figure}

The results of this 60-flight analysis are summarized as a scatter plot in Figure \ref{fig:mag_and_s2}. The horizontal position of each point is meant to signify the value of $S_2$ as 2cm, 4cm, 6cm, or 8cm. Any horizontal deviations from these discrete values are only for visual clarity and do not reflect any deviation in the actual value of $S_2$ when the data was gathered. 
The vertical position is the vector RMSE of all three GPRs (Equation \ref{eq:rmse_norm}). Finally, the colors and symbols identify which battery was used for each repetition with a green $+$ for battery \#04, an orange $\triangle$ for \#02, and purple $\square$ for \#14. 

At $S_2$ = \{2, 4, 6, 8\}cm, each battery flew $N$ = \{5, 4, 3, 8\} repetitions of the 1.5m scanning trajectory. For $S_2$ = 8cm, each battery started at full charge, flew four consecutive reps, was re-charged, then flew four more repetitions. We present the second charge of each battery in the $S_2$=8cm dataset with a rotated $+$. $\triangle$, and $\square$ respectively.

Figure \ref{fig:mag_and_s2} shows two things. First, increasing $S_2$ makes the observations of each flight more consistent with one another. Since the GPR-based map (for each $S_2$) is trained on a battery \#04 flight, the scatter plot really shows us the magnitude of the flight-by-flight variations relative to an arbitrary battery \#04 flight. As we increase $S_2$, the variations tend towards a difference of about 0.3$\mu$T to 0.5$\mu$T. 

Note that for each value $S_2$, there is a single green $+$ that is validated on the same observations used to train the GPR map. Even so, we do not see \textit{any} of the validation tests going much below 0.3$\mu$T meaning this may be something of a lower limit on how accurate our GPR maps can be with our testing platform which would include inaccuracies due to sensor noise (a function of our 200Hz sampling rate) and any electromagnetic noise the quadrotor generates.

The second realization is that, unlike what the three examples in Figure \ref{fig:s2_gprErr} may suggest, there is not a clear relationship between battery choice and flight-by-flight variation. Although battery \#14 (purple $\square$) tends to be most disparate from battery \#04 at $S_2$ of 2cm, 4cm, and 6cm, we see its observations fall in line with the rest by $S_2$ = 8cm. 

We take this moment to point out a flawed conclusion we stated in our previous work.
In Figure 10b of \cite{Kuevor2021},
observations from a rectangular trajectory (the first $\sim$500 points) are validated on a GPR-based map. 
In that plot, there is about 2$\mu$T of an error on each of the $x$ and $z$ GPR validation sets except before takeoff and after landing. 
In \cite{Kuevor2021}, we hypothesized that this error in $x$ and $z$ was caused by a pitching misalignment in the magnetometer for that single flight (relative to the pitch angle for all the other flights). 
However, if this were the case, there likely would not be the brief moments of agreement before takeoff and after landing as shown in Figure 10b of \cite{Kuevor2021}. 
With what we have learned in this work, we now believe this was actually an example of these flight-by-flight variations. 

For the work in ref. \cite{Kuevor2021}, these anomalies may not have happened as often due to us using a different instance of the M330 quadrotor. 
Alternatively, it could be that the nature of attitude estimation (the focus of \cite{Kuevor2021}) is fairly robust to 2-3$\mu$T variations between flights if such anomalies do not significantly change the \textit{angle} of the ambient magnetic field (which has a magnitude of 40-70$\mu$T in our workspace). 
Thus, it is likely that these variations were simply less salient in ref. \cite{Kuevor2021} given our previous application of attitude estimation.

We believe that time-varying magnetometer calibration could address the flight-by-flight variation issues presented here. In ref. \cite{Springmann2012_magCalibration}, they use measurements of the electric current near high-powered devices to estimate and offset vehicle-induced magnetic fields from their measurements. 
However, our vehicle does not have \textit{any} electric current sensors, so such a time-varying calibration solution is out of reach for our platform. 
Instead, we address the problem with our ``compromise'' map.

\subsection{Accuracy of Compromise Map}
\label{sec:accuracy_of_compromise_map}

In this section, we propose a solution to the flight-by-flight variations that incorporate data from several training flights into a single magnetic field map. Since each flight has a chance of giving a different bias in the measured magnetic field, the GPR-based map will overfit if trained on only a single flight test. Thus, we instead, use $n_2$ observations from many flights that span the working volume to create an ``intermediate map'' that optimizes hyperparameters $\{ \bm{\Theta}^*_x(D_x^{n_2}), \; \bm{\Theta}^*_y(D_y^{n_2}), \; \bm{\Theta}^*_z(D_z^{n_2}) \}$ \textit{and predicts with} the $n_2$ observations.
The compromise map then uses the intermediate map's estimates at $n_1$ user-selected locations to predict the magnetic field using only $n_1$ points. The goal of this section is to see how the accuracy of the compromise map varies as a function of $n_1$; specifically, taking after ref. \cite{Akai2017}, to see how the spatial density of the $n_1$ training points effects map accuracy.

\subsubsection{Multi-Altitude Trajectories}
\label{sec:multi_altitude_trajectories}
This section uses data from eight different flight tests listed in Table \ref{tbl:train_vali_uti_flights} where the first four flights are typically used for \textit{training} the intermediate/compromise map while the remaining four are used for \textit{validation}.
The trajectories listed in Table \ref{tbl:train_vali_uti_flights} are all multi-altitude tests where each single-altitude slice is the same trajectory from Figure \ref{fig:1_5m_traj}. 
``Lower Four alts.'' flies at altitudes $Z = [$-0.5, -0.75, -1.0, -1.25]m, ``Upper Four alts.'' at $Z = [$-1.5, -1.75, -2.0, -2.25]m, ``Scan-$\gamma$'' at $Z = [$-0.5, -1.375, -2.25]m, while ``Scan-$\epsilon$'' and ``Scan-$\epsilon^\perp$'' fly at $Z = [$-0.75, -1.5, -2.0]m. 
The idea here is to gather redundant observations throughout the working volume to help our GPR map learn the flight-by-flight variations.

Additionally, Table \ref{tbl:train_vali_uti_flights} lists the number of 2Hz downsampled observations from each flight. The number of 4Hz and 10Hz observations are \textit{approximately} $2\times$ and $5\times$ the values listed in Table \ref{tbl:train_vali_uti_flights}.

\begin{table}
 \centering
 \caption{Separate flights are used for \textit{training} and \textit{validation}. All magnetic field observations were downsampled to 2Hz for \textit{training} and \textit{validation}. Number of observations each training flight contributes to the map is listed. } 
 \label{tbl:train_vali_uti_flights}
 \begin{tabular}{l l c c} 
\toprule
  \textbf{Flight} & \textbf{Flight} & \textbf{Training [2 Hz]} & \textbf{Validation [2 Hz]} \\ 
  \textbf{Number} & \textbf{Description} & (\# Observations) & (\# Observations)\\
 \midrule
 t6\_00 & Lower Four Alts. & 571 & -- \\
 t6\_01 & Upper Four Alts. & 580 & -- \\
 t6\_03 & Scan-$\gamma$ & 442 & -- \\
 t6\_21 & Scan-$\epsilon^\perp$ & 408 & -- \\
 \midrule
 t6\_04 & Lower Four Alts. & -- & 573 \\
 t6\_05 & Upper Four Alts. & -- & 573 \\
 t6\_06 & Scan-$\gamma$ & -- & 441 \\
 t6\_20 & Scan-$\epsilon$ & -- & 384 \\
\bottomrule
 \end{tabular}
\end{table}

\begin{table}
 \centering
 \caption{Multi-altitude map error. Trained on $n_2 = 408$ observations from flight test t6\_21 (Scan-$\epsilon^\perp$).}
 \begin{tabular}{l || c | c c c || r r r } 
\toprule
  Flight ID and & 
  \multicolumn{4}{c ||} {GPR RMSE ($\mu$T)} & 
  \multicolumn{3}{c }   {Error within 2$\sigma$ (\%) }  \\
  
  Description 
  & \multicolumn{4}{c ||}{\textit{Norm|X|Y|Z} } & 
    \multicolumn{3}{c }{\textit{X|Y|Z} } \\

 \midrule

t6\_04 (Lower Alts.) & 0.595 & 0.256 & 0.284 & 0.456 & 74.69 & 87.09 & 52.88 \\
t6\_05 (Upper Alts.) & 0.998 & 0.633 & 0.541 & 0.551 & 6.81 & 19.72 & 32.11 \\
t6\_06 (Scan-$\gamma$) & 0.642 & 0.362 & 0.318 & 0.424 & 56.01 & 86.62 & 68.25 \\
t6\_20 (Scan-$\epsilon$) & 0.556 & 0.229 & 0.293 & 0.413 & 56.25 & 47.40 & 39.06
 \end{tabular}
\label{tbl:t21_only}
\end{table}

\begin{table}
 \centering
 \caption{GPR error for multi-altitude \textit{intermediate} map trained on $n_2=2001$ observations from tests t6\_00, t6\_01, t6\_03, and t6\_21.}
 \begin{tabular}{l || c | c c c || r r r } 
\toprule
  Flight ID and & 
  \multicolumn{4}{c ||} {GPR RMSE ($\mu$T)} & 
  \multicolumn{3}{c }   {Error within 2$\sigma$ (\%) }  \\
  
  Description 
  & \multicolumn{4}{c ||}{\textit{Norm|X|Y|Z} } & 
    \multicolumn{3}{c }{\textit{X|Y|Z} } \\

 \midrule

t6\_04 (Lower Alts.)   & 0.466 & 0.144 & 0.145 & 0.419 & 98.43 & 98.08 & 85.17 \\
t6\_05 (Upper Alts.)   & 0.713 & 0.403 & 0.447 & 0.382 & 45.55 & 23.73 & 90.05 \\
t6\_06 (Scan-$\gamma$)      & 0.378 & 0.184 & 0.155 & 0.292 & 97.96 & 96.83 & 96.83 \\
t6\_20 (Scan-$\epsilon$)    & 0.406 & 0.111 & 0.222 & 0.321 & 99.48 & 92.97 & 93.75 \\
 \end{tabular}
\label{tbl:t0_t1_t3_t21}
\end{table}

\subsubsection{Intermediate Map Accuracy}
To start, we demonstrate the value of adding observations from many flights to a GPR-based map. We call GPR maps that are trained on observations from multiple flights an \textit{intermediate map}. 

Table \ref{tbl:t21_only} gives the performance of a multi-altitude magnetic field map trained on $n_2 = 408$ (2Hz downsampling) observations from flight test t6\_21 and validated on observations from four different flight tests. The table is split into three major columns that give the details of the validation flight and the vector RMSE ($\mu$T) of all three GPRs along with the RMSE of each $x$, $y$, and $z$ GPR respectively. The third major column quantifies how often (as a percentage) an error data point lies within the 2$\sigma$ uncertainty of the respective GPR. The performance metrics from the second and third major columns are the same as those listed in the grayscale plots of Figure \ref{fig:s2_gprErr} (with different training and validation flights in this section).

The underlying issue is the map will overfit to the observations gathered from t6\_21.
In Section \ref{sec:s2_and_flight_by_flight_variations}, we showed that there are flight-by-flight variations in the measured magnetic field. By training a GPR-based magnetic field map on a single flight, we prevent the map from learning to account for these flight-by-flight variations. 
Note that training the map on $n_2 = 2038$ observations (10Hz downsampling) from t6\_21 gives norm RMSE values of 0.629$\mu$T, 1.034$\mu$T, 0.751$\mu$T, and 0.552$\mu$T when validating on t6\_04, t6\_05, t6\_06, and t6\_20 respectively. 
Thus, simply increasing the number of training samples from a single flight test does not improve the map's performance.

By comparison, Table \ref{tbl:t0_t1_t3_t21} trains the GPR hyperparameters (and performs inference) on $n_2 = 2001$ observations (2Hz downsampling) from four different flight tests: t6\_00, t6\_01, t6\_03, and t6\_21. 
Note, from Table \ref{tbl:train_vali_uti_flights}, that we are now training and validating on one instance each of all four types of flights conducted for this analysis: lower four altitudes, upper four altitudes, scan-$\gamma$, and scan-$\epsilon$. 
By comparing Table \ref{tbl:t21_only} to \ref{tbl:t0_t1_t3_t21}, we see that adding observations from a \textit{variety} of flights uniformly reduces RMSE and increases the frequency that error falls within each GPR's 2$\sigma$ error (usually caused by an increase in each GPR's uncertainty). 

Additionally, training on $n_2 = 9999$ observations (10Hz downsampling) from the four training flights gives norm RMSE values of 
0.497$\mu$T, 0.686$\mu$T, 0.406$\mu$T, and 0.408$\mu$T when validating on t6\_04, t6\_05, t6\_06, and t6\_20 respectively. 
Again, a bit to our surprise in this second case, adding more observations from the same ensemble of training flights does not necessarily improve the map's performance. 
However, it is clear that training on multiple flights (Table \ref{tbl:t0_t1_t3_t21}) is better than training on a single flight (Table \ref{tbl:t21_only}).

Aside from the 2Hz vs. 10Hz comparisons done in this section, we do not aim to directly address ways to reduce the cost of training hyperparameters. Instead, we focus on the inference cost $\mathcal{O}\left({n_2}^2 \right)$ with what we call a \textit{compromise map}. 

The idea here is to query the intermediate map (which uses $n_2$ observations for inference) at $n_1$ user-selected locations $\bm{X}^{n_1} \in \mathbb{R}^{3 \times n_1}$ in the working volume. The intermediate map's estimates of the magnetic field at these $n_1$ locations give us a set of  ``measurements'' $\hat{\bm{M}}^{n_1} \in \mathbb{R}^{3 \times n_1}$ which (together with $\bm{X}^{n_1}$) form the ``observations'' used by the compromise map to perform inference. Note that the compromise map will use the same hyperparameters from the intermediate map for its inference. Section \ref{sec:create_query_gpr} gives a more detailed explanation of the process of creating a compromise map from an intermediate map.

\subsubsection{Intermediate Map vs. Compromise Map}
\label{sec:intermediate_map_vs_compromise_map}
We now compare the accuracy of the intermediate map, which uses $n_2 = 2001$ observations (2Hz downsampling) to perform inference, to that of the ``compromise'' map which uses only $n_1 = 511$ inference points. Recall that both use the same sets of hyperparameters optimized over $n_2 = 2001$ observations.

The $n_1=511$ user-selected locations for this analysis are chosen as follows.
Points are distributed evenly through the [-2, 2]m $x$ axis span, [-1.5, 1.5]m $y$ axis span, and [-2.25, -0.5]m $z$ axis span of the working volume. A compromise map location is selected every 0.5m, 0.5m, and 0.25m for the $x$, $y$, and $z$ axes respectively. In total, this gives 504 locations within the working volume. The remaining 7 points are evenly spaced from the ground to an altitude of 0.5m so the compromise map has some observations during the takeoff and landing sequence (which, for all our flights, are above the origin). The number $n_1 = 511$ is an important constraint for another toolbox we used for position localization. Thus, this spatial discretization (0.5m $\times$ 0.5m $\times$ 0.25m) became a common test state in our work.

\begin{figure}
    \centering
     \begin{subfigure}{0.47\textwidth}
      \includegraphics[trim= 10 245 30 210, clip, width=\textwidth]{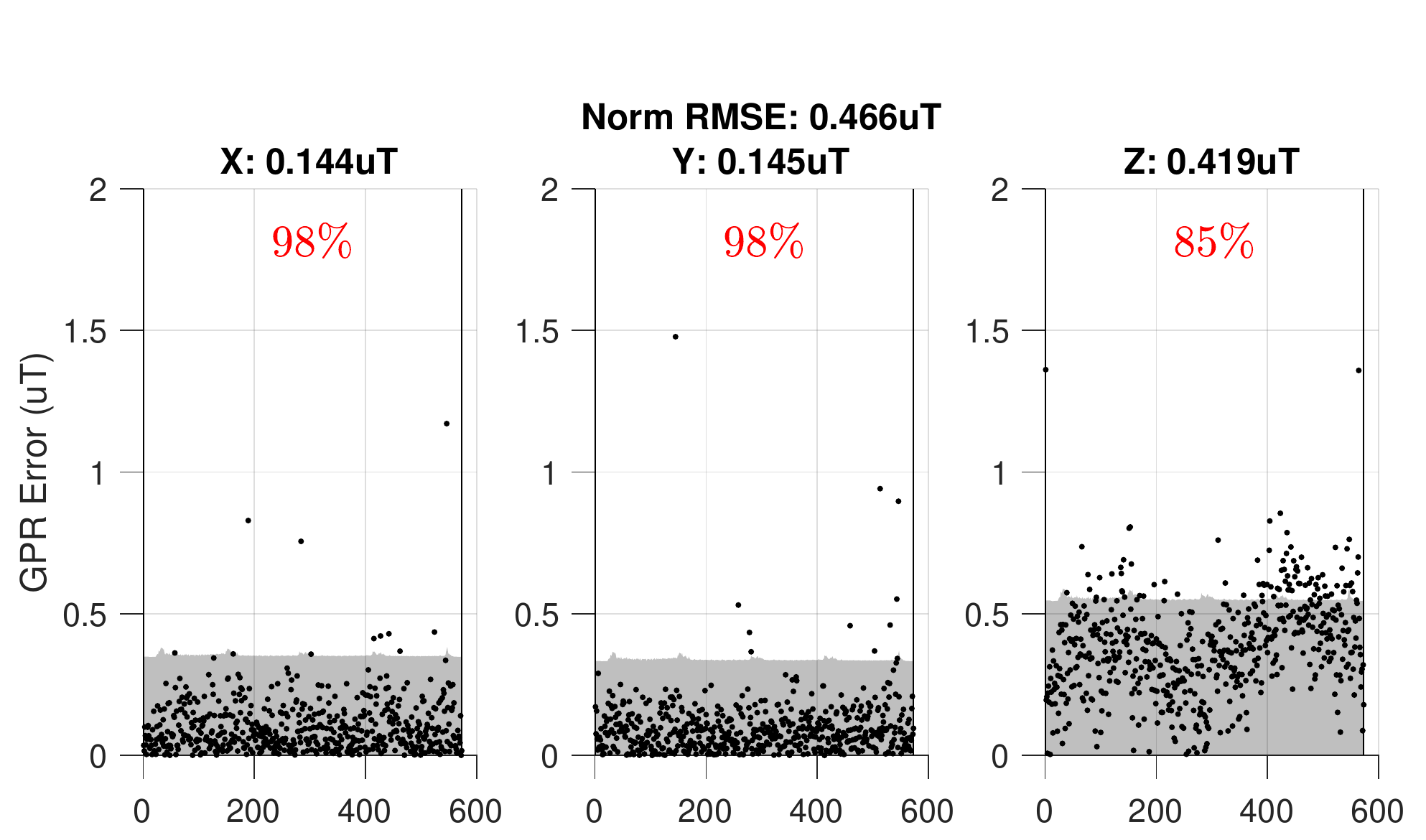}
      \caption{$n_2$ for t6\_04. No outliers.}
      \label{fig:icv_inter_t6_04}
     \end{subfigure}
     \hfill 
     \begin{subfigure}{0.47\textwidth}
      \includegraphics[trim= 10 245 30 210, clip, width=\textwidth]{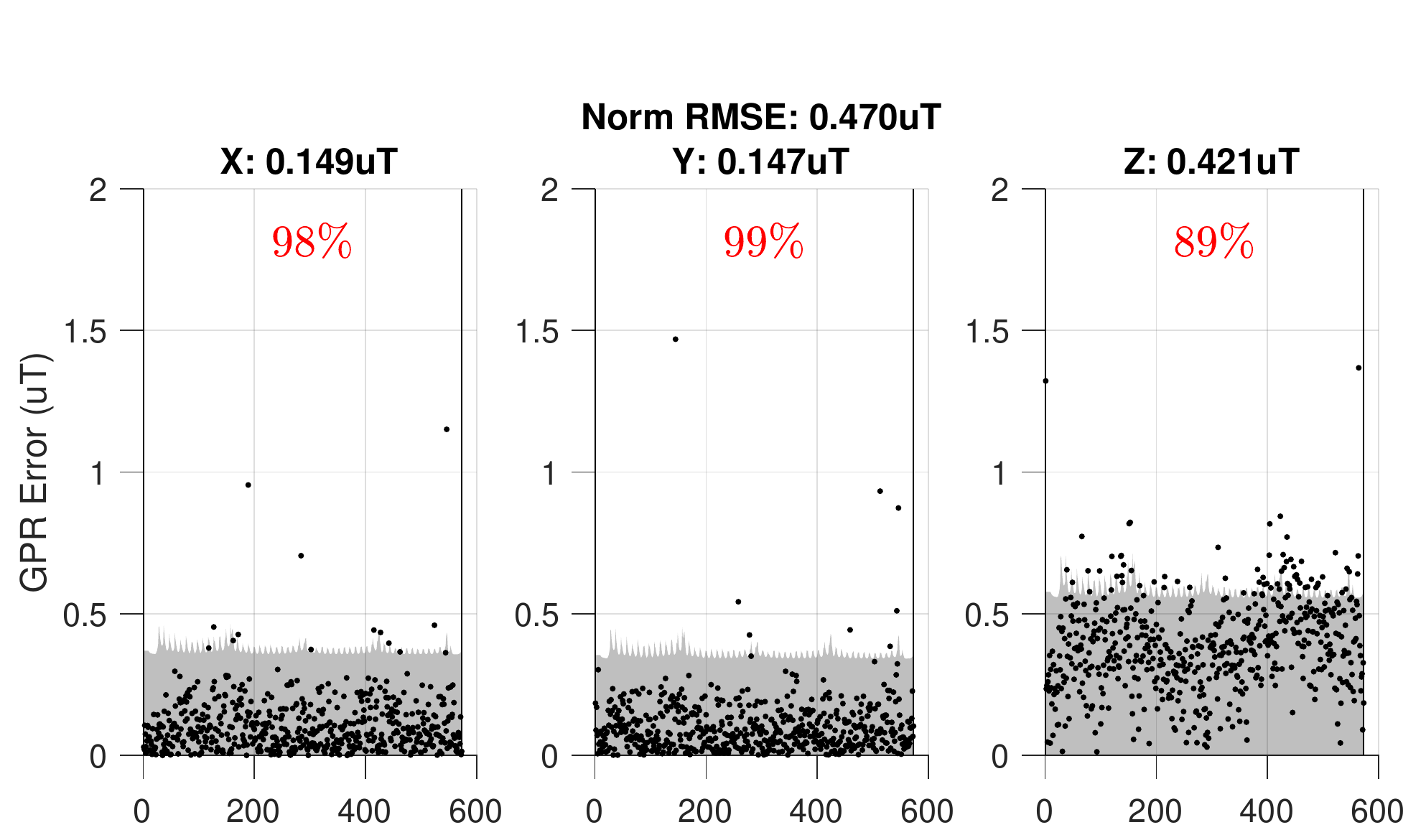}
        \caption{$n_1$ for t6\_04. No outliers.}
      \label{fig:icv_comp_t6_04}
     \end{subfigure}
     \hfill
     \begin{subfigure}{0.47\textwidth}
      \includegraphics[trim= 10 245 30 210, clip, width=\textwidth]{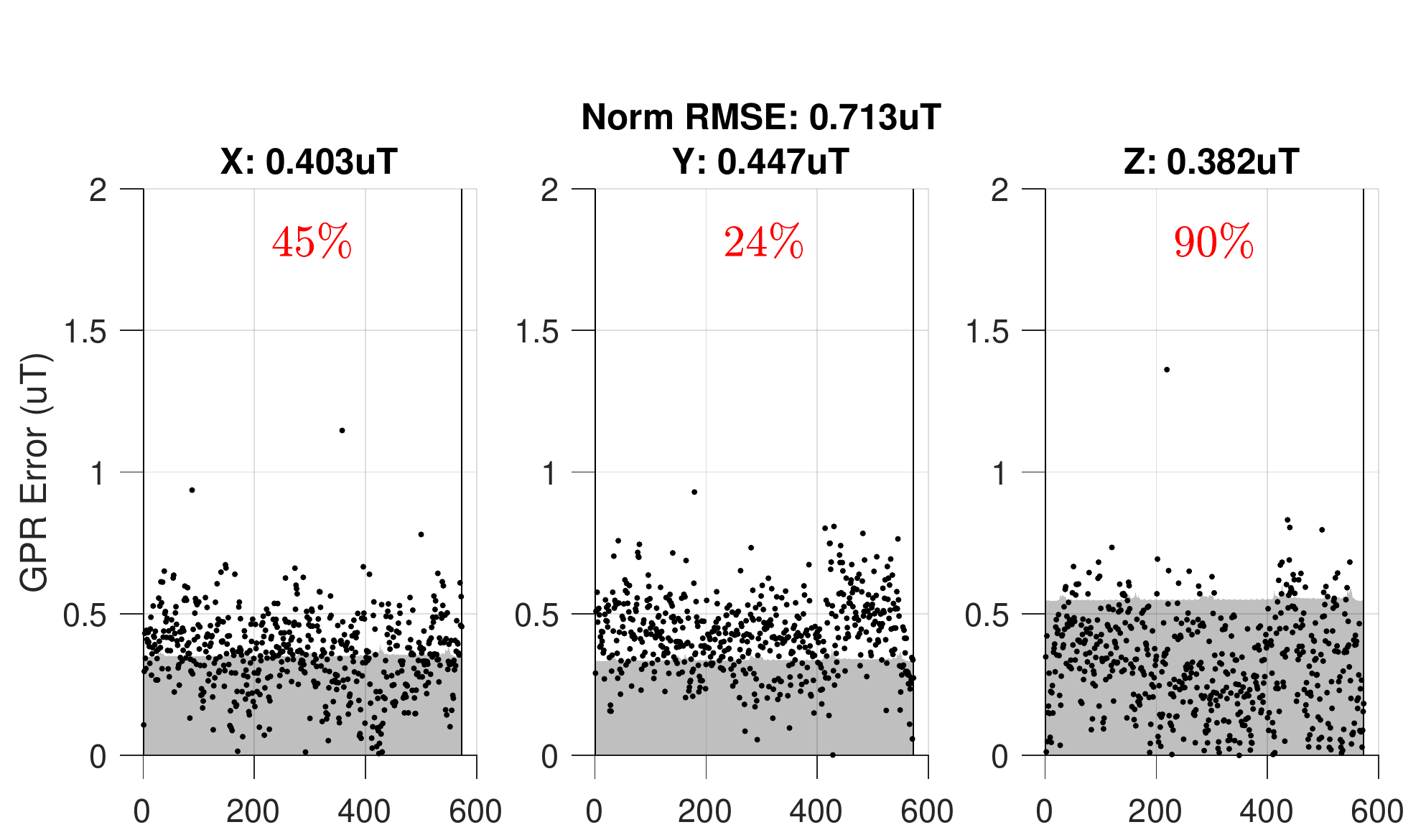}
      \caption{$n_2$ for t6\_05. Outliers \{[2.89], [], [2.42]\}$\mu$T.}
      \label{fig:icv_inter_t6_05}
     \end{subfigure}
     \hfill 
     \begin{subfigure}{0.47\textwidth}
      \includegraphics[trim= 10 245 30 210, clip, width=\textwidth]{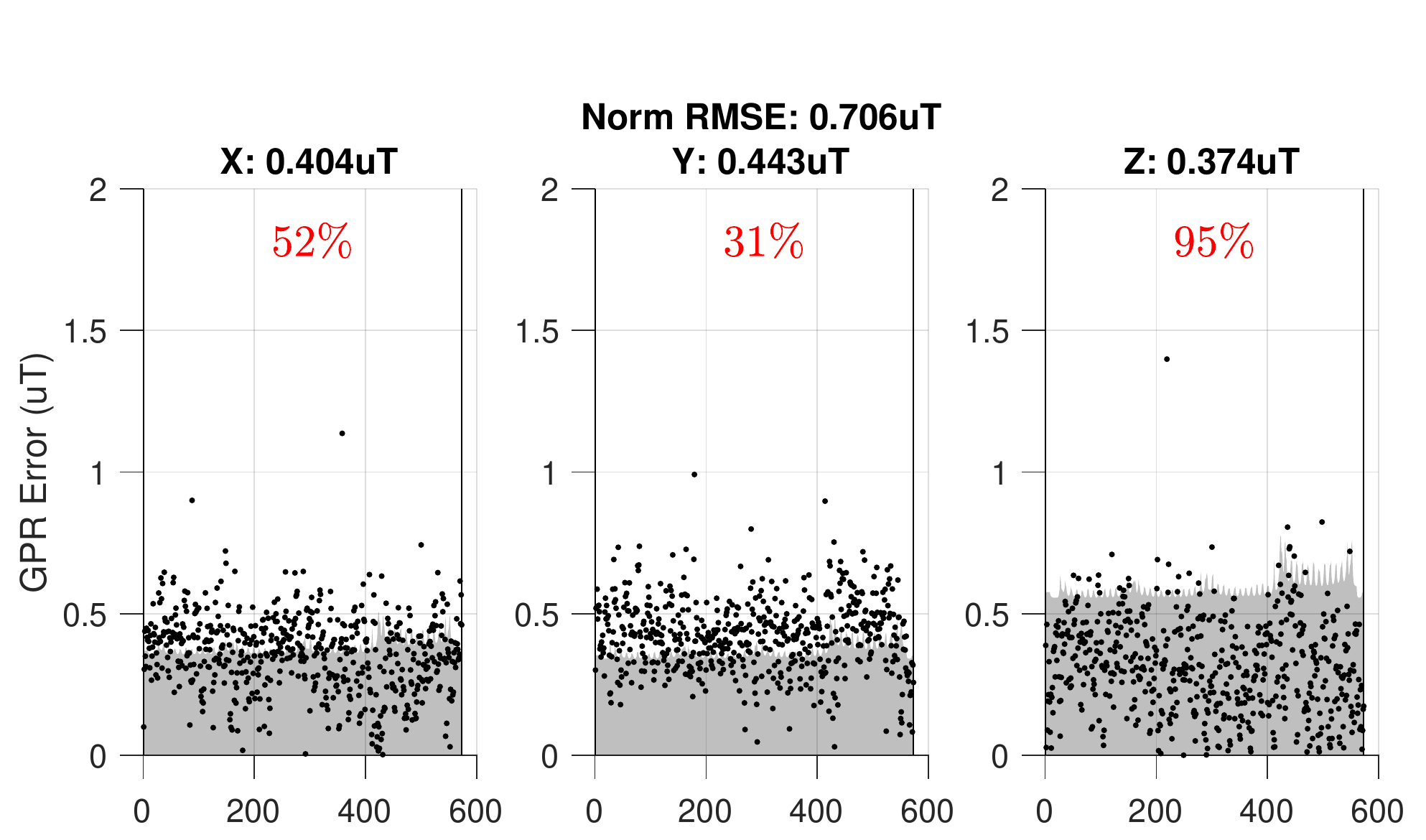}
      \caption{$n_1$ for t6\_05. Outliers \{[2.49],[],[2.39]\}$\mu$T.}
      \label{fig:icv_comp_t6_05}
     \end{subfigure}
     \hfill
     \begin{subfigure}{0.47\textwidth}
      \includegraphics[trim= 10 245 30 210, clip, width=\textwidth]{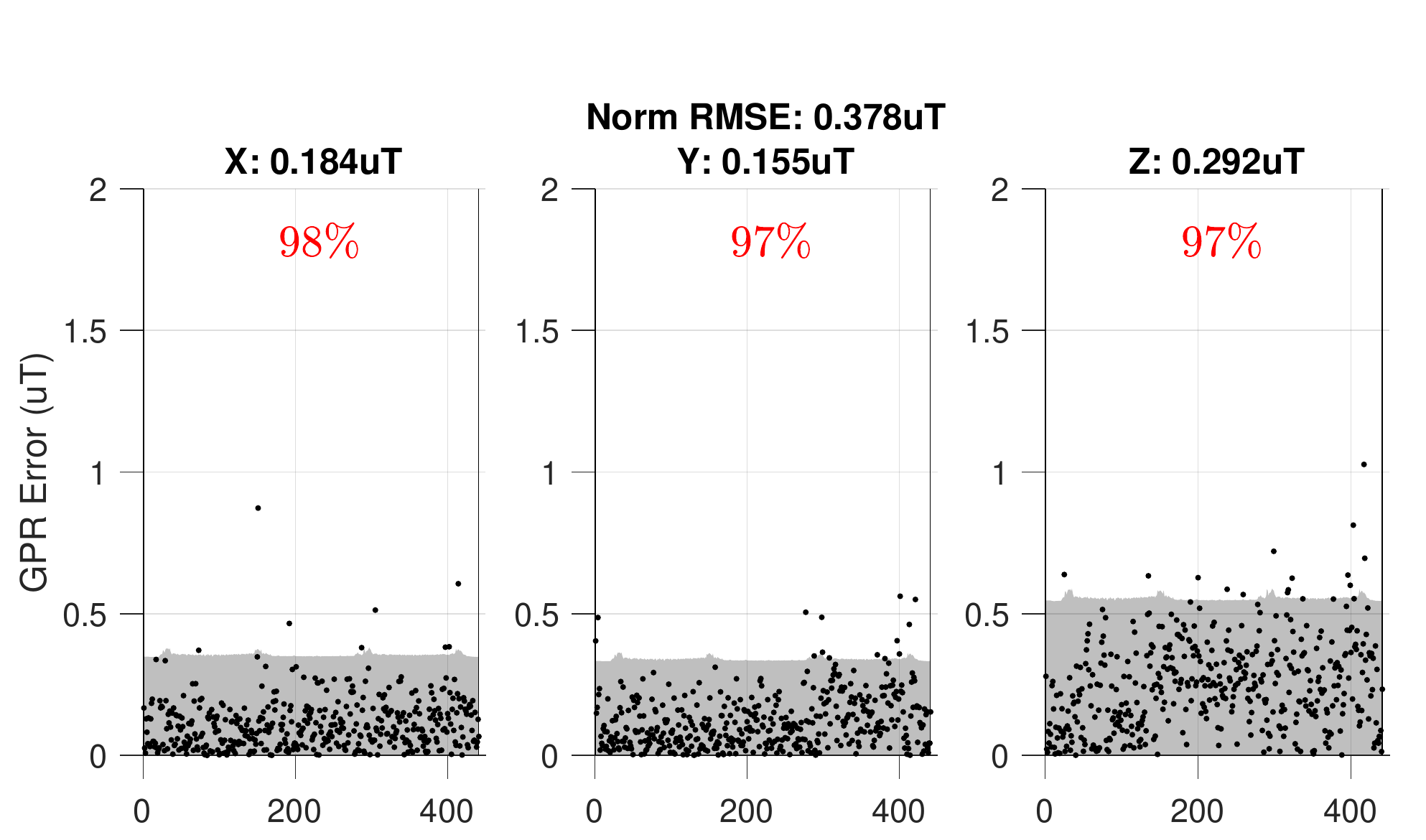}
      \caption{$n_2$ for t6\_06. Outliers \{[2.41],[],[]\}$\mu$T.}
      \label{fig:icv_inter_t6_06}
     \end{subfigure}
     \hfill 
      \begin{subfigure}{0.47\textwidth}
      \includegraphics[trim= 10 245 30 210, clip, width=\textwidth]{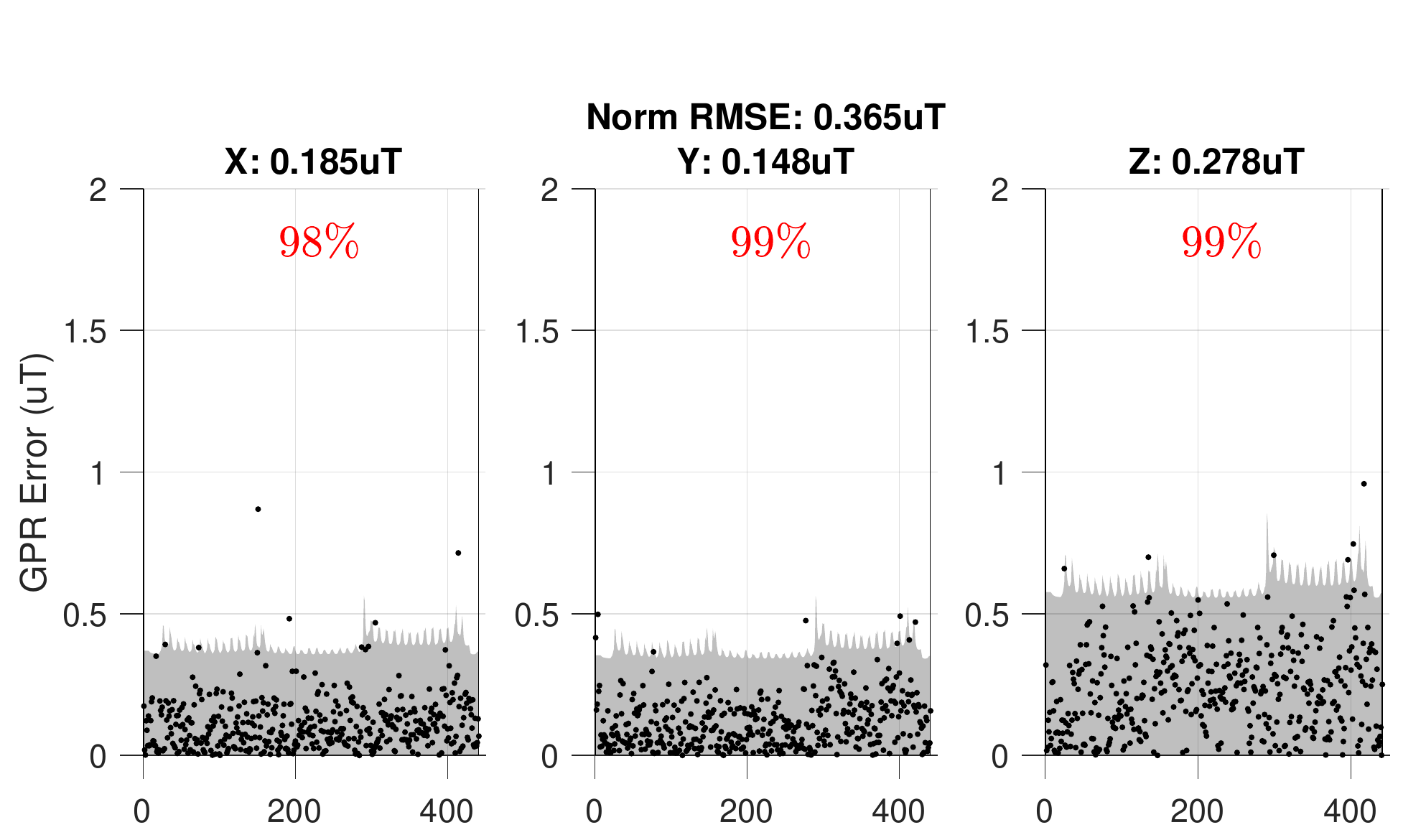}
      \caption{$n_1$ for t6\_06. Outliers \{[2.36], [], []\}$\mu$T.}
      \label{fig:icv_comp_t6_06}
     \end{subfigure}
     \hfill
      \begin{subfigure}{0.47\textwidth}
      \includegraphics[trim= 10 245 30 210, clip, width=\textwidth]{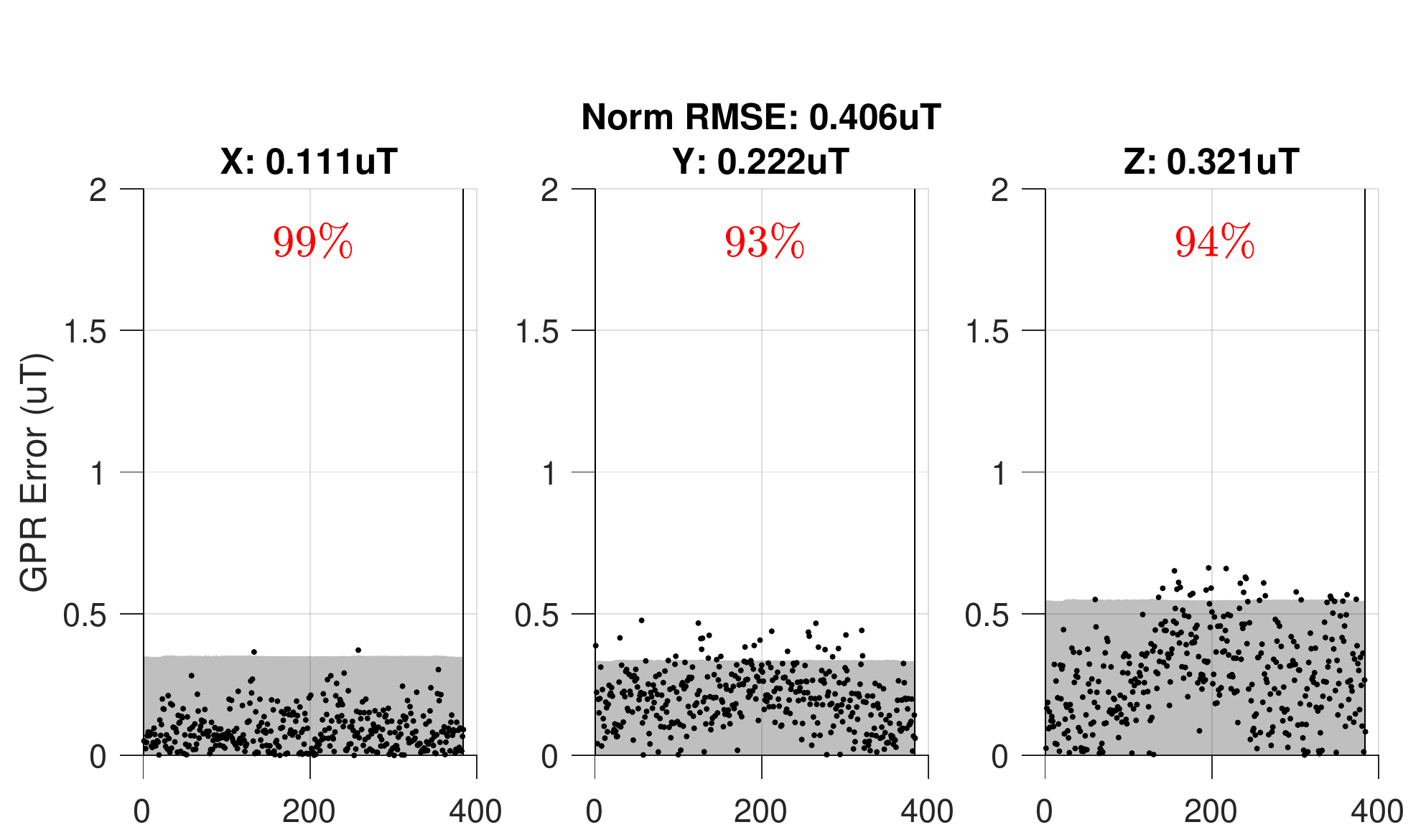}
      \caption{$n_2$ for t6\_20. No outliers.}
      \label{fig:icv_inter_t6_20}
     \end{subfigure}
     \hfill 
      \begin{subfigure}{0.47\textwidth}
      \includegraphics[trim= 10 245 30 210, clip, width=\textwidth]{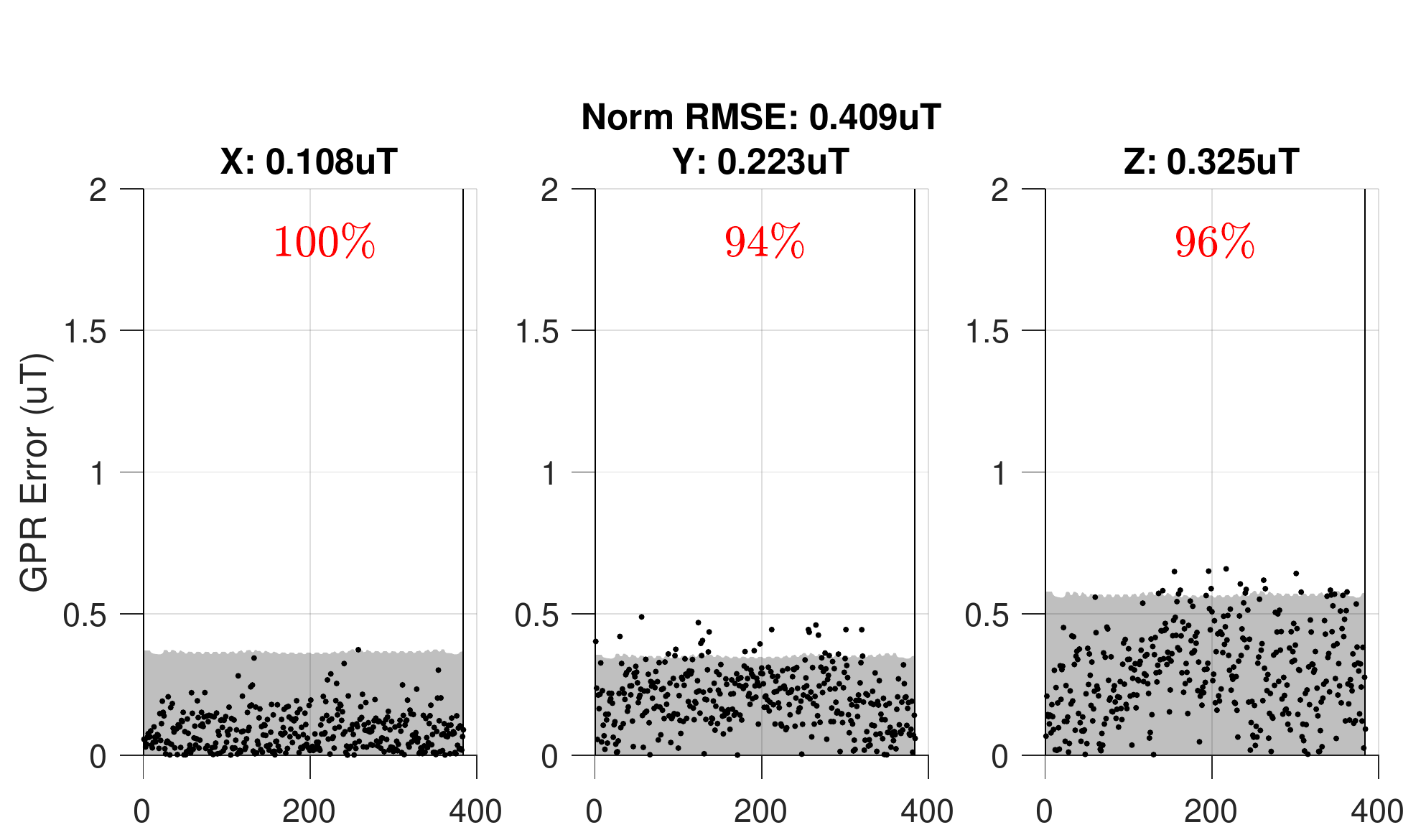}
      \caption{$n_1$ for t6\_20. No outliers.}
      \label{fig:icv_comp_t6_20}
     \end{subfigure}
 \caption
 {Performance of the intermediate map (left column) and compromise map (right column). 
 Outliers larger than 2$\mu$T are listed for \{ [$\mathcal{GPR}_{x}$], [$\mathcal{GPR}_{y}$], [$\mathcal{GPR}_{z}$] \} in each subplot.
 Empty brackets indicate no hidden outliers for the respective $\mathcal{GPR}$.}
 \label{fig:inter_compro_validation}
\end{figure}

We can see these trends better in Figure \ref{fig:inter_compro_validation} which is similar in style to the grayscale plots from Figure \ref{fig:s2_gprErr}, but with the subplots as three columns rather than as three rows. This format fits more figures on a single page to more easily compare the intermediate and compromise maps. 

Figure \ref{fig:inter_compro_validation} gives plots for the intermediate map on the left column and the compromise map on the right. Here, we see the compromise map has both \textit{quantitatively} and \textit{qualitatively} similar performance to the intermediate map despite using nearly a fourth of the points for inference. By comparing the norm RMSE values (in the titles) across the two columns of Figure \ref{fig:inter_compro_validation}, we see the two maps are never off by more than 0.013$\mu$T (13 nanoTesla) in norm RMSE. This difference is within the noise of the magnetic field measurements of our platform. 

For reference, training a GPR map on 408 observations (2Hz downsample) from t6\_21 and validating this map on the same 408 observations gives RMSE values of (0.196$\mu$T, 0.089$\mu$T, 0.108$\mu$T, 0.136$\mu$T) for norm, $x$, $y$, and $z$ RMSE values respectively. This means that on our flight platform, even when the map has overfit for the exact observations it will be validated on, it still cannot discern differences of 0.013$\mu$T. 

Thus, we take the differences between the intermediate map ($n_2 = 2001$) and the compromise map ($n_1 = 511$) in Figure \ref{fig:inter_compro_validation} to be negligible and assert that their RMSE performance is effectively identical. 

Of course, a lot of valuable information can be lost by simply comparing RMSE values. However, visual comparison across the two columns of Figure \ref{fig:inter_compro_validation} further emphasizes that both the intermediate and compromise maps yield very similar prediction results. The key difference is in the uncertainty of the two maps. 

It's easiest to see this in row three (Figures \ref{fig:icv_comp_t6_05} vs. \ref{fig:icv_inter_t6_06}) where the compromise map (right) has spikier gray shading (2$\sigma$ uncertainty) than the intermediate map (left).
The compromise map's increased uncertainty is primarily due to our choice of prediction locations  $\bm{X}^{n_1} \in \mathbb{R}^{3 \times n_1}$ which have 504 points selected from a (0.5m $\times$ 0.5m $\times$ 0.25m) spatial discretization of our working volume. Recall that each altitude of our validation trajectories traverses lanes separated by 0.25m (Figure \ref{fig:1_5m_traj}). Thus, anytime the compromise map is queried at a location between the (0.5m $\times$ 0.5m) $x$-$y$ points, it will report a higher uncertainty in its predicted magnetic field estimate.

The overall increased uncertainty of the compromise map also explains why the respective GPR errors more frequently fall within two standard deviations of their respective uncertainties. Said another way: the plots on the right column of Figure \ref{fig:inter_compro_validation} will almost always have higher red percentages than the comparative GPR on the left column \textit{given our selected spatial discretization of} (0.5m $\times$ 0.5m $\times$ 0.25m). 

In general, we expect the overall uncertainty of the compromise map to decrease with increased spatial density in our selection of locations for $\bm{X}^{n_1}$. The next section will further investigate how the spatial density of the points $\bm{X}^{n_1}$ affect the performance of the compromise map. 

\subsection{Compromise Map - Spatial Density Analysis}
\label{sec:compromise_map_spatial_density_analysis}
This section seeks to understand how the accuracy of the compromise map changes as a function of the spatial density of the locations used in $\bm{X}_1$. Generally, we expect GPR error to increase, the number of error points captured by 2$\sigma$ uncertainty to increase, and the computation time to \textit{decrease} as the training set points become more sparse. In this section, we will only analyze the first of these (GPR accuracy). 

In the last section, our $n_1 = 511$ compromise training points came from a custom spatial density of (0.5m $\times$ 0.5m $\times$ 0.25m) which gave 504 locations, plus an additional 7 locations to have observations along the takeoff and landing segment for each flight. 
In this section, we will work with a uniform spatial density in all directions ($S$ $\times$ $S$ $\times$ $S$). Additionally, to simplify matters a bit, we will use a naive linear spacing of [$min:S:max$] for the locations along each respective spatial axis of the lab. This means that different spatial densities $S$ can yield the same \textit{number} of points $n_1$ but at different locations. Recall that our working volume has spatial limits of [-2, 2]m in $x$, [-1.5, 1.5]m in $y$, and [-2.25, -0.5]m in $z$. 

In this study, we vary $S$ from 0.2m to 1m to emulate the study done in ref. \cite{Akai2017}.  However, our study will quantify the RMSE of the magnetic field map rather than relying exclusively on visual comparisons of the map. For the 17 values of $S$ tested (in increasing order), $n_1$ = [3031,  1775,  931,   655,    447,   259,   259,   199,   133,   112,   97,    97,    79,    67,    47,    47,    47]. Recall, seven of the $n_1$ values listed here are for the takeoff and landing sequence.

\begin{figure}
    \centering
  \captionsetup{justification=centering}
  \includegraphics[width= 0.7\textwidth,trim= 10 160 20 130, clip]{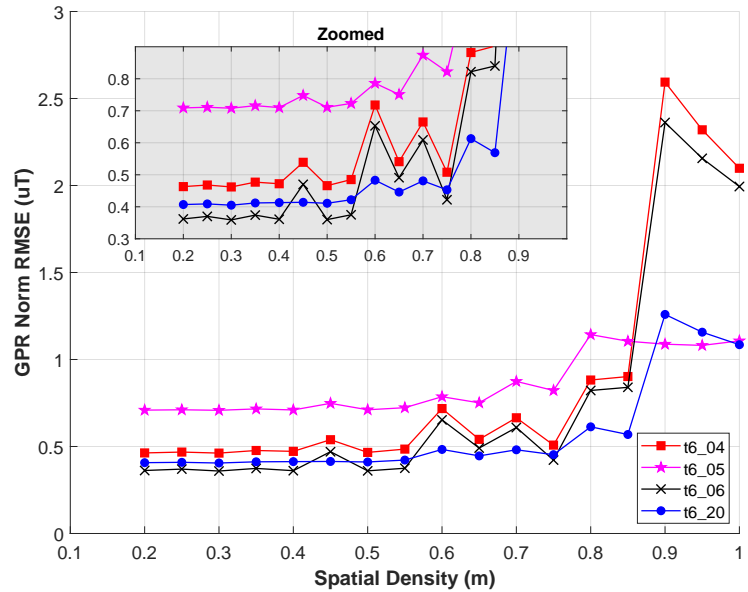}
  \caption{GPR norm RMSE fairly constant for  $S \leq 0.55$m. Flights t6\_04 (red $\square$) and t6\_06 (black $\times$) have observations at $z = -0.5$m causing higher errors for certain values $S$.}
 \label{fig:spatialDensity_lineGraph}
\end{figure}

Figure \ref{fig:spatialDensity_lineGraph} shows the norm RMSE of the compromise map (validated on t6\_04, t6\_05, t6\_06, t6\_20) as a function of the spatial density term $S$. 
A zoomed version of the initial segment of data is embedded in the same figure.
Here we see that the norm RMSE is fairly constant for values $S \leq 0.5$m with a small spike in norm RMSE at $S = 0.45$m. 
This spike is caused by our naive linear spacing which occasionally causes poor sampling along altitudes. At $S = 0.45$m, we get altitudes of $z = $ [-2.25, -1.8, -1.35, -0.9]m.
Recall  from Section \ref{sec:multi_altitude_trajectories} that some of our validation flights have measurements as low as $z =$ -0.5m.

Combining these sources of information, we can see that the black-$\times$ (t6\_06; Scan-$\gamma$) and red-$\square$ (t6\_04; Lower four alts) suffer the worst increases at $S = 0.45$m. Since the compromise map's observations only go as low as $z=-0.9$m, both these trajectories have higher GPR RMSE when validated at their $z=-0.5$m observations. Meanwhile, blue-$\circ$ (t6\_20; Scan-$\epsilon$) shows little change at $S = 0.45$m since all its observations are away from the extremes of the working volume. 
The other transient spikes in Figure \ref{fig:spatialDensity_lineGraph} (e.g., $S = 0.6$m) are caused by similar $z$-axis sampling from our naive linear spacing. t6\_04 and t6\_06 continue to be the most sensitive to certain values $S$ when the map has no observations near $z = -0.5$m.


Despite these transient spikes, there are two clear trends we can see from Figure \ref{fig:spatialDensity_lineGraph}. First, for values of $S \leq 0.55$m (with $S=0.45$m as an exception), the norm RMSE is relatively insensitive to changes in $S$. This agrees quite well with the ref. \cite{Akai2017} where they show (visually) that their magnetic field map is qualitatively similar for values of $S = $ [0.2, 0.4, 0.6]m. In \cite{Akai2017}, their analysis led them to use 0.6m as a standard distance between observations in another experiment they conduct.

For comparison, we briefly summarize the spatial density of observations used by some other indoor magnetic field mapping works.  Some works map along a single axis ($p=1$) such as \cite{Haverinen2009} that separates observations by 0.04m and 0.1m on a ground robot and chest-mounted pedestrian setup respectively through hallway networks of up to 350m in length. 
Additionally, \cite{Suksakulchai2000} makes a $p=1$ map with 0.2m between each observation and \cite{Akai2015} make a $p=1$ map with 0.1m between observations.

For $p=2$, planar maps \cite{Li2012} create a constant-altitude planar map using $S = 0.305$m separation in $x$ and $y$ for their coarse grid and $S = 0.005$m for fine grid spacing. This mapping is done in a 2.1m$\times$2.1m room.
Ref. \cite{Vallivaara2011} uses a type of occupancy grid for their $p=2$ magnetic field SLAM solution with grid size 0.05m$\times$0.05m. The solution ignores all previous magnetic field measurements further than 0.5m from the vehicle's current position.
Ref. \cite{Wu2019} makes $p=2$ maps with lanes separated by 0.38m.
Ref. \cite{Hanley2021} creates $p=2$ maps at different heights along hallways in three separate buildings. The authors use a longitudinal separation distance of about 0.4m (varies per building) and a vertical separation of 0.08m between their observations.

Finally, Ref. \cite{Brzozowski2016} makes a $p=3$ ``map'' of a hallway (no explicit interpolation of observations) with (0.33m $\times$ 0.45m $\times$ $S^?$) where the vertical spacing $S^?$ is not explicitly listed in their work, and
\cite{Lipovsky2021} makes $p=3$ maps with $S= 0.5$m uniform spacing in all axes.
Our previous work used planar lawnmower patterns like in Figure \ref{fig:1_5m_traj}, but with 0.5m spacing in the $y$ axis (instead of the 0.25m spacing used in this work) \cite{Kuevor2021}. 
Furthermore, we used a vertical spacing of 0.75m in our training sets to investigate the accuracy of $p=3$ magnetic field maps when interpolating between and extrapolating outside of pre-mapped altitudes.

In short, for 1D, 2D, or 3D \textit{indoor} magnetic field maps, we have not seen any previous works using separation distances $S$ larger than 0.6m.
The exceptions are \cite{Akai2017} where the authors explicitly study a reasonable upper limit on $S$ and our work \cite{Kuevor2021} which studies the interpolative and extrapolative limits of $p=3$ GPR maps.

We believe the empirical upper bound of $S = 0.6$m described above might be driven by proximity to walls and the floor. 
If this is the case, $S \leq 0.6$m might only be the upper bound for indoor regions that are within a couple meters from some wall or the floor.
We acknowledge that this includes nearly every room or hallway in most buildings, but a large, multi-story atrium might be an exception. 
Portions of such a large, open, yet indoor volume may have lower spatial variation and can be accurately mapped with values $S > 0.6$m.
The size of the two works that explicitly study spatial density indoor magnetic field mapping (3m $\times$3m$\times$2.2m for \cite{Akai2017} and 4m$\times$3m$\times$2.25m in this work) might be too small (i.e., too close to the walls and floor) to test this idea.

It is also possible that the floor, and not walls, are the dominating factor for spatial variation in indoor magnetic fields.
In \cite{Hanley2021}, Hanley et al. show that that the magnetic field near the floor (within 0.5m from the ground) of three university buildings is most distinct from the field at other altitudes up to 2m. 
This might explain why the other works in our spatial density summary, which mostly use ground robots and pedestrians that remain ``near'' the floor, have $S \leq 0.6$m.

The second conclusion, the complement of the first, is that values of $S \geq 0.6$m start to show steady increases in norm RMSE. It is not clear what trends hold for $S > 1$m, but such separation distances are somewhat degenerate given the size of our working volume. Further, it is unlikely for the norm RMSE values to return to their $S = 0.2$m for values of $S > 1$m (which is what we are most interested in).

This analysis is what led us to use $S = 0.5$m for the $x$-$y$ separation distance for the $n_1 = 511$ compromise map used elsewhere in this paper.
We chose $S = 0.25$m for $z$ because we found (empirically) that the GPR's performance is most sensitive to omissions in observed altitudes. 
Recent works on indoor magnetic field mapping have shown similar trends: that indoor magnetic fields change noticeably as a function of altitude \cite{Kuevor2021, Hanley2021}.
For our analysis, similarly for \cite{Kuevor2021, Hanley2021}, it is not clear how much this altitude-based sensitivity is driven by the fact that all our flights (training and validation) are comprised of many single-altitude slices. 


This section has shown that our compromise map can perform accurate magnetic field inference with fewer prediction points $n_1 < n_2$. Remember the goal here was to allow the GPR-based magnetic field map to learn the flight-by-flight variations across several training flights without incurring high computation costs when performing inference.
However, some validation flights like t6\_05 (Figures \ref{fig:icv_inter_t6_05} and \ref{fig:icv_comp_t6_05}) still have steady-state errors that even the compromise map (or the intermediate map) is unable to resolve given a set of training flights. 

In the following section, we compare two methods of mapping the \textit{norm} of the magnetic field in our workspace to investigate if composing the three outputs from a $3 \rightarrow 3$ map gives a less-accurate norm estimate than training a specialized $3 \rightarrow 1$ map on the magnetic field norm.

\subsection{Creating a  \texorpdfstring{$3 \rightarrow 1$}{TEXT} map from estimates of a \texorpdfstring{$3 \rightarrow 3$}{TEXT} map}
\label{sec:3_3_vs_3_1_norm_map}

This section compares two ways of estimating the norm of the magnetic field in a workspace. The first takes the three estimates from our $3 \rightarrow 3$ map and computes the norm of these estimates $\sqrt{ \hat{m}_{x}^2  + \hat{m}_{y}^2 + \hat{m}_{z}^2 }$. The alternative is to create a new $3 \rightarrow 1$ map, called $GPR_{nrm}$, trained on the norm of our measurements

\begin{equation}
    \begin{split}
        \tilde{y}_{nrm} & = \sqrt{ \tilde{y}_x^2 + \tilde{y}_y^2 + \tilde{y}_z^2} \in \mathbb{R} \\
    \end{split}
    \label{eq:nrm_meas}
\end{equation}
to create hyperparameter ($D_{nrm}^{n_2}$) and inference ($D_{nrm}^{n_1}$) sets. The scalar predictions from this new GPR are defined as $\hat{m}_{nrm} \in \mathbb{R}$.

The goal here is to see if a specialized $3 \rightarrow 1$ map can more accurately estimate the magnetic field norm than a composition of the estimates from a $3 \rightarrow 3$ map. 
This is an important distinction since some works create maps on special components of the ambient magnetic field like a $2 \rightarrow 2$ map on the horizontal and vertical components of the field \cite{LeGrand2012, Vandermeulen2013} or a $1 \rightarrow 1$ map on the heading (or declination) of the field \cite{Suksakulchai2000}.
Here, we aim to check if composing the three estimates from a $3 \rightarrow 3$ map accumulates error in estimating the norm of the field in ways that a specialized $3 \rightarrow 1$ map, which estimates the norm directly, would not.

The procedure for training hyperparameters and creating the $n_1 = 511$ compromise map for our new $GPR_{nrm}$ is the same as described in Section \ref{sec:create_query_gpr}, but with measurements as defined in Equation \ref{eq:nrm_meas}.

The RMSE accuracy of our maps requires a prediction from a GPR (or a set of GPRs) at location $\bm{r}^*$ and a measurement taken at the same location. We refer to the prediction given by our typical $3 \rightarrow 3$ map as our \textit{vector} prediction 
\begin{equation}
    \begin{split}
        \hat{\bm{m}}_{vec}(\bm{r}^*)
        = 
        \begin{bmatrix}
        \hat{m}_x(\bm{r}^*) \\
        \hat{m}_y(\bm{r}^*) \\
        \hat{m}_z(\bm{r}^*)
        \end{bmatrix}
        & = 
        \begin{bmatrix}
        \mathcal{GPR}_x(\bm{r}^*) \\
        \mathcal{GPR}_y(\bm{r}^*) \\
        \mathcal{GPR}_z(\bm{r}^*)
        \end{bmatrix}.
        \\
    \end{split}
\end{equation}

\noindent Now we define two error terms meant to compare the accuracy of estimating the magnetic field norm in our workspace. One is for the accuracy of our $3 \rightarrow 3$ vector map ($vec$) while the other is for the $3 \rightarrow 1$ norm map ($nrm$).

\begin{equation}
    \begin{split}
        e_{vec\_nrm} &= \sqrt{ \hat{m}_{x}^2  + \hat{m}_{y}^2 + \hat{m}_{z}^2 } - \tilde{y}_{nrm}  \\
        e_{nrm\_nrm} &=   \hat{m}_{nrm}  - \tilde{y}_{nrm} . \\
    \end{split}
    \label{eq:err_vec_new}
\end{equation}



\noindent The error terms in Equation \ref{eq:err_vec_new}
are used to create RMSE metrics (Equation \ref{eq:rmse}) for the accuracy of each respective GPR against some validation set. 

\begin{table}
 \centering
 \caption{Norm ($nrm$) GPR versus vector ($vec$) GPR on estimating norm of magnetic field.}
 \begin{tabular}{l || r r } 
\toprule

    Flight ID and & \multicolumn{2}{c }{\textit{RMSE} ($\mu$T) } \\
    
  Description & $e_{vec\_nrm}$ & $e_{hrz\_nrm}$  \\

 \midrule

t6\_04 (Lower Four Alts.)   & 0.391 & 0.390 \\
t6\_05 (Upper Four Alts.)   & 0.256 & 0.257 \\
t6\_06 (Scan-$\gamma$)      & 0.245 & 0.246 \\
t6\_20 (Scan-$\epsilon$)    & 0.301 & 0.301 \\
\end{tabular}
\label{tbl:hrz_nrm_metrics}
\end{table}

Table \ref{tbl:hrz_nrm_metrics} computes RMSE on the error metrics of Equation \ref{eq:err_vec_new} 
for the four validation flights we used in the previous section. From this, we see that both methods of estimating the norm of the magnetic field are equivalent. There is never more than 1nT of difference between the two RMSE values which is well within the noise floor of the PNI RM3100 (even without accounting for noise generated by the drone). 

Our result is reassuring in that there is no need to train a fourth GPR $GPR_{nrm}$ if a user wants the three vector components \textit{and} the norm of the magnetic field in their workspace. It is not immediately clear if this result extends to angle-based compositions of the magnetic field map like declination \cite{Suksakulchai2000} or inclination of the ambient field. 

Our last tool in this paper, which we call the ``consistency metric'', uses the frequency of error falling with a GPR's 2$\sigma$ interval to indicate if a validation flight is consistent with a given GPR map. 

\subsection{Consistency Check}
\label{sec:consistency_check}
Despite our efforts to reduce magnetic noise from motors and ESCs (Sections \ref{sec:flight_ctrl_noise} and \ref{sec:s2_and_flight_by_flight_variations}) and incorporate flight-by-flight variations into a compromise map (Section \ref{sec:accuracy_of_compromise_map}), there may still be flight tests with significantly different magnetic field observations than what a given GPR map can predict. Such cases can arise from the flight-by-flight variations of the same vehicle (e.g., Figures \ref{fig:icv_inter_t6_05} and \ref{fig:icv_comp_t6_05}), but may also come into play if different platforms are used for mapping and utilization (e.g., GPR map trained on drone data but utilized for indoor pedestrian localization). 
As such, it is important for users to know when they can rely on the predictions from their GPR-based magnetic field map, and when it may be better to leverage a different sensing modality for their state estimates. 

For this, we introduce the concept of ``consistency'' between a GPR-based map and a validation flight. Given a $p \rightarrow m$ magnetic field map, we say that a validation flight is \textit{consistent} with the map if all $m$ GPRs respectively capture 96\% of their error within two standard deviations (2$\sigma$) of their uncertainty. 

Of course, the 96\% threshold is inspired by the expected number of points that fall within two standard deviations of a univariate normal (Gaussian) distribution. Additionally, we note that a 96\% threshold works well for our test platform, our choice of GPR kernel (squared exponential), the way we optimize hyperparameters (minimization of log marginal likelihood; Section \ref{sec:gpr}), and our target application of position localization. Although these kernel and optimization methods are common, we have not investigated how GPR-based maps generated with different kernels or different hyperparameters behave in regard to our notion of consistency. 

To demonstrate, we create a $n_1 = 511$ compromise map of 2Hz-downsampled observations from the following seven flight tests: t6\_0, t6\_1, t6\_3, t6\_4, t6\_5, t6\_6, t6\_16. 
We validate this compromise map on 10Hz samples from four flight tests (t6\_09, t6\_11, t6\_15, t6\_18) to illustrate how we use the consistency metric.
Note that we are training and validating on different t6\_XX tests compared to the previous sections.
The training flights used here are from an analysis of magnetic field position localization which we will explore in a future paper.

\begin{figure*}
 \begin{subfigure}[b]{0.48\textwidth}
  
  \includegraphics[width=\textwidth,trim= 20 160 20 130, clip]{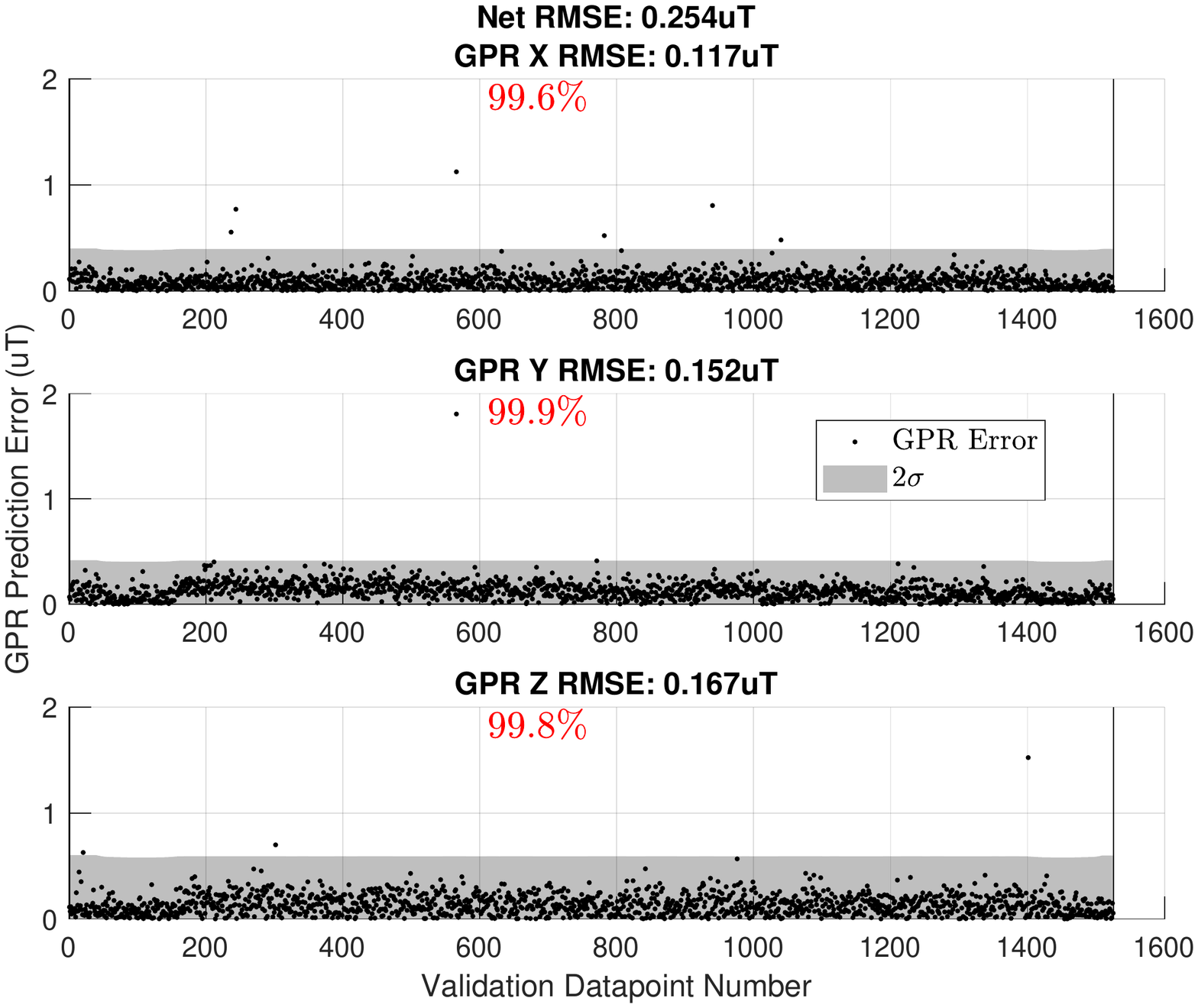}
  \caption{t6\_09. No hidden outliers.}
  \label{fig:consistency_metric_1}
 \end{subfigure}
 \hfill
 \begin{subfigure}[b]{0.48\textwidth}
  
  \includegraphics[width=\textwidth,trim= 20 160 20 130, clip]{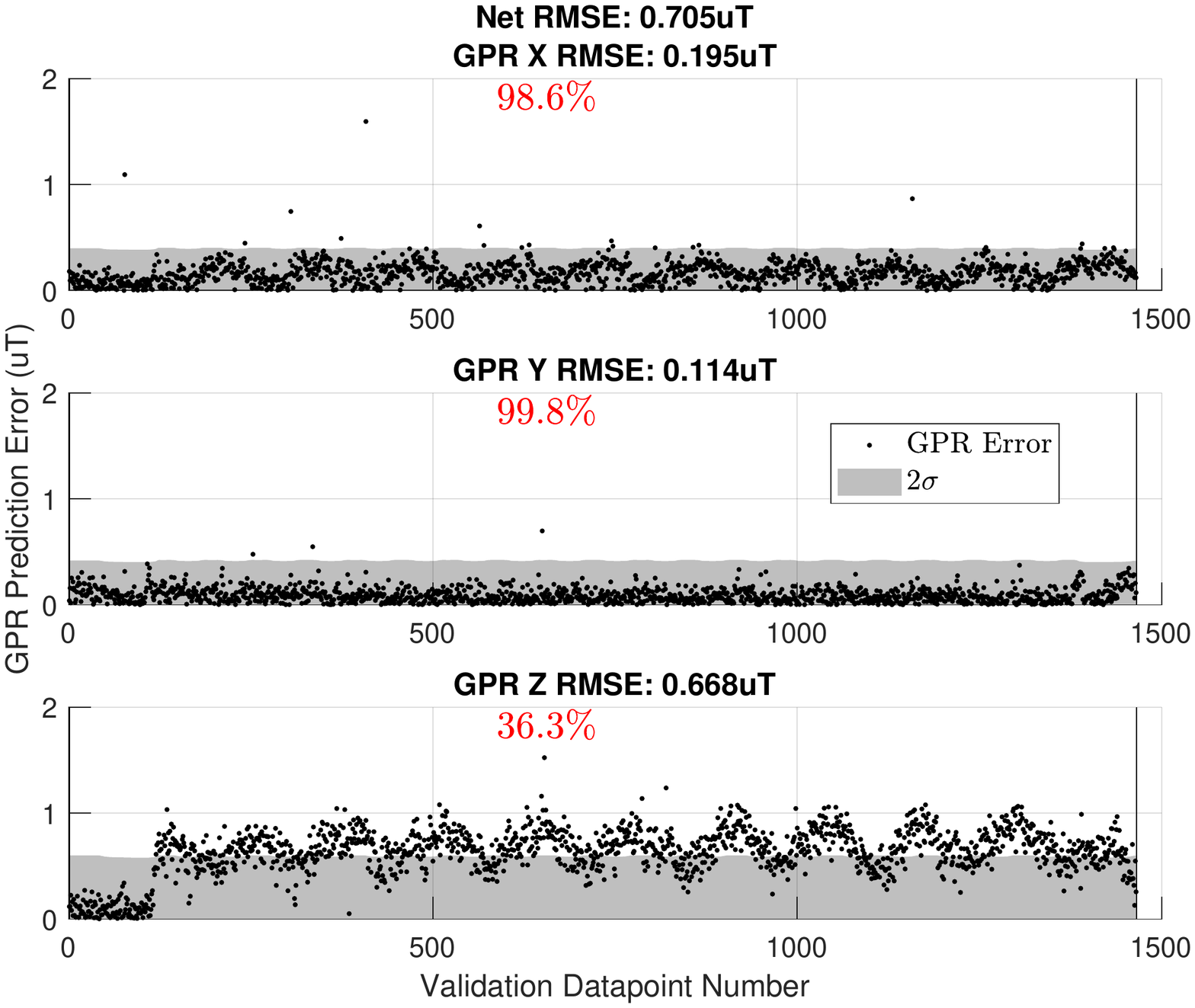}
  \caption{t6\_11. No hidden outliers.}
  \label{fig:consistency_metric_2}
 \end{subfigure}
 \begin{subfigure}[b]{0.48\textwidth}
  
  \includegraphics[width=\textwidth,trim= 20 160 20 130, clip]{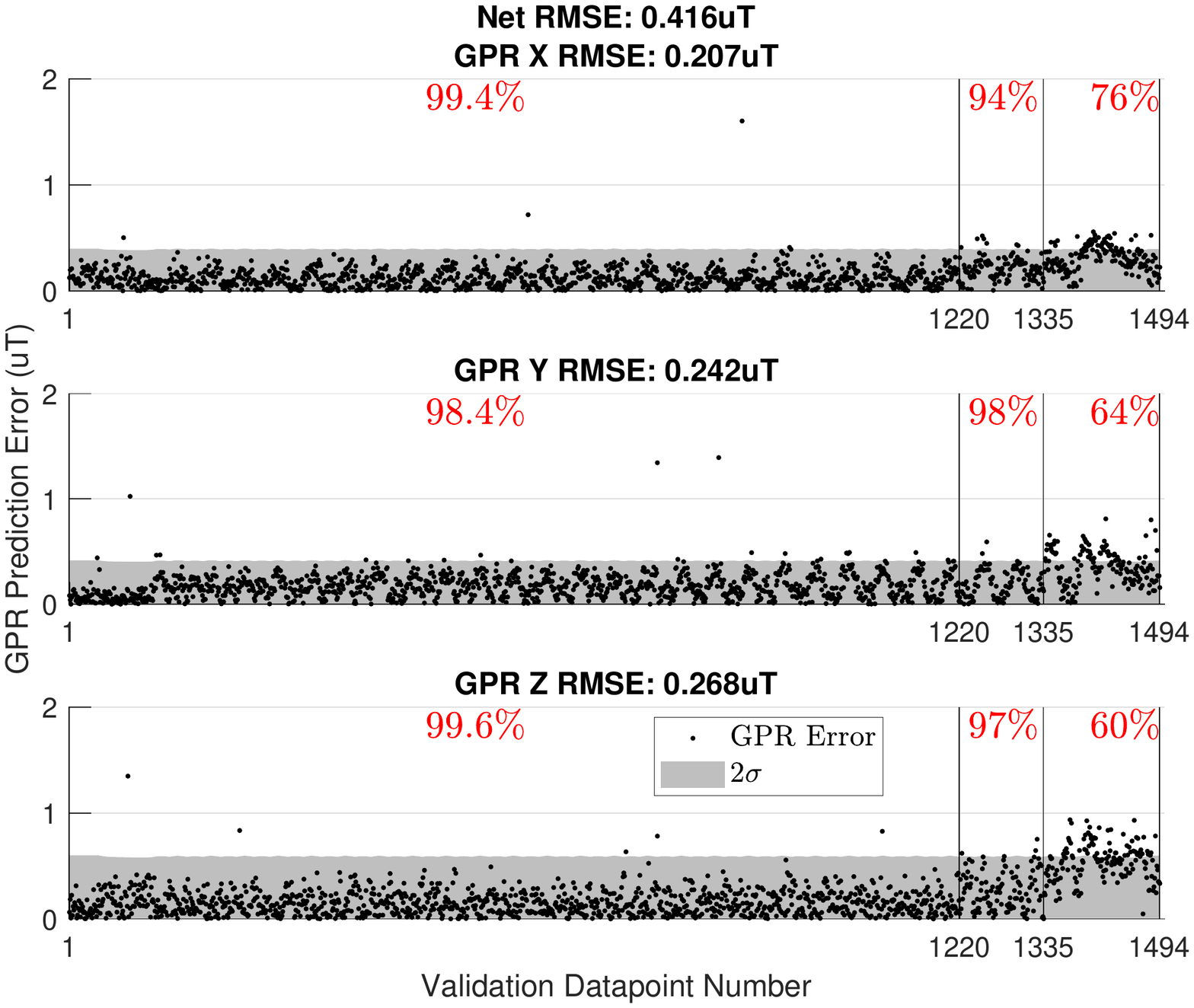}
  \caption{t6\_15. Outliers \{[2.16, 2.56], [2.10], []\}$\mu$T.}
  \label{fig:consistency_metric_3}
 \end{subfigure}
 \hfill
 \begin{subfigure}[b]{0.48\textwidth}
  
  \includegraphics[width=\textwidth,trim= 20 160 20 130, clip]{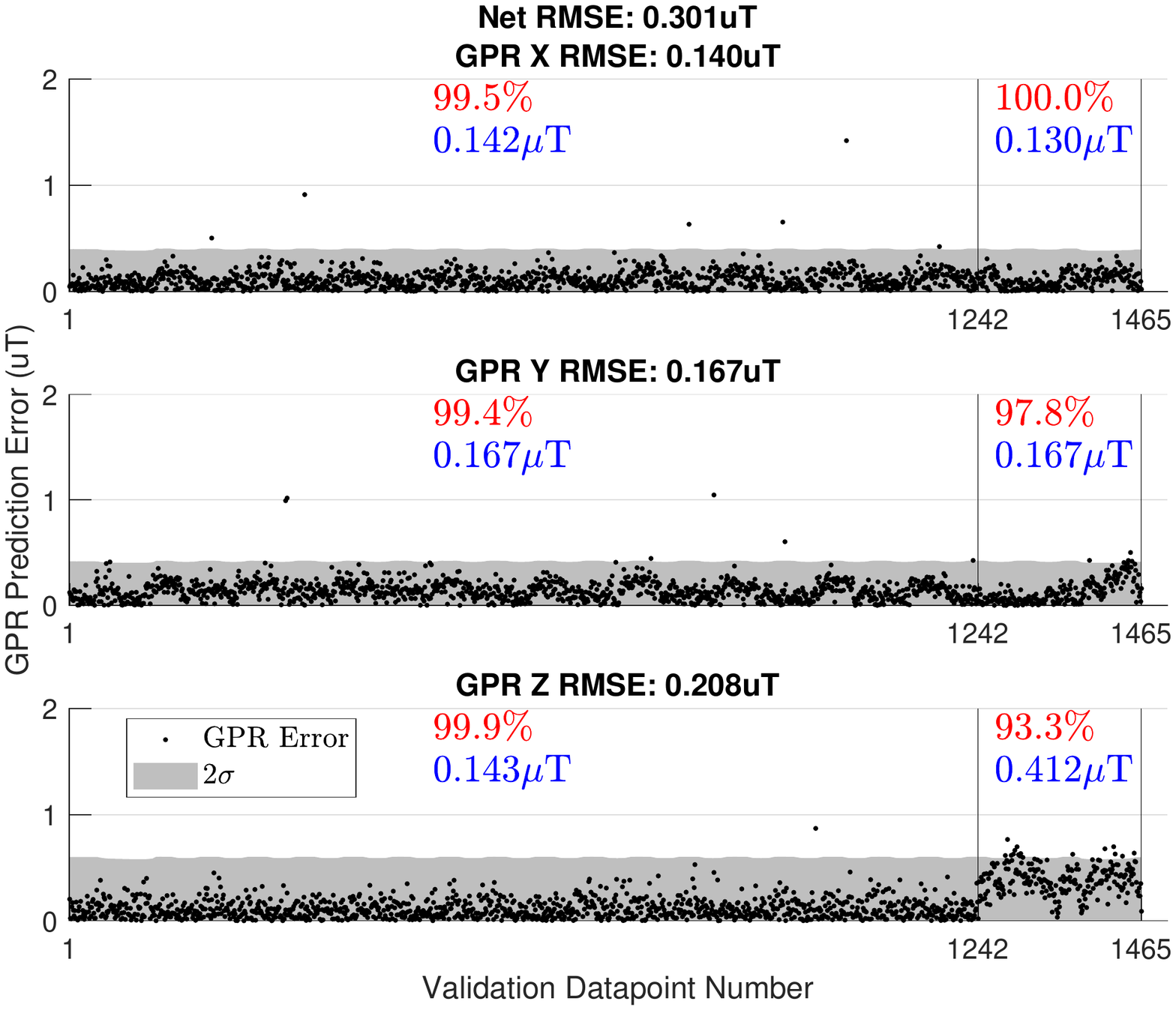}
  \caption{t6\_18. No hidden outliers.}
  \label{fig:consistency_metric_4}
 \end{subfigure}
\caption{Consistency metric (red percentage) across four validation flights. Some error points are larger than 2$\mu$T and are cut off in the graphs. Such outliers are listed for \{ [$\mathcal{GPR}_{x}$], [$\mathcal{GPR}_{y}$], [$\mathcal{GPR}_{z}$] \} in the respective subplot caption.
Empty brackets indicate no hidden outliers for the respective $\mathcal{GPR}$.
}
\label{fig:consistency_metric}
	\vspace{-10pt}
\end{figure*}

Figure \ref{fig:consistency_metric} shows the GPR's performance on the four selected validation flights. We start with a case that easily passes our consistency test (Figure \ref{fig:consistency_metric_1}). Test t6\_11 (Figure \ref{fig:consistency_metric_2}) fails the consistency test since the $\mathcal{GP}_{z}$ does not capture enough of its error within its $2\sigma$ uncertainty. Note that even though $x$ and $y$ GPRs have a great understanding of their error, the poor performance of the $\mathcal{GP}_{z}$ alone means that test t6\_11 is inconsistent with this $3 \rightarrow 3$ map. 
However, if the user needed only the outputs of the $\mathcal{GP}_{x}$ and $\mathcal{GP}_{y}$ (say for localization using magnetic field heading \cite{Suksakulchai2000}), then the poor performance of $\mathcal{GP}_{z}$ may be irrelevant and the map could still be useful for such an application.

Figure \ref{fig:consistency_metric_3} shows an example of the flight-by-flight variation changing partway through a validation flight. In this case, the first 1220 validation points easily pass our consistency metric for all three GPRs.
From 1220 to 1335, things get worse and $\mathcal{GP}_{x}$ barely fails our consistency test. After data point 1335, all three GPRs fail the consistency check. 

This example allows us to emphasize an important point about how we use the consistency metric. Since most of test t6\_15 (up to point 1220) is consistent with the compromise map, computing our 2$\sigma$ percentage across the entire validation set gives (97\%, 95\%, 95\%) for each GPR respectively. Working with these numbers alone (without having Figure \ref{fig:consistency_metric_3} for context), it's tempting to say all of t6\_15 is consistent with the map (although barely). The reality is that since the flight-by-flight variations can change bias values partway through a single flight test, it is sometimes necessary to see if \textit{a portion} of a flight test fails the consistency test rather than checking the 2$\sigma$ metrics across the whole flight.

Finally, validating on t6\_18 gives Figure \ref{fig:consistency_metric_4} which is another gray area in our use of the consistency check. Here, the first 1242 validation points undoubtedly pass the consistency test. However, the remaining points fail the consistency check on the z-component of the magnetic field. 
Note that the $\mathcal{GP}_{y}$ plot in Figure \ref{fig:consistency_metric_4} also changes behavior near index 1380, but our analysis focuses on $\mathcal{GP}_{z}$ since it fails the consistency check before $\mathcal{GP}_{y}$ does.

96\% should be taken as a ``rule of thumb'' rather than a magic threshold where something fundamentally different occurs with our construction of magnetic field maps. 
With this in mind, it is important to consider how the map will be used when deciding whether a validation  test should be rejected on the basis of the consistency check alone. 
Here, for t6\_18 in Figure \ref{fig:consistency_metric_4}, we also include the RMSE values (in blue) for all points before and after 1242 to illustrate the accuracy of our compromise map before and after the mid-flight bias shift.

For our analysis in Ref. \cite{Kuevor2021}, though we did not realize it at the time, these flight-by-flight variations were still present. 
Yet, we were still able to use our magnetic field maps (at the time, trained on a single flight test rather than our new compromise map method) to perform good attitude estimates. 
We believe this is due to the nature of attitude estimation which compares the angle of a measured vector against its corresponding reference vector.
In this case, the flight-by-flight variations did not significantly change the \textit{angle} of the magnetic field reference vectors given by our GPR maps. 

However, our preliminary results on magnetic field position localization show that inconsistencies like those at the end of t6\_15 (Figure \ref{fig:consistency_metric_3}) and the end of  t6\_18 (Figure \ref{fig:consistency_metric_4}) is enough to confuse a particle filter and cause noticeably larger estimation errors.
The use of GPR-based magnetic field maps for position localization (via a particle filter) has been the motivation for much of this paper. 
However, if the intended application is more robust to ``small'' inaccuracies of the GPR map, then a more lenient consistency check threshold may work.

\section{Conclusions and Future Work}
\label{sec:conclusion_future_work}
This work presents practical methods to address vehicle-induced noise when creating and validating GPR-based magnetic field maps with a UAV. First, we showed that certain portions of a vehicle's PID controller can increase the measurable magnetic noise when commanding the motors and ESCs on a flight vehicle. 

Next, we introduced the flight-by-flight magnetic field variations present on our quadrotor and presented two ways to address them. The first involves moving the magnetometer further from the electronics on the vehicle while the second aims to teach the GPR-based map about the flight-by-flight variations through a ``compromise'' map. We show that our compromise map has similar accuracy to the ``intermediate'' map which uses four times as many reference points to estimate the magnetic field through in our working volume. We found that a compromise map trained on observations spaced no more than $S = 0.55$m give comparable accuracy.


Finally, we note that our proposed methods of reduction and incorporation do not encompass all possible mapping errors. 
As such, we introduce a ``consistency check'' as a rule of thumb to help experimentalists quickly assess if magnetic field observations from a particular test dataset are consistent with their magnetic field map.

In the future, we aim to use our new mapping techniques to demonstrate the 3D position localization of a UAV using indoor magnetic field maps. 
Per Section \ref{sec:3_3_vs_3_1_norm_map}, we aim to conduct studies similar to \cite{LeGrand2012, Haverinen2009, Vandermeulen2013} that compare the position localization accuracy of magnetic maps with different output dimensions $m$.
The proposed work would complement our previous work on UAV attitude estimation using indoor field maps \cite{Kuevor2021} which we will consider revisiting with our new mapping methods. 
Additionally, investigating time-varying magnetic fields (due to the position of large, metallic objects like elevators and doors) is important and can be complicated by vehicle-induced magnetic variations that may be confused for a change in the ambient magnetic field.
So far, our work assumed the indoor magnetic field is constant.
This tends to be a good assumption unless large metal objects (like doors and elevators) are moved in the mapped area \cite{Haverinen2009, Solin2015, Michel2018}. As such, we aim to extend our methods into time-varying magnetic field mapping that uses our consistency check (Section \ref{sec:consistency_check}) to decide when a portion of a map may be out of date and needs to be updated.

Finally, a component-wise causal analysis of observed flight-by-flight variations may better help us understand measurable bias and noise sources to modify or eliminate entirely. For example, it is possible that simply using different brands/models of ESCs and motors could make the UAV more magnetically quiet.


\vspace{6pt} 



\authorcontributions{Conceptualization, all; methodology, P.E.K., E.M.A. and J.W.C.; software, P.E.K.; validation, P.E.K.; formal analysis, P.E.K.; investigation, P.E.K.; resources, E.M.A. and J.W.C.; data curation, P.E.K.; writing---original draft, P.E.K.; writing---review and editing, all; visualization, P.E.K.; supervision, M.G., E.M.A., and J.W.C.; funding acquisition, E.M.A and J.W.C. All authors have read and agreed to the published version of the~manuscript.}

\funding{ 
Not applicable.
}

\institutionalreview{Not applicable.}
\informedconsent{Not applicable.}
\dataavailability{ {\color{red} Working on publicly available dataset.} }

\acknowledgments{This work could not have been done without help from my friends in the Michigan eXploration Lab (MXL) and Autonomous Aerospace Systems (A2sys) lab. 
Specifically, thanks to Matthew Romano, Prashin Sharma, Akshay Mathur, Mark Nail, Justin Schachter, Tran Anh Nguyen, and Paul Flanigen for their support in fabricating my test vehicle, tuning the flight controller, assisting with flight tests, developing/soldering PCBs for the magnetometers, and helping to concisely communicating my work through the opening figure.
Special thanks to Mark Moldwin, Alex Hoffmann, and Brady Strabel from the University of Michigan Magnetometer Laboratory (Mag Lab) for assistance with configuring the PNI RM3100 magnetometer for this work.}

\conflictsofinterest{The authors declare no conflict of interest.}





\appendixtitles{no} 
\appendixstart
\appendix
\section[]{Flight Test Name Reference}
\label{apndx:flight_test_names}

The, rather obtuse, naming convention for flight tests in this paper is meant to serve the reader in associating our results with the relevant data in our dataset. Recall that each flight test is denoted as ``tY\_XX'' where \textit{Y} refers to the flight test \textit{series} and \textit{XX} is the two-digit ID of the flight test in that series. The Table \ref{tbl:test_series_dataset_directory} associates each series numbers \textit{Y} used in this paper with their respective directories in our dataset.

\begin{table*}
 \caption{Data used in this paper.} 
 \label{tbl:test_series_dataset_directory}
 \resizebox{0.98\textwidth}{!}{%
 \begin{tabular}{l l l} 
\toprule
  \textbf{Test } & \textbf{Dataset} & \textbf{Use In} \\ 
  \textbf{Series (tY)} & \textbf{Directory} & \textbf{This Paper} \\
 \midrule
  t1 & 20220617\_batteryAndMagTesting & $S_2 = 2$cm\\
  t2 & 20220622\_batteryAndMagPartTwo & $S_2 = 4$cm \\
  t3 & 20220627\_verifyingSufficientMagDist & $S_2 = 6$cm\\
  t4 & 20220707\_8cm\_magAndBatt & $S_2 = 8$cm \\
  t5 & 20220503\_maggieAndEstibonFlights & `noisy' versus `quiet' PID controller gains \\
  t6 & 20220901\_researchData & Intermediate and compromise maps. Consistency metric.\\
\bottomrule
 \end{tabular}}
  \vspace{-10pt}
\end{table*}

\begin{adjustwidth}{-\extralength}{0cm}

\reftitle{References}


\bibliography{refs}

\begin{thebibliography}{999}

\bibitem[Kuevor \em{et~al.}(2021)Kuevor, Cutler, and Atkins]{Kuevor2021}
Kuevor, P.E.; Cutler, J.W.; Atkins, E.M.
\newblock {Improving Attitude Estimation Using
  Gaussian-Process-Regression-Based Magnetic Field Maps}.
\newblock {\em Sensors} {\bf 2021}, {\em 21},~6351.
\newblock {\url{https://doi.org/10.3390/s21196351}}.

\bibitem[Kok and Solin(2018)]{Kok2018}
Kok, M.; Solin, A.
\newblock {Scalable Magnetic Field SLAM in 3D Using Gaussian Process Maps}.
\newblock {\em 2018 21st International Conference on Information Fusion
  (FUSION)} {\bf 2018}, pp. 1353--1360,
  \href{http://xxx.lanl.gov/abs/1804.01926}{{\normalfont [1804.01926]}}.
\newblock {\url{https://doi.org/10.23919/ICIF.2018.8455789}}.

\bibitem[Vallivaara \em{et~al.}(2011)Vallivaara, Haverinen, Kemppainen, and
  Roning]{Vallivaara2011}
Vallivaara, I.; Haverinen, J.; Kemppainen, A.; Roning, J.
\newblock {Magnetic field-based SLAM method for solving the localization
  problem in mobile robot floor-cleaning task}.
\newblock In Proceedings of the 2011 15th International Conference on Advanced
  Robotics (ICAR). IEEE,  2011, pp. 198--203.
\newblock {\url{https://doi.org/10.1109/ICAR.2011.6088632}}.

\bibitem[Vallivaara \em{et~al.}(2010)Vallivaara, Haverinen, Kemppainen, and
  Roning]{Vallivaara2010}
Vallivaara, I.; Haverinen, J.; Kemppainen, A.; Roning, J.
\newblock {Simultaneous localization and mapping using ambient magnetic field}.
\newblock In Proceedings of the 2010 IEEE Conference on Multisensor Fusion and
  Integration. IEEE,  2010, pp. 14--19.
\newblock {\url{https://doi.org/10.1109/MFI.2010.5604465}}.

\bibitem[Akai and Ozaki(2015)]{Akai2015}
Akai, N.; Ozaki, K.
\newblock {Gaussian processes for magnetic map-based localization in
  large-scale indoor environments}.
\newblock {\em IEEE International Conference on Intelligent Robots and Systems}
  {\bf 2015}, {\em 2015-December},~4459--4464.
\newblock {\url{https://doi.org/10.1109/IROS.2015.7354010}}.

\bibitem[Akai and Ozaki(2017)]{Akai2017}
Akai, N.; Ozaki, K.
\newblock {3D magnetic field mapping in large-scale indoor environment using
  measurement robot and Gaussian processes}.
\newblock In Proceedings of the 2017 International Conference on Indoor
  Positioning and Indoor Navigation (IPIN). IEEE,  2017, Vol. 2017-Janua, pp.
  1--7.
\newblock {\url{https://doi.org/10.1109/IPIN.2017.8115960}}.

\bibitem[Suksakulchai \em{et~al.}(2000)Suksakulchai, Thongchai, Wilkes, and
  Kawamura]{Suksakulchai2000}
Suksakulchai, S.; Thongchai, S.; Wilkes, D.; Kawamura, K.
\newblock {Mobile robot localization using an electronic compass for corridor
  environment}.
\newblock In Proceedings of the SMC 2000 Conference Proceedings. 2000 IEEE
  International Conference on Systems, Man and Cybernetics. 'Cybernetics
  Evolving to Systems, Humans, Organizations, and their Complex Interactions'
  (Cat. No.00CH37166). IEEE,  2000, Vol.~5, pp. 3354--3359.
\newblock {\url{https://doi.org/10.1109/ICSMC.2000.886523}}.

\bibitem[Li \em{et~al.}(2012)Li, Gallagher, Dempster, and Rizos]{Li2012}
Li, B.; Gallagher, T.; Dempster, A.G.; Rizos, C.
\newblock {How feasible is the use of magnetic field alone for indoor
  positioning?}
\newblock In Proceedings of the 2012 International Conference on Indoor
  Positioning and Indoor Navigation (IPIN). IEEE,  2012, pp. 1--9.
\newblock {\url{https://doi.org/10.1109/IPIN.2012.6418880}}.

\bibitem[Haverinen and Kemppainen(2009)]{Haverinen2009}
Haverinen, J.; Kemppainen, A.
\newblock {Global indoor self-localization based on the ambient magnetic
  field}.
\newblock {\em Robotics and Autonomous Systems} {\bf 2009}, {\em
  57},~1028--1035.
\newblock {\url{https://doi.org/10.1016/j.robot.2009.07.018}}.

\bibitem[Wu \em{et~al.}(2019)Wu, Wen, Peng, Tang, and Wang]{Wu2019}
Wu, Z.; Wen, M.; Peng, G.; Tang, X.; Wang, D.
\newblock {Magnetic-Assisted Initialization for Infrastructure-free Mobile
  Robot Localization}.
\newblock In Proceedings of the 2019 IEEE International Conference on
  Cybernetics and Intelligent Systems (CIS) and IEEE Conference on Robotics,
  Automation and Mechatronics (RAM). IEEE,  2019, pp. 518--523,
  \href{http://xxx.lanl.gov/abs/1911.09313}{{\normalfont [1911.09313]}}.
\newblock {\url{https://doi.org/10.1109/CIS-RAM47153.2019.9095809}}.

\bibitem[Robertson \em{et~al.}(2013)Robertson, Frassl, Angermann, Doniec,
  Julian, Puyol, Khider, Lichtenstern, and Bruno]{Robertson2013}
Robertson, P.; Frassl, M.; Angermann, M.; Doniec, M.; Julian, B.J.; Puyol,
  M.G.; Khider, M.; Lichtenstern, M.; Bruno, L.
\newblock {Simultaneous localization and mapping for pedestrians using
  distortions of the local magnetic field intensity in large indoor
  environments}.
\newblock {\em 2013 International Conference on Indoor Positioning and Indoor
  Navigation, IPIN 2013} {\bf 2013}.
\newblock {\url{https://doi.org/10.1109/IPIN.2013.6817910}}.

\bibitem[Wang \em{et~al.}(2017)Wang, Zhang, Liu, Dong, and Xu]{Wang2017}
Wang, X.; Zhang, C.; Liu, F.; Dong, Y.; Xu, X.
\newblock {Exponentially weighted particle filter for simultaneous localization
  and mapping based on magnetic field measurements}.
\newblock {\em IEEE Transactions on Instrumentation and Measurement} {\bf
  2017}, {\em 66},~1658--1667.
\newblock {\url{https://doi.org/10.1109/TIM.2017.2664538}}.

\bibitem[Afzal \em{et~al.}(2011)Afzal, Renaudin, and Lachapelle]{Afzal2011}
Afzal, M.H.; Renaudin, V.; Lachapelle, G.
\newblock {Multi-magnetometer based perturbation mitigation for indoor
  orientation estimation}.
\newblock {\em Navigation, Journal of the Institute of Navigation} {\bf 2011},
  {\em 58},~279--292.
\newblock {\url{https://doi.org/10.1002/j.2161-4296.2011.tb02586.x}}.

\bibitem[Yin and Zhang(2018)]{Yin2018}
Yin, G.; Zhang, L.
\newblock {Magnetic heading compensation method based on magnetic
  interferential signal inversion}.
\newblock {\em Sensors and Actuators A: Physical} {\bf 2018}, {\em 275},~1--10.
\newblock {\url{https://doi.org/10.1016/j.sna.2018.03.043}}.

\bibitem[Karimi \em{et~al.}(2020)Karimi, Babaians, Oelsch, Aykut, and
  Steinbach]{Karimi2020}
Karimi, M.; Babaians, E.; Oelsch, M.; Aykut, T.; Steinbach, E.
\newblock {Skewed-redundant Hall-effect Magnetic Sensor Fusion for
  Perturbation-free Indoor Heading Estimation}.
\newblock {\em Proceedings - 4th IEEE International Conference on Robotic
  Computing, IRC 2020} {\bf 2020}, pp. 367--374.
\newblock {\url{https://doi.org/10.1109/IRC.2020.00064}}.

\bibitem[Hanley \em{et~al.}(2021)Hanley, {De Oliveira}, Zhang, Kim, Wei, and
  Bretl]{Hanley2021}
Hanley, D.; {De Oliveira}, A.S.; Zhang, X.; Kim, D.H.; Wei, Y.; Bretl, T.
\newblock {The Impact of Height on Indoor Positioning with Magnetic Fields}.
\newblock {\em IEEE Transactions on Instrumentation and Measurement} {\bf
  2021}, {\em 70}.
\newblock {\url{https://doi.org/10.1109/TIM.2021.3059317}}.

\bibitem[Brzozowski \em{et~al.}(2016)Brzozowski, Kazmierczak, Rochala, Wojda,
  and Wojtowicz]{Brzozowski2016}
Brzozowski, B.; Kazmierczak, K.; Rochala, Z.; Wojda, M.; Wojtowicz, K.
\newblock {A concept of UAV indoor navigation system based on magnetic field
  measurements}.
\newblock {\em 3rd IEEE International Workshop on Metrology for Aerospace,
  MetroAeroSpace 2016 - Proceedings} {\bf 2016}, pp. 636--640.
\newblock {\url{https://doi.org/10.1109/MetroAeroSpace.2016.7573291}}.

\bibitem[Brzozowski and Kazmierczak(2017)]{Brzozowski2017}
Brzozowski, B.; Kazmierczak, K.
\newblock {Magnetic field mapping as a support for UAV indoor navigation
  system}.
\newblock In Proceedings of the 2017 IEEE International Workshop on Metrology
  for AeroSpace (MetroAeroSpace). IEEE,  2017, pp. 583--588.
\newblock {\url{https://doi.org/10.1109/MetroAeroSpace.2017.7999535}}.

\bibitem[Frassl \em{et~al.}(2013)Frassl, Angermann, Lichtenstern, Robertson,
  Julian, and Doniec]{Frassl2013}
Frassl, M.; Angermann, M.; Lichtenstern, M.; Robertson, P.; Julian, B.J.;
  Doniec, M.
\newblock {Magnetic maps of indoor environments for precise localization of
  legged and non-legged locomotion}.
\newblock {\em IEEE International Conference on Intelligent Robots and Systems}
  {\bf 2013}, pp. 913--920.
\newblock {\url{https://doi.org/10.1109/IROS.2013.6696459}}.

\bibitem[Liu \em{et~al.}(2021)Liu, Yu, Huang, Shi, Gao, and He]{Liu2021}
Liu, G.; Yu, B.; Huang, L.; Shi, L.; Gao, X.; He, L.
\newblock {Human-Interactive Mapping Method for Indoor Magnetic Based on
  Low-Cost MARG Sensors}.
\newblock {\em IEEE Transactions on Instrumentation and Measurement} {\bf
  2021}, {\em 70}.
\newblock {\url{https://doi.org/10.1109/TIM.2021.3052026}}.

\bibitem[Gozick \em{et~al.}(2011)Gozick, Subbu, Dantu, and
  Maeshiro]{Gozick2011}
Gozick, B.; Subbu, K.P.; Dantu, R.; Maeshiro, T.
\newblock {Magnetic maps for indoor navigation}.
\newblock {\em IEEE Transactions on Instrumentation and Measurement} {\bf
  2011}, {\em 60},~3883--3891.
\newblock {\url{https://doi.org/10.1109/TIM.2011.2147690}}.

\bibitem[Ayanoglu \em{et~al.}(2018)Ayanoglu, Schneider, and
  Eitel]{Ayanoglu2018}
Ayanoglu, A.; Schneider, D.M.; Eitel, B.
\newblock {Crowdsourcing-Based Magnetic Map Generation for Indoor
  Localization}.
\newblock In Proceedings of the 2018 International Conference on Indoor
  Positioning and Indoor Navigation (IPIN). IEEE,  2018, pp. 1--8.
\newblock {\url{https://doi.org/10.1109/IPIN.2018.8533832}}.

\bibitem[Almeida \em{et~al.}(2021)Almeida, Pedrosa, and Curado]{Almeida2021}
Almeida, D.; Pedrosa, E.; Curado, F.
\newblock {Magnetic Mapping for Robot Navigation in Indoor Environments}.
\newblock {\em 2021 International Conference on Indoor Positioning and Indoor
  Navigation, IPIN 2021} {\bf 2021}.
\newblock {\url{https://doi.org/10.1109/IPIN51156.2021.9662528}}.

\bibitem[Zheng \em{et~al.}(2021)Zheng, Li, Xing, and Zhang]{Zheng2021}
Zheng, Y.; Li, S.; Xing, K.; Zhang, X.
\newblock {Unmanned aerial vehicles for magnetic surveys: A review on platform
  selection and interference suppression}.
\newblock {\em Drones} {\bf 2021}, {\em 5}.
\newblock {\url{https://doi.org/10.3390/drones5030093}}.

\bibitem[Versteeg and McKay(2007)]{Versteeg2007}
Versteeg, R.; McKay, M.
\newblock {Feasibility Study for an Autonomous UAV -Magnetometer System --
  Final Report on SERDP SEED 1509:2206}.
\newblock Technical Report September, Idaho National Laboratory,  2007.
\newblock {\url{https://doi.org/10.2172/923485}}.

\bibitem[Koyama \em{et~al.}(2013)Koyama, Kaneko, Ohminato, Yanagisawa,
  Watanabe, and Takeo]{Koyama2013}
Koyama, T.; Kaneko, T.; Ohminato, T.; Yanagisawa, T.; Watanabe, A.; Takeo, M.
\newblock {An aeromagnetic survey of Shinmoe-dake volcano, Kirishima, Japan,
  after the 2011 eruption using an unmanned autonomous helicopter}.
\newblock {\em Earth, Planets and Space} {\bf 2013}, {\em 65},~657--666.
\newblock {\url{https://doi.org/10.5047/eps.2013.03.005}}.

\bibitem[Vasiljevic \em{et~al.}(2022)Vasiljevic, Martinovic, Batos, and
  Bogdan]{Vasiljevic2022}
Vasiljevic, G.; Martinovic, D.; Batos, M.; Bogdan, S.
\newblock {Validation of two-wire power line UAV localization based on the
  magnetic field strength}.
\newblock {\em 2022 International Conference on Unmanned Aircraft Systems,
  ICUAS 2022} {\bf 2022}, pp. 434--441.
\newblock {\url{https://doi.org/10.1109/ICUAS54217.2022.9836106}}.

\bibitem[Li \em{et~al.}(2019)Li, Zahran, Zhuang, Gao, Luo, He, Pei, Chen, and
  El-Sheimy]{Li2019}
Li, Y.; Zahran, S.; Zhuang, Y.; Gao, Z.; Luo, Y.; He, Z.; Pei, L.; Chen, R.;
  El-Sheimy, N.
\newblock {IMU/magnetometer/barometer/mass-flow sensor integrated indoor
  quadrotor UAV localization with robust velocity updates}.
\newblock {\em Remote Sensing} {\bf 2019}, {\em 11},~1--22.
\newblock {\url{https://doi.org/10.3390/rs11070838}}.

\bibitem[Zahran \em{et~al.}(2019)Zahran, Moussa, Sesay, and
  El-Sheimy]{Zahran2019}
Zahran, S.; Moussa, A.M.; Sesay, A.B.; El-Sheimy, N.
\newblock {A New Velocity Meter Based on Hall Effect Sensors for UAV Indoor
  Navigation}.
\newblock {\em IEEE Sensors Journal} {\bf 2019}, {\em 19},~3067--3076.
\newblock {\url{https://doi.org/10.1109/JSEN.2018.2890094}}.

\bibitem[Tiemann \em{et~al.}(2018)Tiemann, Ramsey, and Wietfeld]{Tiemann2018}
Tiemann, J.; Ramsey, A.; Wietfeld, C.
\newblock {Enhanced UAV indoor navigation through SLAM-Augmented UWB
  Localization}.
\newblock {\em 2018 IEEE International Conference on Communications Workshops,
  ICC Workshops 2018 - Proceedings} {\bf 2018}, pp. 1--6.
\newblock {\url{https://doi.org/10.1109/ICCW.2018.8403539}}.

\bibitem[Niu \em{et~al.}(2021)Niu, Zhang, Guo, Pun, and Chen]{Niu2021}
Niu, G.; Zhang, J.; Guo, S.; Pun, M.O.; Chen, C.S.
\newblock {UAV-Enabled 3D Indoor Positioning and Navigation Based on VLC}.
\newblock In Proceedings of the ICC 2021 - IEEE International Conference on
  Communications. IEEE,  2021, pp. 1--6.
\newblock {\url{https://doi.org/10.1109/ICC42927.2021.9500633}}.

\bibitem[Lipovsk{\'{y}} \em{et~al.}(2021)Lipovsk{\'{y}}, Draganov{\'{a}},
  Novotň{\'{a}}k, Szőke, and Fiľko]{Lipovsky2021}
Lipovsk{\'{y}}, P.; Draganov{\'{a}}, K.; Novotň{\'{a}}k, J.; Szőke, Z.;
  Fiľko, M.
\newblock {Indoor mapping of magnetic fields using uav equipped with fluxgate
  magnetometer}.
\newblock {\em Sensors} {\bf 2021}, {\em 21}.
\newblock {\url{https://doi.org/10.3390/s21124191}}.

\bibitem[Bla{\v{z}}ek \em{et~al.}(2018)Bla{\v{z}}ek, Lipovsk{\'{y}},
  He{\v{s}}ko, and Rep{\v{c}}{\'{i}}k]{Blazek2018}
Bla{\v{z}}ek, J.; Lipovsk{\'{y}}, P.; He{\v{s}}ko, F.; Rep{\v{c}}{\'{i}}k, D.
\newblock {Electromagnetic image of small UAV in very low frequency range}.
\newblock {\em Journal of Electrical Engineering} {\bf 2018}, {\em
  69},~438--441.
\newblock {\url{https://doi.org/10.2478/jee-2018-0069}}.

\bibitem[Rasmussen and Williams(2006)]{carledwardrasmussen2005}
Rasmussen, C.E.; Williams, C.K.I.
\newblock {\em Gaussian Processes for Machine Learning}; MIT Press,  2006.

\bibitem[Springmann and Cutler(2012)]{Springmann2012_magCalibration}
Springmann, J.C.; Cutler, J.W.
\newblock {Attitude-Independent Magnetometer Calibration with Time-Varying
  Bias}.
\newblock {\em Journal of Guidance, Control, and Dynamics} {\bf 2012}, {\em
  35},~1080--1088.
\newblock {\url{https://doi.org/10.2514/1.56726}}.

\bibitem[Wu \em{et~al.}(2020)Wu, Pei, Li, Gao, and Bai]{Wu2020}
Wu, H.; Pei, X.; Li, J.; Gao, H.; Bai, Y.
\newblock An improved magnetometer calibration and compensation method based on
  Levenberg–Marquardt algorithm for multi-rotor unmanned aerial vehicle.
\newblock {\em Measurement and Control} {\bf 2020}, {\em 53},~276--286.
\newblock {\url{https://doi.org/10.1177/0020294019890627}}.

\bibitem[Team and Survey(2020)]{WMM2020}
Team, N.G.M.; Survey, B.G.
\newblock World Magnetic Model 2020,  2020.
\newblock {\url{https://doi.org/10.25921/11v3-da71}}.

\bibitem[Regoli \em{et~al.}(2018)Regoli, Moldwin, Pellioni, Bronner, Hite,
  Sheinker, and Ponder]{Regoli2018}
Regoli, L.H.; Moldwin, M.B.; Pellioni, M.; Bronner, B.; Hite, K.; Sheinker, A.;
  Ponder, B.M.
\newblock {Investigation of a low-cost magneto-inductive magnetometer for space
  science applications}.
\newblock {\em Geoscientific Instrumentation, Methods and Data Systems} {\bf
  2018}, {\em 7},~129--142.
\newblock {\url{https://doi.org/10.5194/gi-7-129-2018}}.

\bibitem[Romano \em{et~al.}(2022)Romano, Ye, Pye, and Atkins]{Romano2022}
Romano, M.; Ye, A.; Pye, J.; Atkins, E.
\newblock Cooperative Multilift Slung Load Transportation Using Haptic
  Admittance Control Guidance.
\newblock {\em Journal of Guidance, Control, and Dynamics} {\bf 2022}, pp.
  1--14.
\newblock {\url{https://doi.org/10.2514/1.G006587}}.

\bibitem[{Le Grand} and Thrun(2012)]{LeGrand2012}
{Le Grand}, E.; Thrun, S.
\newblock {3-Axis magnetic field mapping and fusion for indoor localization}.
\newblock In Proceedings of the 2012 IEEE International Conference on
  Multisensor Fusion and Integration for Intelligent Systems (MFI). IEEE,
  2012, pp. 358--364.
\newblock {\url{https://doi.org/10.1109/MFI.2012.6343024}}.

\bibitem[Vandermeulen \em{et~al.}(2013)Vandermeulen, Vercauteren, and
  Weyn]{Vandermeulen2013}
Vandermeulen, D.; Vercauteren, C.; Weyn, M.
\newblock {Indoor localization Using a Magnetic Flux Density Map of a Building
  Feasibility study of geomagnetic indoor localization}.
\newblock {\em AMBIENT 2013 : The Third International Conference on Ambient
  Computing, Applications, Services and Technologies Magnetic} {\bf 2013}, pp.
  42--49.

\bibitem[Solin \em{et~al.}(2015)Solin, Kok, Wahlstr{\"{o}}m, Sch{\"{o}}n, and
  S{\"{a}}rkk{\"{a}}]{Solin2015}
Solin, A.; Kok, M.; Wahlstr{\"{o}}m, N.; Sch{\"{o}}n, T.B.; S{\"{a}}rkk{\"{a}},
  S.
\newblock {Modeling and interpolation of the ambient magnetic field by Gaussian
  processes}.
\newblock {\em IEEE Transactions on Robotics} {\bf 2015}, {\em 34},~1112--1127,
   \href{http://xxx.lanl.gov/abs/1509.04634}{{\normalfont [1509.04634]}}.
\newblock {\url{https://doi.org/10.1109/TRO.2018.2830326}}.

\bibitem[Michel \em{et~al.}(2018)Michel, Genev{\`{e}}s, Fourati, and
  Laya{\"{i}}da]{Michel2018}
Michel, T.; Genev{\`{e}}s, P.; Fourati, H.; Laya{\"{i}}da, N.
\newblock {Attitude estimation for indoor navigation and augmented reality with
  smartphones}.
\newblock {\em Pervasive and Mobile Computing} {\bf 2018}, {\em 46},~96--121.
\newblock {\url{https://doi.org/10.1016/j.pmcj.2018.03.004}}.

\end{thebibliography}

%


\end{adjustwidth}
\end{document}